%% file: smart_arxiv_v2.tex
\documentclass[10pt,twocolumn,letterpaper]{article}

\usepackage[pagenumbers]{cvpr} 

\usepackage[accsupp]{axessibility} 
\usepackage{graphicx}
\usepackage{amsmath}
\usepackage{amssymb}
\usepackage{booktabs}
\usepackage{xspace}
\usepackage{epigraph}
\usepackage{caption}
\usepackage{subcaption}

\usepackage[dvipsnames]{xcolor}
\usepackage{multirow}
\usepackage{enumitem}

\usepackage{pifont}
\newcommand{\cmark}{\green{\ding{51}}}%
\newcommand{\xmark}{\red{\ding{55}}}%

\input{definitions}
\usepackage[pagebackref,breaklinks,colorlinks]{hyperref}

\usepackage[capitalize]{cleveref}
\crefname{section}{Sec.}{Secs.}
\Crefname{section}{Section}{Sections}
\Crefname{table}{Table}{Tables}
\crefname{table}{Tab.}{Tabs.}

\def\confName{CVPR}
\def\confYear{2023}

\begin{document}

\title{Are Deep Neural Networks SMARTer than Second Graders?}
\author{Anoop Cherian$^1$\quad Kuan-Chuan Peng$^1$\quad Suhas Lohit$^1$\quad Kevin A. Smith$^2$\quad Joshua B. Tenenbaum$^2$\\
$^1$Mitsubishi Electric Research Labs (MERL)\quad
$^2$Massachusetts Institute of Technology (MIT)\\
{\tt\small \{cherian, kpeng, slohit\}@merl.com}\qquad
{\tt\small \{k2smith, jbt\}@mit.edu}\\
\url{http://smartdataset.github.io/smart101}
}

\maketitle

\begin{abstract}
Recent times have witnessed an increasing number of applications of deep neural networks towards solving tasks that require superior cognitive abilities, e.g., playing Go, generating art, ChatGPT, etc. Such a dramatic progress raises the question: \emph{how generalizable are neural networks in solving problems that demand broad skills?} To answer this question, we propose \shortname: a~\emph{\fullname} and the associated \name dataset, for evaluating the abstraction, deduction, and generalization abilities of neural networks in solving visuo-linguistic puzzles designed specifically for children in the 6--8 age group. Our dataset consists of 101 unique puzzles; each puzzle comprises a picture and a question, and their solution needs a mix of several elementary skills, including arithmetic, algebra, and spatial reasoning, among others. To scale our dataset towards training deep neural networks, we programmatically generate  entirely new instances for each puzzle, while retaining their solution algorithm. To benchmark performances on \name, we propose a vision and language meta-learning model using varied state-of-the-art backbones. Our experiments reveal that while powerful deep models offer reasonable performances on puzzles in a supervised setting, they are not better than random accuracy when analyzed for generalization. We also evaluate the recent ChatGPT and other large language models on a part of \name and find that while these models show convincing reasoning abilities, the answers are often incorrect.
\end{abstract}

\vspace*{-0.7cm}
\section{Introduction}
\epigraph{``\small{\emph{An attempt will be made to find how to make machines use language, form abstractions and concepts, solve kinds of problems now reserved for humans, and improve themselves.}}''}{\textit{The Dartmouth Summer Project on AI, 1956}}

Deep learning powered AI systems have been increasing in their data modeling abilities at an ever more vigor in the recent times, with compelling applications emerging frequently, many of which may even seem to challenge human abilities. A few notable such feats include but are not limited to game playing (\eg, AlphaGo~\cite{silver2016mastering}), language-guided image generation (\eg, the recent DALLE-2~\cite{ramesh2022hierarchical} and ImageGen~\cite{saharia2022photorealistic}), creative story writing  (\eg, using GPT-3~\cite{brown2020language}), solving university level math problems~\cite{drori2022neural}, algorithmic inference~\cite{fawzi2022discovering}, and general question-answering/dialog (\eg, ChatGPT~\cite{openai2023gpt4} and variants). Such impressive performances have prompted an introspection into the foundation of what constitutes artificial intelligence and deriving novel tasks that could challenge deep models further~\cite{lake2017building,rule2020child,chollet2019measure,mitchell2021abstraction}.

\begin{figure}[t]
\centering
\includegraphics[width=6.5cm,trim={3cm 2cm 3.5cm 2cm},clip]{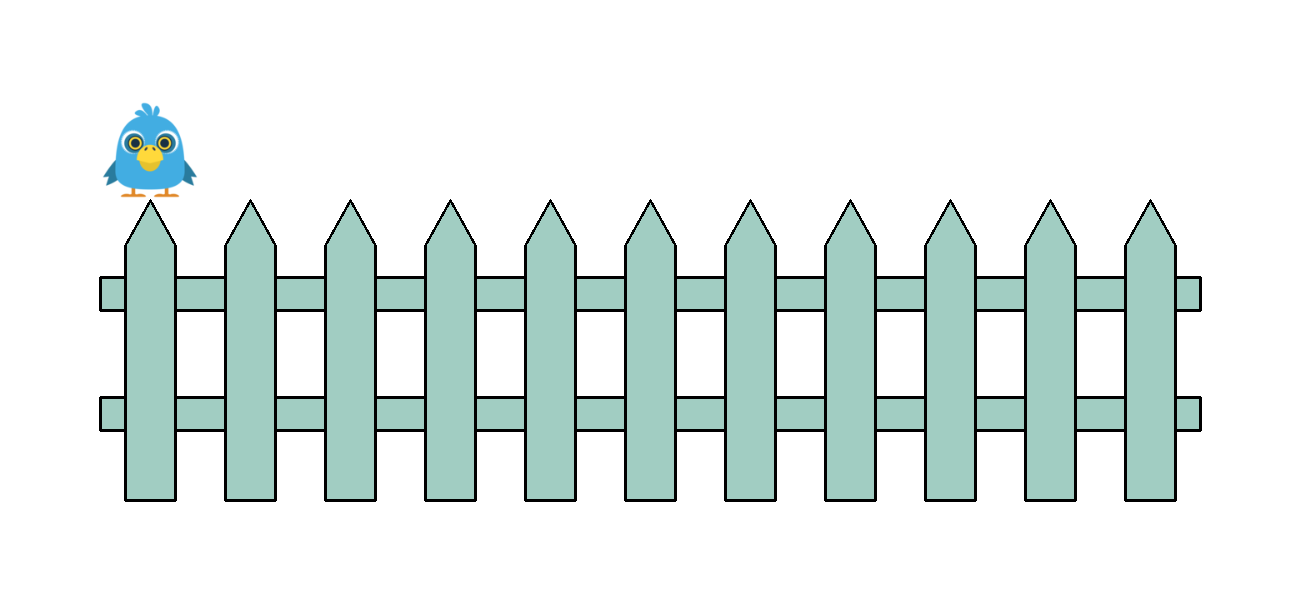}
\vspace*{-0.4cm}
\caption*{\textbf{Question:} \emph{Bird Bobbie jumps on a fence from the post on the left end to the other end. Each jump takes him 4 seconds. He makes 4 jumps ahead and then 1 jump back. Then he again makes 4 jumps ahead and 1 jump back, and so on. In how many seconds can Bobbie get from one end to the other end?}\\
\textbf{Answer Options:}~A: 64\quad	B: 48\quad	C: 56\quad	D: 68\quad	E: 72}
\vspace{-2.5mm}
\caption{An example puzzle instance from our \name dataset generated using our programmatic augmentation method. Solving this puzzle needs various skills such as \emph{counting} the number of posts, \emph{spatially} locating \emph{Bobbie}, and using the \emph{details in the question} to derive an algorithm for the solution. At a foundational level, a reasoning agent needs to recognize abstracted objects such as posts and identify the \emph{bird}.  The answer is shown below.\protect\footnotemark}
\vspace{-1mm}
\label{fig:teaser}
\end{figure}
\footnotetext{The answer to the puzzle in Figure~\ref{fig:teaser} is: C.}

While deep neural networks offer compelling performances on specialized tasks on which they are trained on, (i) how well do they model abstract data, attend on key entities, and transfer knowledge to solve new problems? (ii) how fluid are they in acquiring new skills? and (iii) how effective are they in the use of language for visual reasoning? We task ourselves to understand and seek a way to answer these questions for state-of-the-art (SOTA) vision and language deep learning models. An approach that has been taken several times in the past is to design specialized datasets that can measure the cognitive abilities of well-trained neural networks. For example, in CLEVR~\cite{johnson2017clevr}, a diagnostic dataset is proposed that comprises visuo-linguistic spatial reasoning problems. The abstraction abilities of neural networks have been explored towards solving types of Bongard problems~\cite{nie2020bongard,jiang2022bhoi} and human IQ puzzles (e.g., Ravens progressive matrices) have been extended to evaluate neural reasoning abilities ~\cite{zhang2019raven,benny2021scale,spratley2020closer,teney2020vprom,hu2021stratified,barrett2018measuring,zhuo2021unsupervised,pekar2020generating}. However, while the puzzles in these prior works are often seemingly diverse, they are often confined to a common setting and may need only specialized skill sets, bringing in inductive biases that could be exploited by well-crafted deep learning models, thereby solving such puzzles with near perfect accuracy~\cite{shi2019explainable,spratley2020closer}.  

In this paper, we take a look back at the foundations of intelligence, by asking the question: \emph{Are state-of-the-art deep neural networks capable of emulating the thinking process of even young children?}  To gain insights into answering this question, we introduce the \fullname (SMART) -- a visuo-linguistic task and the associated \name dataset built from 101 distinct children's puzzles. As this is the first step in this direction, we keep the puzzles simple -- to ensure this, we took inspiration from the puzzles in the Math Kangaroo USA Olympiad~\cite{kangaroo} which has puzzle sets professionally designed for children in the age group of 6--8. Each puzzle in our dataset has a picture describing the problem setup and an associated natural language question. To solve the puzzle, one needs to use the question to gather details from the picture and infer a simple mathematical algorithm that leads to a solution to be matched against multiple answer options. In Figure~\ref{fig:teaser}, we illustrate our task with an example puzzle from our dataset. Unlike prior datasets with similar goals, each of the 101 puzzles in our dataset is distinct and needs a broad range of elementary mathematical skills for their solutions, including skills in algebra, basic arithmetic, geometry, ordering, as well as foundational skills to interpret abstract images, and execute counting, spatial reasoning, pattern matching, and occlusion reasoning. To the best of our knowledge, this is the first dataset that offers such a richly diverse set of visuo-linguistic puzzles in an open-world setting, with a psychometric control on their difficulty levels against human performance.\footnote{This is derived from the assumption that the puzzles are professionally designed with a particular audience in mind.} 

To benchmark performances on the \name dataset, we propose an end-to-end meta-learning based neural network~\cite{finn2017model}, where we use a SOTA pre-trained image encoder backbone (\eg, Transformers/ResNets) to embed the picture part of the puzzles, and a strong large language model (\eg, GPT/BERT) to model the questions. As each puzzle may have a different range for their answers (\eg, selection from a few choices, sequential answers, \etc), we propose to treat each puzzle as a separate task, with task-specific neural heads and training objectives, while a common vision-language backbone is used on all the puzzles.
 
We provide experiments using our learning framework under various evaluation settings, analyzing the ability of SOTA vision and language backbones for: (i) in-distribution generalization, when training and test data are from the same distributions of puzzle instances, and out-of-distribution generalization, when training and test data are from: (ii) distinct answer distributions, or (iii) different puzzles. We find the backbones performing poorly in our model on (i) and (ii), while failing entirely on (iii), suggesting that solving our dataset would demand novel research directions into algorithmic reasoning.

We experiment on various settings, evaluating the ability of our model to (i) solve puzzles when trained and tested on the same distribution of instances, (ii) out of distribution generalization when training and testing data are disjoint at the answer level, and (iii) out of distribution generalization when the training and testing sets are disjoint at the puzzle levels. We find that our model performs poorly on the tasks (i) and (ii), while failing entirely on (iii), suggesting that solving our dataset would demand novel research directions into neural abstractions and algorithmic reasoning abilities.  

We summarize below the key contributions of this paper.
\begin{enumerate}
[topsep=0pt,itemsep=-1ex,partopsep=1ex,parsep=1ex]
    \item With the goal of making progress towards improving the visuo-linguistic algorithmic reasoning abilities of neural networks, we introduce a novel task: \shortname, and the associated large-scale \name dataset.
    \item We propose a programmatic augmentation strategy for replicating abstract puzzles. 
    \item We design a baseline meta-solver neural architecture for solving the puzzles in our task.
    \item We present experiments using our approach in various algorithmic generalization settings, bringing out key insights on the performance of SOTA neural networks on this task. We also compare performances against humans and using large language models.
\end{enumerate}

\section{Related works}
To set the stage, we briefly review below a few prior methods and datasets proposed towards understanding the reasoning abilities of deep neural networks. 
\begin{table*}[h]
    \centering
    \resizebox{.98\textwidth}{!}{
    \begin{tabular}{@{}c@{}c@{}c@{}c@{}}
    \toprule
    Dataset & Involve language & Dataset size & Task nature\\
    \midrule
    Bongard-LOGO~\cite{nie2020bongard} & \xmark & 12K & few-shot concepts, abstract shape reasoning \\
    Bongard-HOI~\cite{jiang2022bhoi} & \xmark & 53K & few-shot concepts, human-object interaction \\
    ARC~\cite{chollet2019measure} & \xmark &800 & generate image based on abstract rules \\
    Machine Number Sense~\cite{zhang2020machine} & \xmark & 280K & solving arithmetic problems\\
    RAVEN~\cite{zhang2019raven} & \xmark &70K & finding next image in sequence\\
    Image riddles~\cite{aditya2018combining} & \cmark (fixed question) &3333 & finding common linguistic descriptions \\
    VLQA~\cite{sampat2020vlqa} &\cmark (variable questions) &9267 &spatio-temporal reasoning, info lookup, mathematical, logical, causality, analogy, \etc\\
    PororoQA~\cite{kim2017deepstory} &\cmark (variable questions) &8913 &reason from cartoon videos about action, person, abstract, detail, location, \etc\\
    CLEVR~\cite{johnson2017clevr} &\cmark (variable questions) & 100K & exist, count, query attributes, compare integers/attribute\\
    \textbf{\name} (ours) & \cmark (variable questions) & 200K & 8 predominant algorithmic skills and their compositions (see Figure~\ref{fig:stats})  \\
    \bottomrule
    \end{tabular}
    }
    \vspace{-2mm}
    \caption{Comparison between our \name dataset with existing datasets related to visual reasoning.}
    \vspace{1.2mm}
    \label{tab:datasets}
\end{table*}

\noindent\textbf{Solving IQ puzzles:} via creating computer programs has been a dream since the early days of exploration into AI~\cite{hernandez2016computer,minsky1961steps,minsky1988society}; Evan's ANALOGY~\cite{evans1964program} and Hofstader's CopyCat, among others~\cite{hofstadter1995fluid} are famous tasks in this direction. With the resurgence of deep learning, there have been several attempts at re-considering such puzzles, with varied success. In Table~\ref{tab:datasets}, we briefly review such tasks and datasets (see Ma{\l}ki{\'n}ski and Ma{\'n}dziuk~\cite{malkinski2022review} for an in-depth survey). While, the goal of these works have been towards capturing human cognition through machine learning models, their tasks are often specialized and when provided enough data, the neural networks apparently leverage shortcomings in the dataset towards achieving very high accuracy~\cite{zhang2019gap,hernandez2016computer,spratley2020closer}, defaulting the original goals.

\noindent\textbf{Neuro-symbolic learning and program synthesis:} approaches consider solving complex tasks via decomposing a scene into entities and synthesizing computer programs that operate on these entities; thereby plausibly emulating human reasoning. The DreamCoder approach~\cite{ellis2020dreamcoder} for program synthesis to draw curves, solving Bongard problems using program induction~\cite{sonwane2021using}, solving Raven's matrices using neuro-symbolic methods~\cite{hersche2022neuro}, and Bongard LOGO~\cite{nie2020bongard} are a few recent and successful approaches towards neuro-algorithmic reasoning, however, their generalization to tasks beyond their domains is often unexplored. 

\noindent\textbf{Visual and language:} tasks for understanding and reasoning on natural images~\cite{radford2021learning,antol2015vqa,hudson2019learning,andreas2016neural,johnson2017clevr} have been very successful using deep neural networks, lately~\cite{radford2021learning,lu2019vilbert,singh2022flava,thrush2022winoground,bitton2022winogavil,kiela2020hatefulmemes,xie2018snlive,suhr2019nlvr2,li2021adversarial,sheng2021human,zellers2019vcr,singh2019textvqa}. Similar to such tasks, our goal in \name is to jointly interpret vision and language modalities for solving various reasoning tasks. However, different from such approaches, our images are not necessarily natural images, instead are mostly sketches without textures; thereby avoiding the unexpected and implicit inductive biases. 

\noindent\textbf{Understanding children's cognition:} for solving a variety of age-appropriate problems has been intensively studied over the years~\cite{cook2011science,lake2017building,gweon2010infants} via studying their ability to form abstract, hierarchical representations of the world, acquire language and develop a theory of mind~\cite{gopnik2004theory}. A particularly useful and common approach to understanding children's cognitive abilities, albeit imperfectly, is to present them with puzzles such as those in IQ tests~\cite{triona2007new,myers2012graphic,lehrer2012supporting}. To the best of our knowledge, it is the first time that a dataset has been built in this direction, that can allow exploration of generalized reasoning abilities at a level of children's cognition, and that can be potentially useful not only in computer vision, but also for studying a breadth of abilities spanning psychology, neuroscience, and cognitive science.

\section{Proposed approach}
\subsection{Task and the \name dataset}
As alluded to above, our goal is to understand the abilities and shortcomings of SOTA deep models for visuo-linguistic reasoning. With this goal in mind, we propose the \fullname and the \name dataset, consisting of visuo-linguistic puzzles in a multiple-choice answer selection setting. 

Each puzzle in \name consists of an image $\img$, a natural language question $\quest$, and a set of five multiple choice answers $\ansset$, and the task is to have an AI model $\model_\theta$, parameterized by $\theta$, that can provide the correct answer $\ans$ to a given problem tuple $\tuple{\img, \quest, \ansset}$, \ie,
\begin{equation}
    \model_\theta(\img, \quest) \rightarrow \ans\in\ansset.
\end{equation}
To learn the parameters $\theta$ of the model $\model_\theta$, we use a dataset $\rootdataset=\set{\rp_1, \rp_2, \cdots, \rp_K}$ consisting of a set of $K=101$ distinct puzzles. We call each $\rp$ a \emph{root puzzle}. To train deep learning models, we need large datasets, and to this end, we create new non-identical puzzle instances for each root puzzle. That is, for each $\rp\in\rootdataset$, we programmatically produce $\puzzset=\set{\puzinst_1, \puzinst_2, \cdots, \puzinst_{n_{\rp}}}$, where $\puzinst$ denotes a new instance of root puzzle $\rp$. Thus, our full dataset $\dataset=\cup_{\pi\in\rootdataset} \puzzset$. 

To choose the root puzzles, one may consider a variety of sources, \eg, puzzle books, IQ tests, online resources, \etc ~In this work, we derived them from the Math Kangaroo (MK) USA Olympiad~\cite{kangaroo}, which is an annually held mathematical competition meant for kids from first to tenth grade. For this paper, we selected problems designed for children of ages 6--8 (typically first and second graders). Given that MK is  a professionally-held competition, it contains high quality content with  significant diversity in children's skills needed for solving the puzzles and offer careful categorization of the algorithmic complexity/difficulty needed for solving them. Table~\ref{tab:puzzle_samples} shows some example root puzzles from our \name. Further, and most importantly, the puzzles being part of a competition, helps gather statistically significant scores on children's performances, which is perhaps difficult to obtain otherwise.
\input{example_puzzles}

\subsection{Programmatic puzzle augmentation}
In this subsection, we detail our approach to replicate a root puzzle into its diverse instances; potentially expanding the dataset to a size that is large enough for adequately training deep neural networks. While, one may resort to standard data augmentation methods (such as cropping, rotations, \etc) to produce data from the root puzzles, such an approach may be unsuitable, because: (i) such operations may make the problem invalid, \eg, flipping an image to augment it might make a question on the orientation of an object incorrect, and (ii) such augmentations might not change the puzzle content much, \eg, rotating an image of a circle. A different direction is perhaps to create more puzzles via human help, \eg, Amazon Turkers. However, this will need specialized creative skills that could be difficult to obtain and can be expensive. 

Intuitively, as we are seeking a model to learn an underlying \emph{algorithm} for solving the puzzles, we should consider puzzle augmentations that make a model algorithmically-equivariant to their solutions. Inspired by this insight, we propose to programmatically augment the puzzles via remaking a root puzzle using a computer program and randomly changing the program settings to diversify the puzzles. Specifically, as our goal is for a reasoning method to learn an ``\emph{algorithm}'' to solve a puzzle (rather than using only the perception modules), we randomly change the visual, lingual, and contextual puzzle attributes using content from a variety of domains, thereby bringing in significant diversity in each recreated puzzle instance. To accomplish this, the new puzzle images are sampled from varied sources, \eg, the Icons-50 dataset~\cite{hendrycks2018benchmarking}, random internet cliparts, \etc, and their spatial organizations, colors, textures, shapes, etc. are all randomly-sampled.  

While the above approach for puzzle augmentation seems straightforward, it needs to be noted that to replicate each root puzzle, sometimes special expertise is needed to produce suitable images, the associated questions, and produce answers that are correct. To illustrate this intricacy, in Table~\ref{tab:puzzle_samples}, we illustrate three puzzles and their augmentations using our approach. Below, we provide details of their augmentation programs.

\noindent\textbf{Table~\ref{tab:puzzle_samples}~Row 1.} We first randomly sample two different types of shapes $s_1$ and $s_2$ from a shape set, with random spatial locations and sizes. Optionally, we also including distractors. Second, we randomly sample the flower instances from the Icons-50 dataset~\cite{hendrycks2018benchmarking} and paste them to the images such that the boundaries of $s_1$ and $s_2$  do not intersect with those of the icons. Third, we randomly sample the relationship associated with $s_1$ and $s_2$ from $\{$inside, outside$\}$ to create the question and compute the answer.

\noindent\textbf{Table~\ref{tab:puzzle_samples}~Row 2.} For a problem setting with $n$ circles (and roads), the replication of this puzzle amounts to finding an $X=\begin{bmatrix}X_{11}, X_{12}\\X_{21}, X_{22}\end{bmatrix}$, where $X_{11}=X_{22}$ and $X_{12}=X_{21}$ with $X_{ij}$'s being $n\times n$ integer matrices under the constraint that their rows and columns sum to $k$ (the number of houses in the puzzle). This problem is cast as an integer programming problem and solved using the GLPK toolkit~\cite{glpk} for random puzzle attributes. 

\noindent\textbf{Table~\ref{tab:puzzle_samples}~Row 3.}
We sample the number of nodes $N$ from  $\left[4, N_{max}\right]$, and sample random graphs with number of edges in $\left[N, \frac{N(N - 1)}{2}\right]$. We use  the NetworkX Python package \cite{networkx,SciPyProceedings_11} for rendering random graphs, post which we randomly sample source and target nodes to generate a question. Next, we find all simple paths between the vertices, compute their lengths, and choose one  target path in the generated question to form the correct answer.
\begin{figure}
\centering
\includegraphics[width=8.3cm,trim={0cm 2cm 1cm 2cm},clip]{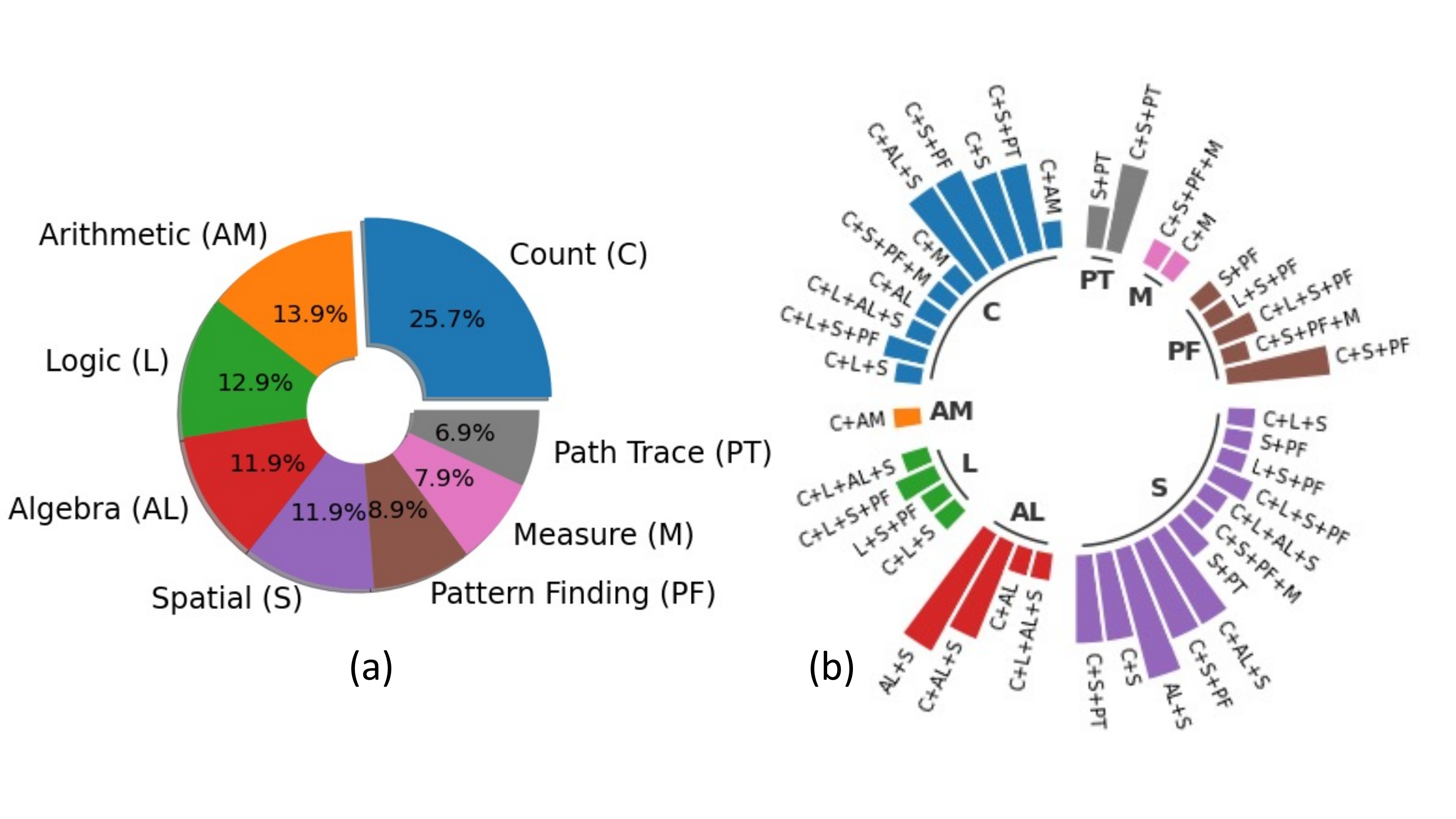}
\vspace{-6mm}
\caption{Analysis of the various statistics of problems in the \name dataset. (a) shows the distribution of problems among the eight classes of predominant math skills needed to solve them. In (b), we plot the composition of various skills that are potentially needed to solve a problem.}
\vspace{-2mm}
\label{fig:app_stats}
\end{figure}

\subsection{Details of the dataset}
We categorize the 101 root puzzles in the \name dataset into eight different classes based on the type of basic skill needed to solve them, namely: (i) variants of counting (\eg, counting lines, basic shapes, or object instances), (ii) basic arithmetic (\eg, simple multiplication), (iii) logical reasoning (\eg, \emph{Is X taller than Y but shorter than Z?}), (iv) algebra (\eg, \emph{Is the sum of the sides of a cube X?}), (v) spatial reasoning (\eg, \emph{Is X behind Y?}), (vi) pattern finding (\eg, \emph{If the pattern in X is repeated, which point will it pass through?}), (vii) path finding (\eg, \emph{which option needs to be blocked so that X will not reach Y in a maze?}), and (viii) measurement (\eg, \emph{for a grid $X$ if each cell is 1 cm, how long is $X$?}). In Figure~\ref{fig:stats}, we show the distribution of puzzles in \name across these classes.\footnote{Note that this categorization was done among the authors via a manual categorization and voting on the root puzzles.}

As one can see from the sample puzzles provided in Table~\ref{tab:puzzle_samples}, it is not just the above skills that one needs to solve them, instead their solution demands a composition of the above skills. For example, to solve the puzzle in the first row of Table~\ref{tab:puzzle_samples}, one needs to recognize the \emph{pattern} for similar flowers, \emph{spatially reason} whether each flower is within or outside a given shape, and \emph{count} the flowers. The class distribution in Figure~\ref{fig:stats}(a) characterizes the basic skill needed (\eg, counting) to solve this problem, and might not provide the full skill diversity. Thus, in Figure~\ref{fig:stats}(b), we provide a more comprehensive analysis of the various compositions of skills needed to solve the of problems in \name. As is clear from this pie chart, each puzzle in our dataset demands a multitude of skills -- attesting to the complexity of the task and the challenge it offers. 

\noindent\textbf{Question Augmentation.} To create new questions for puzzle instances, we follow a combination of three different strategies: (i) for puzzle questions with numbers, we replace them with new numbers sampled from a range, (ii) replace the sentence structure with manually-generated templates, and (iii) use slotted words in the template, where the words in the slots are sampled from potential synonyms, while ensuring the question is grammatically correct, sensible, and captures the original goal and difficulty of the puzzle.

\section{\name reasoning model}

\begin{figure}
    \centering
    \includegraphics[width=0.9\linewidth,trim={3.2cm 3cm 6.8cm 4.0cm}, clip]{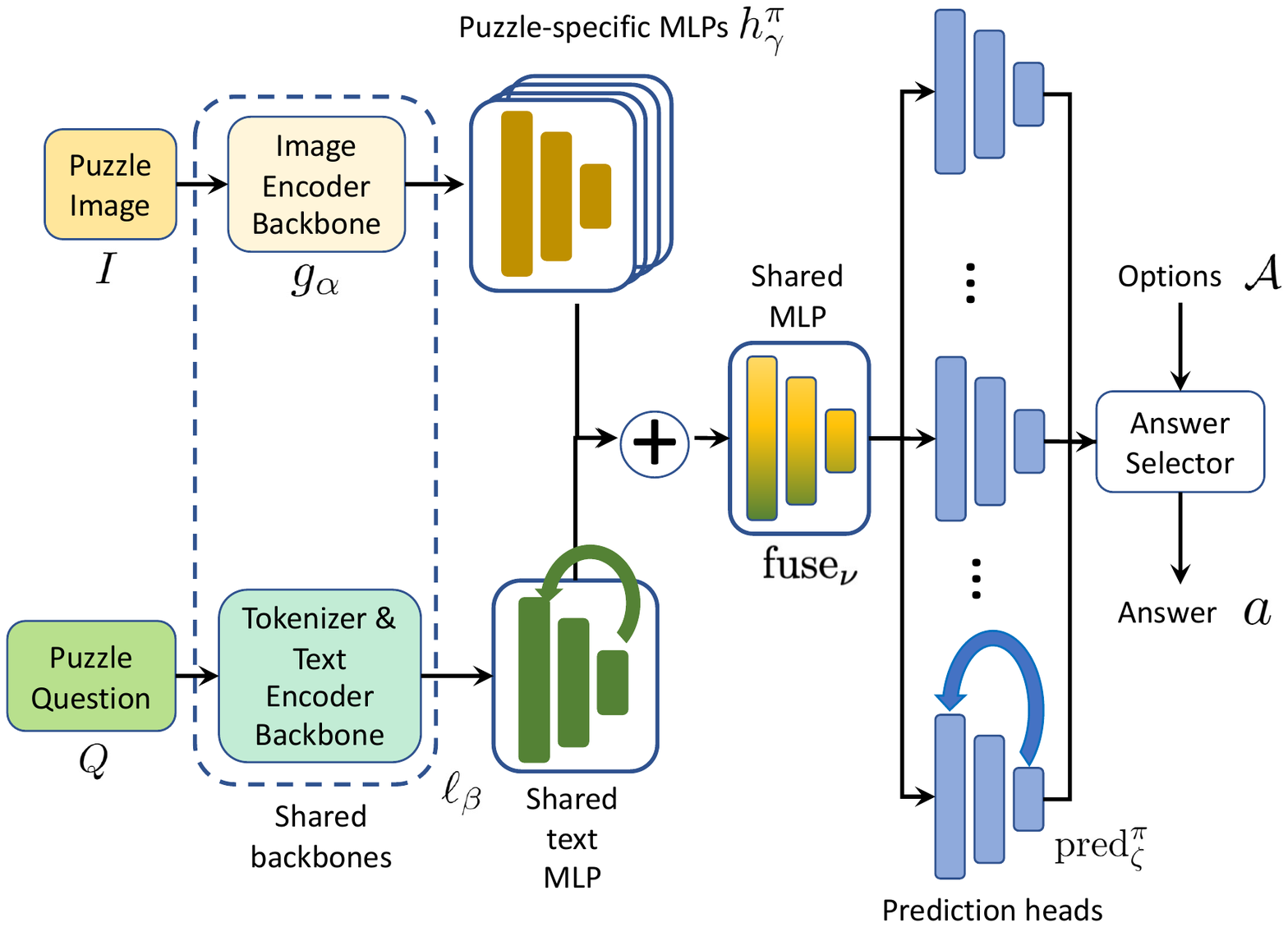}
    \vspace{-4mm}
    \caption{An illustration of our learning and reasoning model.}
    \vspace{-0.6cm}
    \label{fig:arch}
\end{figure}
Each puzzle in \name has distinct problem characteristics and diverse ranges for their answers (\eg, numeric, alphabets, sentences, and words); thus, using a single loss for all puzzles may be sub-optimal. While, one may resort to multi-task learning (MTL), however having samples from all puzzles to train in MTL may need large batches that can be difficult to scale. Further, we desire our model to be trained and evaluated in few-shot settings. A natural way to resolve all the above challenges is to consider a meta-learning architecture~\cite{finn2017model}, pictorially described in Figure~\ref{fig:arch}. 

Mathematically, let $\imagehead$ and $\lanhead$ be the \emph{image backbone} and the \emph{language backbone} (combined with an RNN to aggregate the word embeddings) shared across all the puzzles in $\dataset$ respectively, where $\alpha$ and $\beta$ capture their parameters. As distinct root puzzle images have specific characteristics for the solution (e.g., some of the images have their answer options embedded within the image), we found it useful to have a puzzle-specific image head. To this end, we attach a small (2-layered) multi-layer perceptron (MLP), denoted $\imgpuzzhead$, to the output of the image backbone, where $\imgpuzzhead$ is specific to each root puzzle $\pi$ and has its own parameters $\gamma$. Using these modules,  our prediction model for puzzle $\pi$ is:
\begin{equation}
    \model^{\pi}_\theta(\img, \quest) :=  \pred(\fusion((\imgpuzzhead(\imagehead(\img)) + \lanhead(\quest))),
    \label{eq:model_eq}
\end{equation}
where $\fusion$ denotes a shared MLP to fuse the image and language embeddings and $\pred$ is a puzzle-specific prediction head that maps a given puzzle tuple to the domain of the puzzle answers (with its own parameters).  For example, a puzzle answer may be a sequence, for which $\pred$ would be an RNN, while for another puzzle, the response could be an integer in 1--100, for which $\pred$ could be an MLP classifier with 100 softmax outputs. We abstractly represent trainable parameters in various modules by $\theta$. 

To train the model in Eq.~\eqref{eq:model_eq}, we optimize:
\begin{equation}
    \min_{\Theta} \expect_{\pi\sim\rootdataset} \expect_{\tuple{\img, \quest, \ans}\sim\puzzset} \loss_{\pi}\bigl(\model^{\pi}_\theta(\img, \quest) - \ans\bigr),
    \label{eq:loss}
\end{equation}
where $\Theta=\cup_{\pi\in\rootdataset} \set{\theta}_{\pi}$ and $\loss_\pi$ is a puzzle-specific loss that is activated based on the root puzzle $\pi$ for an instance $\tuple{\img, \quest, \ans}$ in a given batch. Specifically, we sample the tasks (puzzles) and instances from those tasks to form mini-batches to train the puzzle-specific heads for several iterations, followed by combining the gradients from the tasks to update the backbones through the puzzle heads, as in~\cite{finn2017model}. Note that $\ans$ is the correct answer and $\loss_\pi$ could be: (i) a softmax cross-entropy loss (selecting in a discrete answer range) or (ii) an $\ell_1$ regression loss predicting a scalar value.

At inference, we select the answer from the options as:
\begin{equation}
    \hat{\ans} = \argmax_{\alpha\in\ansset} \similarity\bigl(\model^\pi_\theta(\img, \quest), \alpha\bigr),
\end{equation}
where $\similarity$ captures the similarity of a predicted answer value against the choices in $\ansset$, and $\similarity$ is specific to the problem $\rp$ (\eg, euclidean distance for numerals).

\section{Experiments}
In this section, we detail the experimental protocol to evaluate the models for solving \name. 
\subsection{Data splits}
We propose four different data splits that evaluate varied generalization properties of a method/model to solve \name. The splits are: (i) \textbf{Puzzle Split} (PS) with the goal to evaluate extreme generalization. In this setting, we split the root puzzles into 77-3-21 (train-val-test).\footnote{In PS test, we use 2 counting, 5 logic, 4 algebra, 1 path, 1 measurement, 4 spatial, 3 arithmetic, and 1 pattern puzzles.} The performance is evaluated on the test set consisting of puzzles that the model has never seen during training (as a zero-shot solver). (ii) As PS is perhaps extremely challenging for today's machine learning approaches, we include a \textbf{Few-shot Split} (FS), where the model sees $m~(=10)$ instances from all the 21 puzzles used as the test set in PS. (iii) \textbf{Instance Split} (IS) evaluates the in-distribution performance of a model (supervised learning). For IS, we split all the instances of all root puzzles into 80-5-15 (\%). IS receives puzzle-specific information on all puzzles and is the easiest setting for a model to perform. (iv) \textbf{Answer Split} (AS) that evaluates the generalization to answers that a model has not seen during training. In this split, we compute the distribution of all answers ($\ans$ in Eq.~\ref{eq:loss}) across instances for a root puzzle, find the median answer, and remove all instances that have this median answer from the training set; these instances are used only during inference. See Appendix~\ref{sec:data_splits} for more details on the data splits.

\subsection{Evaluation}
We use two metrics to evaluate performance: (i) the solution accuracy $\sacc$ that computes the frequency with which the correct \emph{solution} was produced by a model and (ii) the option selection accuracy $\oacc$ that measures the frequency with which the correct \emph{option} was selected by a model. To clarify, for the root puzzle in Table~\ref{tab:puzzle_samples} Row 1, let us say a model produced an answer 8. Since 8 is not in the option set, the closest option 9 will be selected, \ie, the correct option will be selected even if the wrong answer is produced. In this case, its $\oacc$=100\%. while its $\sacc$=0\%. 

\subsection{Backbone models}
We evaluate popular pretrained image, language, and vision-and-language backbones\footnote{All the pretrained backbone models are downloaded from public repositories, specifically \url{https://huggingface.co/models}.} using the reasoning architecture in Figure~\ref{fig:arch}.

\noindent\textbf{Image Backbones.} We consider three groups of models: (i) ResNets, (ii) Transformers, and (iii) contrastively pre-trained models. For (i), we use ResNet-50 and ResNet-18~\cite{He2015}. For (ii) we use several variants, including Vision-Transformers (ViT)~\cite{dosovitskiy2020image}, Swin-Transformers~\cite{liu2021swin} (Swin-T and Swin-B) and Cross-Transformers~\cite{wang2021crossformer}. While we fine-tune ViT and Swin-Ts from pre-trained models, we train Cross-Transformers from scratch on our dataset. For self-supervised pre-trained models, we use SimSiam~\cite{chen2021exploring} based on ResNet-50 and Masked Autoencoders ~\cite{he2022masked}(MAE). 

\noindent\textbf{Language Backbones.} As alluded to above, we use either a learned feature embedding (Emb.) for encoding the questions (using a vocabulary of $\sim$7K words created on \name) or a SOTA embedding model and its associated tokenizer. We consider 3 text embedding models: (i) GPT-2~\cite{radford2019language}, (ii) BERT~\cite{devlin2018bert}, and (iii) GloVe~\cite{pennington2014glove}.  

\noindent\textbf{Vision-and-Language Models.} We also consider multimodal pre-trained models that are specifically trained for aligning vision with language. In this setting, we consider the recent CLIP~\cite{radford2021clip} and FLAVA~\cite{singh2022flava}. 

\subsection{Implementation details and data preprocessing}
\label{sec:impl}
\noindent\textbf{Training setup.}
For ease of implementation and benchmarking, we use the same training settings for all our (backbone) models. Specifically, we use a batch size of 64, and train/fine-tune the models using the Adam optimizer with a learning rate of 0.001. The maximum question length is set to 110 tokens. Some of the vision-and-language models (such as CLIP and BERT) have a maximum token limit of 77, to which we trim the corresponding questions accordingly so that no important information relevant to the puzzles are removed. 

We use a 2-layer MLP (with ReLU activations) for our (puzzle and instance) heads with an output dimensionality of the image and language MLPs set to 128. The language features from the MLP are aggregated using a single layer GRU. For the sequential prediction head (for generating sequential answers), we use another single layer GRU. When using a single prediction classifier head for all the puzzles (e.g., in PS and FS splits), we use a maximum output dimensionality of 256 (for the softmax function), which is the maximum of the correct answer values that we can have across all puzzles. All the models are trained for 100 epochs or until their validation performances saturate. 

\noindent\textbf{Training losses.}
As the images across root puzzles may be significantly different and need varied image analysis skills, we use a puzzle specific image head as shown in Figure~\ref{fig:arch}. The models are trained end-to-end using a loss computed either using: (i) a softmax cross-entropy loss, or (ii) an $\ell_1$ regression loss, or (iii) softmax cross-entropy loss between the predicted and ground truth answers for the entities in a sequential answer (we have 6 puzzles in \name that has sequential answers). For (i), we use the output dimensionality as the discrete ranges of the answers in the training set (e.g., if the maximum value of $\ans$ over all instances of a puzzle is $\nu$, then we use $\nu$ as the output dimensionality of the corresponding puzzle head), and use cross-entropy loss to train it against the ground truth answer.\footnote{Here, the ground truth answer is not the 5 candidate answer options, instead is the value of the actual answer to the puzzle instance} Note that the output dimensionality varies from puzzle-to-puzzle and some puzzles use GRU to produce sequential answers. For (ii), we use an $\ell_1$ loss between the correct answer and the predicted scalar real value, and in this case we use a single scalar output as the model prediction (or a sequence of scalars for sequential outputs).

\noindent\textbf{Preprocessing.}
The puzzle images in \name have varied resolutions depending on the puzzles. For example, some puzzle images wide and thinner, while many have an equal aspect ratio. As most backbones use $224\times 224$ resolution for their inputs, for efficiency, we resize all the original high resolution puzzle images to this size. Next, for the specific type of backbone used for training (e.g. Transformers, ResNets, etc.), we apply the recommended image normalization (e.g., mean subtraction and such) that comes along with the respective pre-trained model (e.g., PyTorch models on the hub provides such pre-processing steps as part of their models). For the question embedding stream, we tokenize the words either using a tokenizer that comes with a pre-trained model (e.g., CLIP~\cite{radford2019language} or FLAVA~\cite{singh2022flava}) or use embeddings based on a dictionary constructed across all the puzzle instances in our dataset. After filtering for less frequent words (using a frequency threshold of 3), our language model uses a vocabulary of size 7037 words. To train the models, we convert all the options into integers, \eg, using the ordinals for alphabets.  

\begin{table*}[ht]
    \centering
    \resizebox{\textwidth}{!}{
    \begin{tabular}{cccccccccc}
    \toprule[1pt]
        Puzzle Category~$\rightarrow$ & Count & Arithmetic & Logic & Path Trace & Algebra  & Measure & Spatial & Pattern Finding & Average\\
        \midrule 
        \multicolumn{10}{c}{Puzzle Split (PS) -- Extreme Generalization Experiments} \\
        \midrule
        Avg. $2^{nd}$ Grader Performance &  \textbf{72.8}&	\textbf{81.3}& \textbf{82.2} &	\textbf{81.1} &	\textbf{64.5}&\textbf{90.4} &	\textbf{74.8} &  \textbf{88.6} & \textbf{77.1} \\
        \hline
        Greedy (baseline) &19.1/21.4 &  14.0/21.4 &  18.5/21.1 &  21.8/21.1 &  13.5/21.5 &  23.1/20.9 &  18.2/21.2 &  21.4/21.4 & 17.7/21.3\\ 
        Uniform (baseline) &7.74/20.0 &  8.00/20.0 &  7.61/20.0 &  18.9/20.0 &  6.94/20.0 &  5.62/20.0 &  14.2/20.0 &  20.0/20.0 & 11.20/20.0\\
        \hline
        MAE + BERT & 7.2/12.0 &  3.3/23.1 &  10.4/34.1 &  9.6/22.0 &  7.3/14.7 &  3.7/15.2 &  8.5/16.5 &  2.6/16.4 & 7.21/19.1\\
        SimSiam + BERT & 6.4/18.4 &  4.8/20.9 &  7.7/41.4 &  2.5/22.2 &  4.2/25.3 &  7.9/20.5 &  11.8/22.2 &  0.2/17.2 & 6.41/23.9\\
        \midrule
        Swin\_T + BERT & 810.5/17.3 &  4.7/24.7 &  5.6/29.3 &  11.4/21.5 &  6.5/16.8 &  10.3/23.3 &  11.9/16.3 &  17.3/19.1 & 9.25/20.1\\ 
        ViT-16 + BERT & 9.41/22.7 & 5.77/26.8 & 6.95/25.1 & 4.72/18.7 & 5.57/15.1 & 8.68/21.3 & 11.6/21.5 & 18.9/19.7 & 8.51/21.6\\
        \midrule
        CLIP &  9.1/15.7 &  1.4/18.5 &  7.4/30.6 &  14.2/21.4 &  7.5/18.6 &  8.9/22.2 &  12.4/18.4 &  19.0/19.6 & 11.9/24.1\\
        FLAVA & 8.3/20.2 & 4.0/22.2 & 8.1/31.3 & 9.5/20.3 & 3.1/22.2 & 19.0/32.0 & 9.7/18.1 & 20.9/21.2 & 7.21/19.0\\
        \midrule
        R50 + BERT (FT + Cls.) & 10.9/18.3 & 6.96/15.8 & 12.8/20.8 & 19.6/19.7 & 7.95/15.1 & 16.9/26.7 & 13.4/17.7 & 0.0/21.2 & 11.7/18.9\\
        R50 + BERT (FT + Reg.) & 12.0/22.8 & 5.08/21.3 & 4.24/16.2 & 18.4/18.4 & 4.89/22.2 & 15.1/25.9 & 11.9/17.9 & 19.0/19.0 & 8.21/19.7\\
        \midrule
        \multicolumn{10}{c}{Few-Shot Split (FS) Experiments, $m=10$} \\
        \midrule
        R50 + BERT (Cls.) &  17.3/28.0 &  11.2/25.8 &  18.0/37.6 &  19.2/19.2 &  7.9/21.9 &  14.8/31.2 &  18.7/25.8 &  17.8/17.8 & 15.2/25.3\\
        R50 + BERT (Reg.) & 13.3/25.2 & 8.3/24.7 & 11.2/23.3 & 17.3/18.6 & 6.6/18.9 & 19.5/34.2 & 18.5/26.4 & 21.1/21.1 &  13.6/23.3\\
        \midrule
        \multicolumn{10}{c}{Instance Split (IS) -- Supervised Learning Experiments} \\
        \midrule 
        Greedy (baseline) &21.7/22.6& 8.97/21.5 & 18.5/21.0 & 22.7/21.2 & 10.2/21.1 & 12.8/21.1  & 22.3/21.3& 20.6/21.3 & 17.3/21.6\\ 
        Uniform (baseline) & 9.41/20.0 & 3.65/20.0 & 7.91/20.0 & 11.1/20.0 & 5.01/20.0 & 3.63/20.0 & 15.5/20.0 & 16.7/20.0 & 8.41/20.0\\
        \midrule
        Swin-T  + Emb. & 23.1/35.1 & 33.7/41.0 & 20.3/28.8 & 16.7/18.6 & 17.7/29.5 & 26.3/34.3 & 24.5/29.1 & 17.5/26.5 &  22.5/30.8\\ 
        Swin-B + Emb. & 22.0/34.0 &  29.4/36.5 &  17.7/26.1 &  16.7/17.0 &  17.1/30.2 &  25.0/34.2 &  26.2/30.7 &  21.5/29.6 & 21.6/29.9\\
        Cross-Transformer + Emb. & 20.5/30.4 & 6.3/15.3 & 15.5/22.9 & 15.1/15.6 & 8.7/23.9 & 10.7/18.2 & 21.7/24.7 & 19.0/27.3 & 14.7/22.8\\
        ViT-16 + Emb. & 25.6/36.4 & 39.7/47.1 & 21.2/30.8 & 15.5/16.3 & 20.1/33.8 & 39.4/40.8 & 29.0/33.0 & 20.3/29.6 & 25.9/33.5\\
        MAE + Emb. & 25.4/36.7 &  34.2/43.2 &  21.6/31.5 &  16.4/16.7 &  20.0/33.3 &  32.0/39.7 &  28.2/32.9 &  18.6/26.6 &  24.5/33.0\\
        \midrule
         SimSiam + Emb. & 44.9/56.1 & 35.1/43.5 & 45.7/50.8 & 25.0/26.6 & 23.4/35.1 & 64.7/73.5 & 55.0/57.2 & 42.8/49.1 & 39.5/47.0\\
        R18 + Emb. & 44.0/54.0 & 8.8/19.8 & 41.1/47.6 & 24.5/26.7 & 13.7/26.5 & 30.9/40.2 & 43.3/45.5 & 29.5/34.8 & 29.4/37.4\\ 
        \midrule
         R50 + Emb. & 46.6/57.8 & 38.0/45.9 & 43.2/50.1 & 24.6/26.4 & 23.3/35.1 & 56.9/67.4 & 57.9/58.6 & 44.8/51.0 & 39.8/47.5\\
        R50 + GloVe & 46.0/56.3 & 39.2/48.5 & 53.9/56.4 & 26.7/28.9 & 21.5/32.4 & 58.9/68.5 & 48.5/50.4 & 43.3/47.8 &  40.0/47.2\\
        R50 + GPT2 & 47.0/57.9 & 44.8/53.1 & 55.1/58.6 & 26.1/28.4 & 27.2/39.3 & 61.0/71.3 & 49.0/50.2 & 42.5/48.4 &  42.1/49.6\\
        R50 + BERT &  48.5/59.3 &  46.1/54.9 &  56.7/60.2 &  26.5/28.4 &  28.5/39.7 &  65.6/75.4 &  44.3/46.2 &  39.9/45.3 &  \gray{42.8/50.2}\\
        \midrule
        CLIP & 41.3/52.9 & 18.2/29.3 & 33.3/41.1 & 19.8/21.9 & 12.9/24.9 & 27.8/42.8 & 32.2/36.2 & 29.9/36.1 & 27.3/36.4\\
        FLAVA & 47.7/58.1 &  20.2/29.7 &  41.4/47.1 &  25.4/27.1 &  19.6/31.2 &  30.5/41.9 &  33.2/35.7 &  38.3/44.2 & 32.3/40.2\\
        \midrule
        \multicolumn{10}{c}{Answer Split (AS)  -- Answer Generalization Experiments}\\
        \midrule
        R50 + BERT (FT + Cls.) & 0.1/23.8 & 1.5/13.2 & 0.0/16.8 & 0.0/1.6 & 0.4/17.3 & 0.0/21.1 & 0.0/6.0 & 0.0/15.0 & 0.19/10.2\\
        R50 + BERT (FT + Reg.) & 12.0/28.4 & 10.4/25.7 & 19.6/30.8 & 9.5/10.6 & 3.64/18.3 & 9.42/28.6 & 14.1/21.1 & 25.5/30.9 & 16.3/23.4\\
    \bottomrule[1pt]
    \end{tabular}
    }
    \vspace{-2.5mm}
    \caption{\name performances of various image and language backbones in our framework on PS, FS, IS, and AS splits. We also report the second-grader performances. Each entry shows $\sacc/\oacc$ (\%; higher is better). The image backbones in IS are all fine-tuned.}
    \vspace{-3mm}
    \label{tab:category-wise-performance}
\end{table*}

\section{Experimental Results}
In Table~\ref{tab:category-wise-performance}, we present our results using our reasoning framework and varied backbones on both $\sacc$ and $\oacc$ metrics, and against human performance.

\subsection{Second Grader Performance} The main goal of this paper is to gauge the performance of SOTA deep neural networks against those of second-graders. In Table~\ref{tab:category-wise-performance}, we report averaged category-wise performances of children who participated in the Math Kangaroo competition. Overall, children perform at nearly 77\% average accuracy on all the 21 PS puzzles. See Appendices~\ref{sec:mk_details} and~\ref{sec:puzzle_categories} for more details on MK, root puzzle categories, and instance augmentation.

\subsection{Baselines} To ensure that \name answer options are devoid of any biases, we report two baseline performances that do not involve any learning, namely: (i) \emph{greedy}, that selects the most frequent answer from the training set instances for each root puzzle, and (ii) \emph{uniform}, that randomly samples an answer. Table~\ref{tab:category-wise-performance} shows that $\oacc$ for all the baseline methods is nearly 20\%, suggesting that our answer options are uniformly distributed among the five choices.

\subsection{Analysis of Generalization}
The fundamental goal of this paper is to understand the generalization abilities of SOTA deep models. In  Table~\ref{tab:category-wise-performance} (under Puzzle Split), we report results analyzing extreme generalization using Transformers, CLIP, and FLAVA. In these experiments, we used the  publicly available pretrained backbones and trained only puzzle heads. From the table, we find that SOTA models fail entirely, often selecting a random answer ($\oacc$$\sim$20\%). We also evaluate our best setting (R50 + BERT) via fine-tuning (FT) R50 with classification (Cls.) and regression (Reg.) losses; however, without any improvement.

To ameliorate extreme generalization, we explore the few-shot (FS) setting where the model is shown $m$ instances of a puzzle during training that is otherwise hidden in the PS split. Even for an $m=10$, Table~\ref{tab:category-wise-performance} (FS) shows that the $\sacc$ improves by nearly 4\% (from $\sim11$\% in PS to $\sim15$\% in FS), suggesting that the model has perhaps learned several useful embeddings and may learn new skills quickly. 

Next, using R50 + BERT, in classification and regression settings, we evaluate answer generalization (on the AS split). Table~\ref{tab:category-wise-performance} (AS) shows our classification model fails entirely on AS (0\% on $\sacc$). This is unsurprising as on the AS split, the deep model is masked from seeing a particular answer, which is used only during testing. However, Table~\ref{tab:category-wise-performance} (FS) also shows that using regression allows the model to interpolate the seen answers, leading to an $\sacc$ of $16.3\%$.

\subsection{Supervised Learning (IS) Performances} For these experiments, we use the learned word embeddings (Emb.). Surprisingly, we find that in IS, ResNet models (R18/R50/SimSiam) perform significantly better than most Transformer models on average (Table~\ref{tab:category-wise-performance}-IS). To ensure this is not an implementation artifact, we repeat our experiments either via training the models from scratch (Cross-Transformers) or fine-tuning pretrained models (Swin-B, Swin-T, ViT-16, and MAE). These models offer varied amounts of global and local self-attention for reasoning. Table~\ref{tab:category-wise-performance} shows that most Transformer variants we compare to do relatively well in \emph{Arithmetic} ($\sim$40\% on $\sacc$ for ViT-16, $\sim$34\% for Swin-T and MAE, \etc), while they perform the least on tasks that need path tracing. We find that prtrained vision-and-language models (FLAVA and CLIP) perform slightly better than Transformers and show improved performances on counting, logic, and pattern finding. Using R50 image backbone, we further evaluate the performances against various language model choices. We find in Table~\ref{tab:category-wise-performance}-IS that richer (pretrained) language models such as GloVe, GPT2 or BERT improve the performance over Emb., with benefits in almost all puzzle categories.

\section{Detailed Experiments}
\subsection{Comparisons to Human Performance (IS split)}
 As alluded to above, the human study was conducted using the data provided to us by the organizers of MK, and included performance statistics from MK2020 and MK2021. Extending upon the human performance analysis provided in the Table~\ref{tab:category-wise-performance}, we provide more analysis here. As described earlier, we use only some puzzles (of appropriate difficulty level) in \name, and thus we evaluated the deep learning performance only on those. The MK statistic provided to us includes the number of participant responses received for each puzzle and how many of them were correct, which is then used to compute the correct response rate, which we call \emph{accuracy} in our bar plots (provided below). 

\noindent\textbf{Are the comparisons fair?} For the following comparisons between human and AI, we used a ResNet 50 backbone with learned embeddings (R50 + Emb.), which is one of our best performing models as seen in Table~\ref{tab:category-wise-performance}, and trained using the instance-split (IS). Note that in contrast to students taking the MK test, who have never seen the exact puzzles they are solving, the deep learning method is trained on nearly 1600 instances of the root puzzles; while this may appear an unfair comparison, note that without this training, deep methods fail entirely (nearly as seen in Table~\ref{tab:category-wise-performance} and their performances are close to random accuracy. Further, we apply the deep learning method on 300 test instances to aggregate their performances, as against one test example given to the participants in MK. While, these differences could be characterized as not precisely measuring the human-vs-machine performances, note that our overarching question is to see if an AI model can have the skill set that children have, even when the AI model has been trained on instances of the puzzles solving which explicitly demands learning such skills and provide a plausible best case on the performance against humans.  

\begin{figure*}[ht]
    \centering
    \begin{subfigure}{0.49\linewidth}
        \includegraphics[width=8cm]{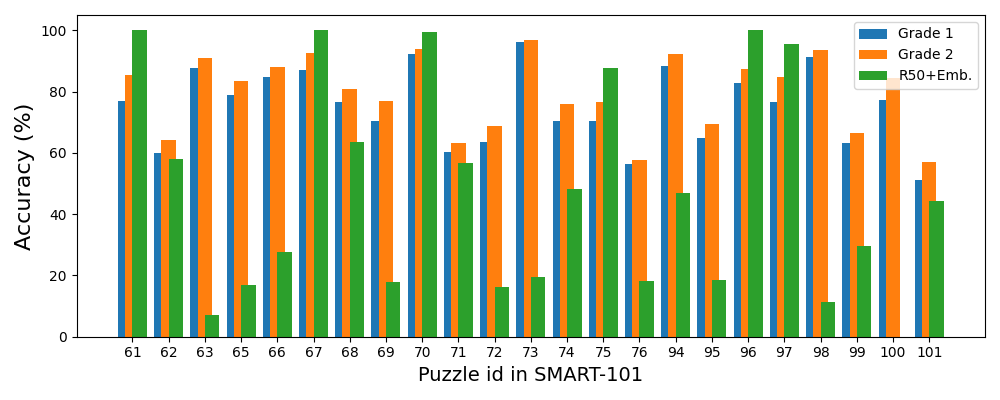}
        \caption{Overall }
        \label{fig:hp_all}
    \end{subfigure}
    \begin{subfigure}{0.49\linewidth}
        \includegraphics[width=8cm]{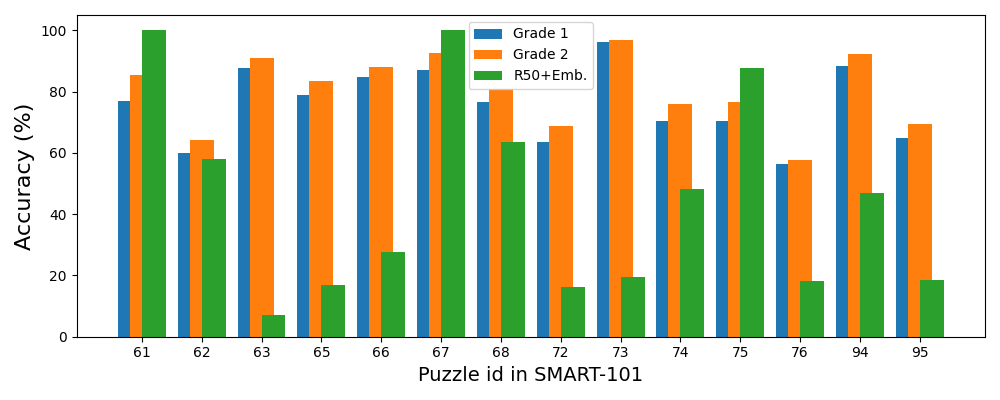}
        \caption{ One-step puzzles}
        \label{fig:hp_diff_one}
    \end{subfigure}\\
    \begin{subfigure}{0.49\linewidth}
        \includegraphics[width=8cm]{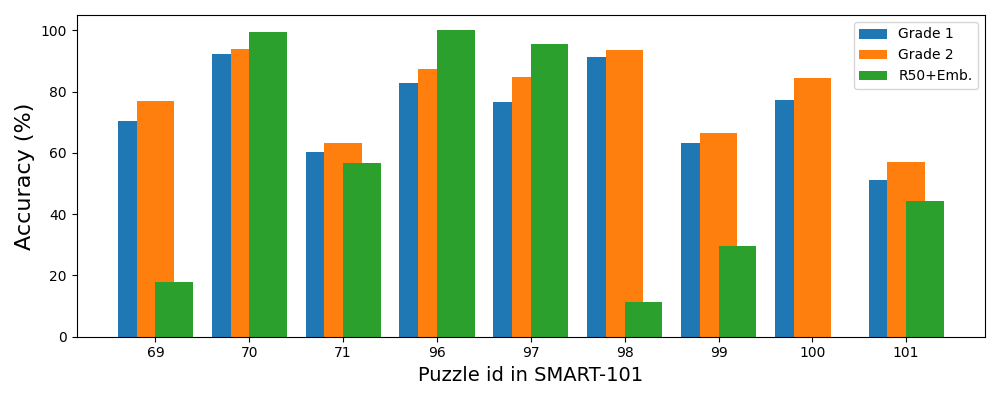}
        \caption{Two-three step puzzles}
        \label{fig:hp_diff_two}
    \end{subfigure}
    \begin{subfigure}{0.49\linewidth}
        \includegraphics[width=8cm]{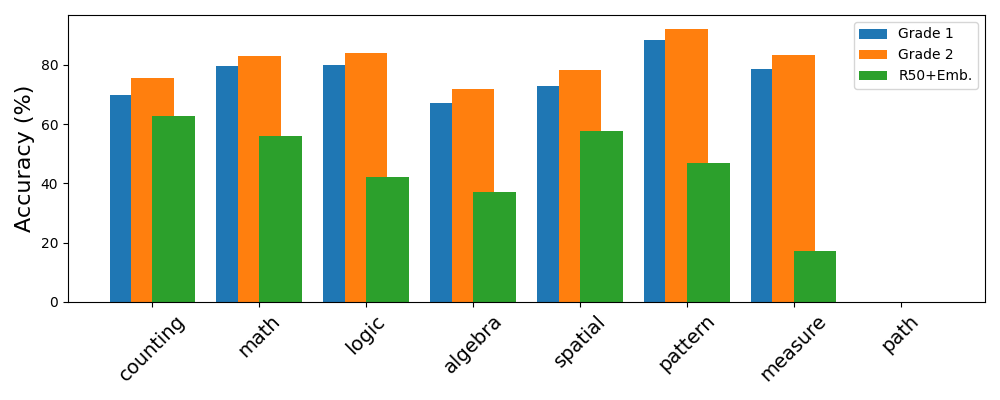}
        \caption{Average performance over skill set.}
        \label{fig:hp_diff_cat}
    \end{subfigure}
    \caption{Comparison of ResNet-50 + Emb. (IS split) performance against that of participants (grades 1--2) in the MK competition held in 2020 and 2021. For (d), we have not included root puzzles needing skills such as 'path finding', as they were absent from the subset of puzzles we selected for \name from that particular year of MK.}
    \label{fig:human-AI-stats}
\end{figure*}

\subsubsection{Analysis against human performances} In Figure~\ref{fig:human-AI-stats}, we compare the human performance against our best-performing AI model (on IS). As can be seen from the bar-plots, our AI model demonstrates better results on 6 out of 23 problems we considered. Among single-step puzzles (Figure~\ref{fig:hp_diff_one}), the AI model is better than humans on five and one on two/three-step (difficult) puzzles (Figure~\ref{fig:hp_diff_two}). On one puzzle, viz. puzzle 100 in Figure~\ref{fig:hp_diff_two}, the AI model fails entirely (0\% accuracy). This is perhaps expected, given this is a sequential puzzle and needs to solve a cryptography problem. In general, we found that the deep model struggles in solving sequential puzzles which need all the items in the sequence to match exactly with the ground truth. In Figure~\ref{fig:hp_diff_one}, puzzles such as 61, 73 (algebra), 62 (path finding), 63 (logic), 66 (spatial), and 95 (arithmetic), the model performs very poorly as well. However, on 64 (arithmetic), 67 (logic), 93 (spatial), 94, 99 (counting), it performs nearly 100\%.  On the difficult puzzles (Figure~\ref{fig:hp_diff_two}), the model performs well on 68 (arithmetic), 98, 99 (algebra), 101 (logic). In Figure~\ref{fig:hp_diff_cat}, we provide a skill set based categorization of the performances. Overall, we see the model performs reasonably well on counting, arithmetic, spatial reasoning, and pattern matching, however struggles on logic, algebra, measuring. Note that we only had a single puzzle for ``measure'' class, and did not have any for the ``path finding'' category in the root puzzle sets we used from MK2020 and MK2021.\footnote{A very careful reader might wonder how we had the performances for all the eight skill categories in Table~\ref{tab:category-wise-performance}. This is using data from elsewhere in MK where the test setting is slightly different from an exam setting, instead is a practice mode.}

\subsubsection{Pearson Correlation Analysis} Additionally, we compute the Pearson correlation coefficient between MK grade 1 and grade 2 participant performances, given by the Correct Response Rate (CRR), and the performance we obtained using the best performing deep model using the instance-split. We do this for the puzzles by both combining puzzles of both difficulty levels -- one-step puzzles and two/three-step puzzles -- together as well as separately. The results are shown in Table \ref{tab:correlation}. We observe that the correlations between human performance and both $O_{acc}$ and $S_{acc}$  performance, while slightly on the lower side, appear to be positively correlated with the model performance. Interestingly, the correlation is higher for one-step puzzles (nearly 0.3) and lower for two/three step puzzles (around 0.1-0.2), suggesting that perhaps there is a discrepancy between the human judgement of puzzle difficulty against what the AI model finds challenging to learn. We find that grade 2 participants achieve mean CRRs of $79.59\%$ overall, $84.68\%$ for one-step puzzles, and $71.68\%$ for two/three-step puzzles. The corresponding results for Grade 1 students are $75.09\%$, $80.15\%$, and $67.22\%$, respectively. We see similar trends when using the best performing deep model on the IS split which yields $S_{acc}$ of $39.75\%$, $43.41\%$ and $34.07\%$ respectively. 

\begin{table*}[ht]
    \centering
    \resizebox{0.7\textwidth}{!}{
    \begin{tabular}{c|ccc|ccc}
    \toprule[1pt]
      & \multicolumn{3}{c|}{$O_{acc}$} & \multicolumn{3}{c}{$S_{acc}$} \\
         \hline
        & Overall & One-step & Two/Three-step & Overall & One-step & Two/Three-step \\
        \midrule
        Grade 2 CRR & 0.17 & 0.31 & 0.19 & 0.33 & 0.33 & 0.30 \\
        Grade 1 CRR & 0.12 & 0.28 & 0.06 & 0.29 & 0.30 & 0.22 \\
    \bottomrule[1pt]
    \end{tabular}
    }
    \caption{Pearson correlation coefficients between Grade 1 and Grade 2 Correct Response Rate (CRR) and $\oacc$ and $\sacc$ for all puzzles in Fig. \ref{fig:hp_all} (Overall), one-step puzzles listed in Fig. \ref{fig:hp_diff_one} and two/three-step puzzles listed in Fig. \ref{fig:hp_diff_two}}
     \label{tab:correlation}
    \vspace{-3mm}
\end{table*}

\begin{table}[ht]
    \centering
    \resizebox{0.7\columnwidth}{!}{
    \begin{tabular}{ccc}
    \toprule[1pt]
    Method & $\sacc\uparrow$ & $\oacc\uparrow$ \\ 
    \midrule 
     \multicolumn{3}{c}{Instance split}\\
    \midrule
        R50 + BERT &  \textbf{42.8} & \textbf{50.2} \\ 
    \midrule
        No meta learning/MTL & 29.7 & 37.3 \\
        Image only (no question) &  28.3 & 36.3 \\
        Question only (no image) &  15.1 & 23.2 \\
        Single image head & 25.0 & 34.3 \\
    \bottomrule
    \end{tabular}
    }
    \vspace{-2.5mm}
    \caption{Ablation studies using the R50 + BERT model.}
    \vspace{-3.5mm}
    \label{tab:ablation}
\end{table}

\begin{table*}[ht]
    \centering
    \resizebox{\textwidth}{!}{
    \begin{tabular}{ccccccccccc}
    \toprule[1pt]
        Model  & Cls./Reg. & Count & Arithmetic & Logic & Path Trace & Algebra  & Measure & Spatial & Pattern Finding & Average\\
        \midrule 
        Meta Learn & Cls. & 23.9/37.3 & 32.7/41.2 & 32.7/40.7 & 22.1/22.2 & 10.2/27.5 & 17.4/32.8 & 28.4/35.3 & 33.9/33.9 & \textbf{24.4/33.4}\\
        MTL  & Cls. & 26.1/42.6 &  14.9/25.5 &  27.2/40.2 &  19.6/19.6 &  12.8/31.3 &  23.2/34.4 &  24.0/29.5 &  21.4/21.4 & 20.3/29.9\\
        \hline
        Meta learn & Reg. & 19.9/33.7 & 13.9/26.3 & 18.3/26.9 & 18.7/18.7 & 10.3/24.4 & 11.6/25.8 & 20.8/29.8 & 21.9/22.3 & \textbf{16.7/26.5}\\
        MTL  & Reg. & 17.7/28.6 & 11.2/24.8 & 9.5/14.2 & 18.8/18.8 & 7.3/15.5 & 17.9/31.9 & 17.1/24.4 & 21.1/21.1 & 14.2/21.6\\
    \bottomrule[1pt]
    \end{tabular}
    }
    \vspace{-2.5mm}
    \caption{Ablation study on meta learning vs. multi-task learning. The results are on the FS split using R50 + BERT backbones.}
     \label{tab:performance_per_meta}
    \vspace{-3mm}
\end{table*}
\subsection{Ablation studies} Table~\ref{tab:ablation} reports the ablations on using meta-learning versus multi-task learning (MTL), as well as on using puzzle-specific image heads. As is expected, when adding the puzzle heads, the performances improve. We find that using meta-learning is important and leads to a dramatic ($\sim$12\%) improvement in performance. Our results also confirm that both vision and language modalities are essential for solving \name. In more detail, Table~\ref{tab:performance_per_meta} compares few-shot performance on the FS split with and without meta-learning (instead using multi-task learning). As is clear from the table, using meta-learning leads to a substantial improvement in performance, especially on the classifier setting (+4\%). 

\begin{figure}
    \centering
    \includegraphics[width=4cm]{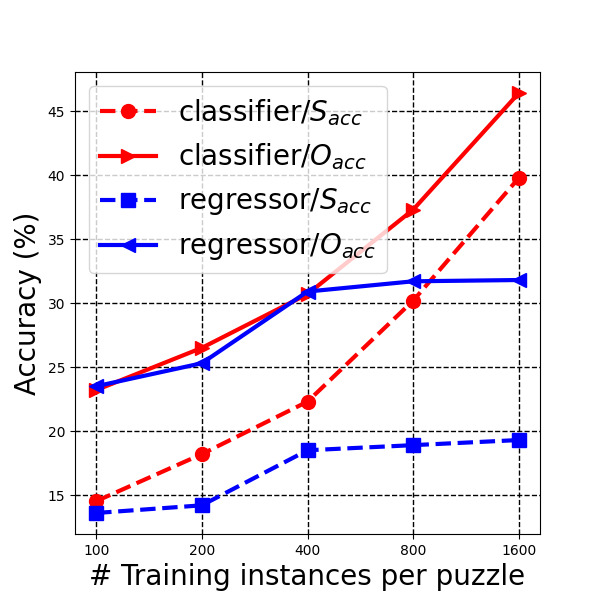}
    \includegraphics[width=4cm]{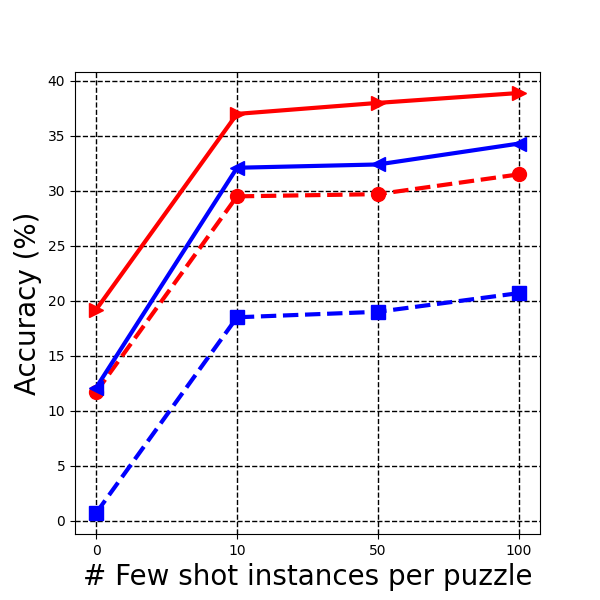}
    \caption{Performance against increasing training set size. Left: IS split and right: PS split with few-shot instances. Both experiments use R50 + learned embeddings (Emb.).}
    \label{fig:sensitivity}
\end{figure}

\noindent\textbf{How many training images to use?} In Figure~\ref{fig:sensitivity} (left), we analyze the sensitivity of the models for increasing training size by varying training sizes from 100 to 1600 instances per puzzle and training the backbones on the entire puzzle set. The plot shows that more training data is useful with $\sacc$ increasing almost quadratically with data size when using a classification loss (perhaps because of better gradients against the regression loss).

\begin{figure*}[ht]
    \centering
    \begin{subfigure}{1\linewidth}
        \includegraphics[width=16cm]{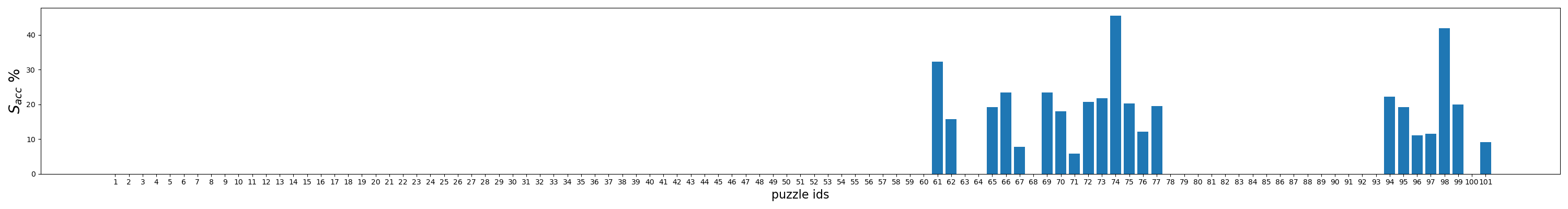}
        \caption{CLIP, PS split, and Classification (avg. $\sacc$=11.9\%).}
        \label{fig:perf_clip_ps_cls}
    \end{subfigure}
    \begin{subfigure}{1\linewidth}
        \includegraphics[width=16cm]{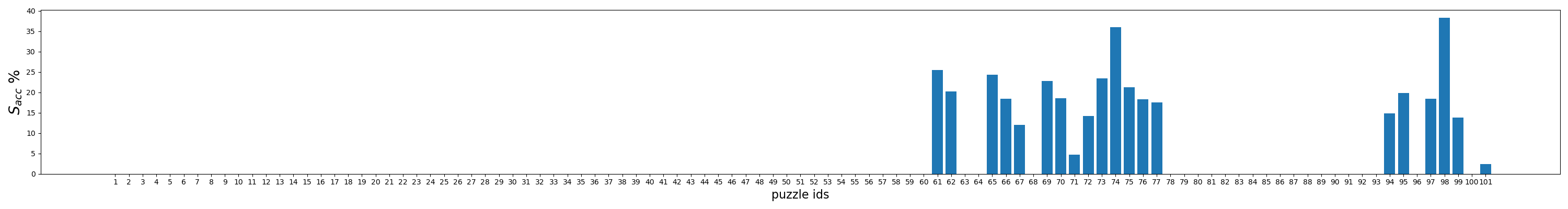}
        \caption{ResNet-50 + BERT, PS split and Classifier (avg. $\sacc$=11.7\%). }
        \label{fig:perf_r50_bert_ps_cls}
    \end{subfigure}    
    \begin{subfigure}{1\linewidth}
        \includegraphics[width=16cm]{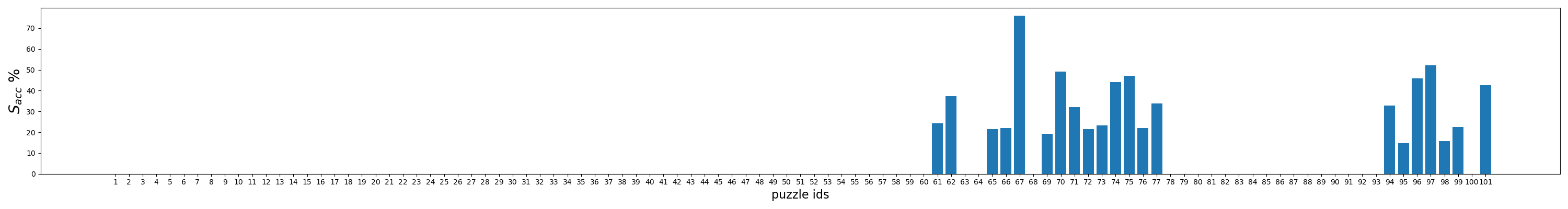}
        \caption{ResNet-50 + BERT, FS split, and Classifier (avg. $\sacc$=24.4\%).}
        \label{fig:perf_r50_bert_fs_cls}
    \end{subfigure}
    \begin{subfigure}{1\linewidth}
        \includegraphics[width=16cm]{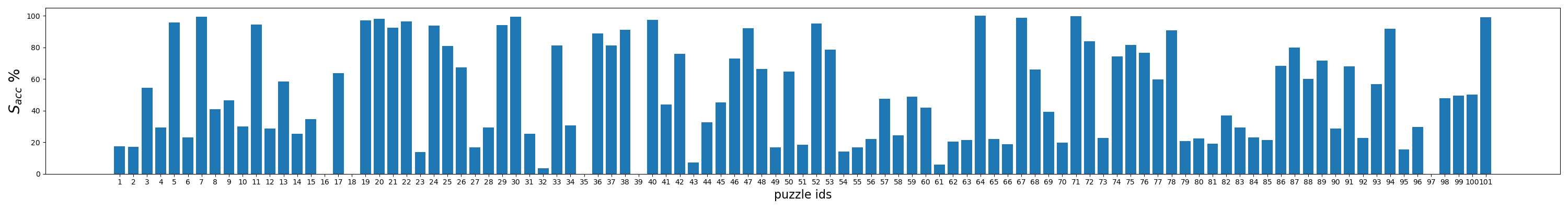}
        \caption{ResNet-50 + BERT, IS split, and Classifier (avg. $\sacc$=42.8\%)}
        \label{fig:perf_r50_bert_is_cls}
    \end{subfigure}
    \begin{subfigure}{1\linewidth}
        \includegraphics[width=16cm]{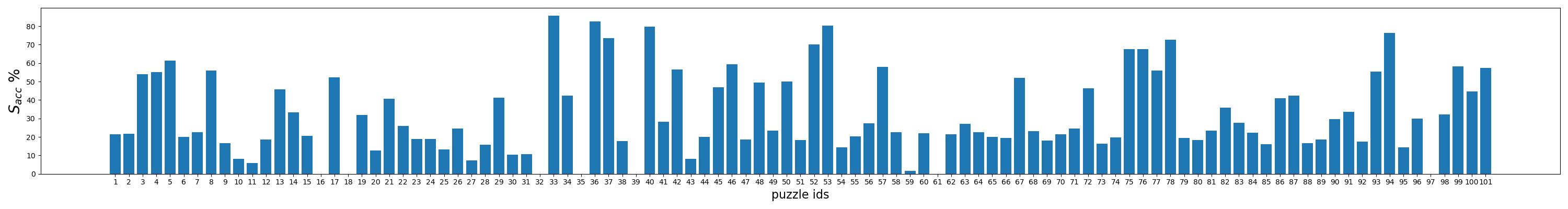}
        \caption{Swin-b+ Emb., IS split, and Classifier (avg. $\sacc$=21.6\%)}
        \label{fig:perf_swin_bert_is_cls}
    \end{subfigure}
    \vspace{-3mm}
    \caption{We compare the $\sacc$ accuracy on the instances within every root puzzle (test set) between our best performing (ResNet-50+BERT) IS model (d) against selected variants such as: (a) pretrained CLIP model for PS split, (b) R50 + BERT on PS with fine-tuned R50, and (c) R50 + BERT on FS split with fine-tuned R50 using 100 few-shot examples. We also compare to a Transformer backbone variant (swin-b with learned embeddings) on the IS split. For the PS and FS plots above, note that we evaluate only on the 21 test puzzles, and thus other puzzles have zero accuracies. The x-axis in the plots show the root puzzle numbers. }
    \label{fig:puzz_acc_stats}
\end{figure*}

\noindent\textbf{Few-shot Performance Sensitivity.} We analyze the impact of the number of few-shot samples against performance in Figure~\ref{fig:sensitivity}~(right) where we see a slow, but increasing gains with more few-shot examples. This experiment used R50 + Emb. While there is a steady increase in performance even for $m=10$, the improvements start to flatten, perhaps because some of the puzzles benefit significantly from the few-shot examples, but other puzzles are very hard that the few-shot examples are perhaps not sufficient (see Figure~\ref{fig:puzz_acc_stats}). 

\subsection{Detailed comparisons across splits}
In Table~\ref{tab:category-wise-split-performance}, we provide category-wise results on the various data splits, such as IS, AS, PS, and FS splits. While, the PS split performances (that look for extreme generalization on puzzles that the model has not seen during training) are close to random accuracy, the AS and FS splits show better performances.  Note that the FS experiments used only 100 samples from the respective root puzzles during training. As the table clearly shows, the trends are very consistent with regard to the accuracies on each of the skill categories, with the PS having the least performance, followed by FS, AS, and IS. Further, for AS, the regression setting works significantly better than the classification setting, as expected. 

\begin{table*}[ht]
    \centering
    \resizebox{\textwidth}{!}{
    \begin{tabular}{cccccccccccc}
    \toprule[1pt]
        Row&Split & cls./reg. & Count & Arithmetic & Logic & Path Trace & Algebra  & Measure & Spatial & Pattern Finding & Average\\
        \midrule 
         1&Instance (IS) & cls. & 48.5/59.3 &  46.1/54.9 &  56.7/60.2 &  26.5/28.4 &  28.5/39.7 &  65.6/75.4 &  44.3/46.2 &  39.9/45.3 &  42.8/50.2\\
        2&Answer (AS) & cls. & 0.1/23.8 & 1.5/13.2 & 0.0/16.8 & 0.0/1.6 & 0.4/17.3 & 0.0/21.1 & 0.0/6.0 & 0.0/15.0 & 0.19/10.2\\
        3&Puzzle (PS) & cls. & 10.9/18.3 & 6.96/15.8 & 12.8/20.8 & 19.6/19.7 & 7.95/15.1 & 16.9/26.7 & 13.4/17.7 & 0.0/21.2 & 11.7/18.9\\
        4&Few-shot (FS) & cls. & 23.9/37.3 & 32.7/41.2 & 32.7/40.7 & 22.1/22.2 & 10.2/27.5 & 17.4/32.8 & 28.4/35.3 & 33.9/33.9 & 24.4/33.4\\
        \midrule
        5&IS & reg. & 36.3/53.8 & 26.4/46.5 & 28.8/44.2 & 20.9/23.8 & 22.3/38.8 & 41.9/64.5 & 30.3/38.8 & 33.3/41.1 & 29.4/43.8\\
        6&AS & reg. & 12.0/28.4 & 10.4/25.7 & 19.6/30.8 & 9.5/10.6 & 3.64/18.3 & 9.42/28.6 & 14.1/21.1 & 25.5/30.9 & 16.3/23.4\\
        7&PS & reg. & 12.0/22.8 & 5.08/21.3 & 4.24/16.2 & 18.4/18.4 & 4.89/22.2 & 15.1/25.9 & 11.9/17.9 & 19.0/19.0 & 8.21/19.7\\
        8&FS & reg. & 19.8/33.6 & 13.9/26.3 & 18.2/26.9 & 18.7/18.7 & 10.3/24.4 & 11.6/25.8 & 20.8/29.8 & 21.9/22.3 & 16.7/26.5\\
    \bottomrule[1pt]
    \end{tabular}
    }
    \vspace{-2.5mm}
    \caption{Category-wise puzzle performances on various data splits for classification or regression settings using R50 + BERT model, with fine-tuned R50 model. We used 100 puzzle instances for the $m$-shot split (FS). Each entry shows the $\sacc/\oacc$ (\%; higher is better).}
     \label{tab:category-wise-split-performance}
    \vspace{-3mm}
\end{table*}

\begin{figure}[t]
    \centering
    \includegraphics[width=8cm,trim={0cm 0.5cm 0cm 0cm},clip]{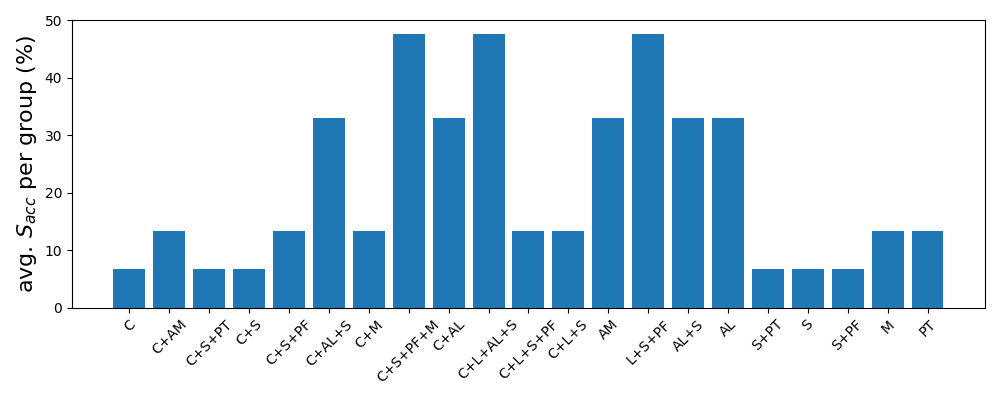}
    \caption{Distribution of performances across top-performing compositional skill categories.}
    \label{fig:comp_stats_perf}
\end{figure}

\begin{figure}[t]
    \centering
    \includegraphics[width=\columnwidth]{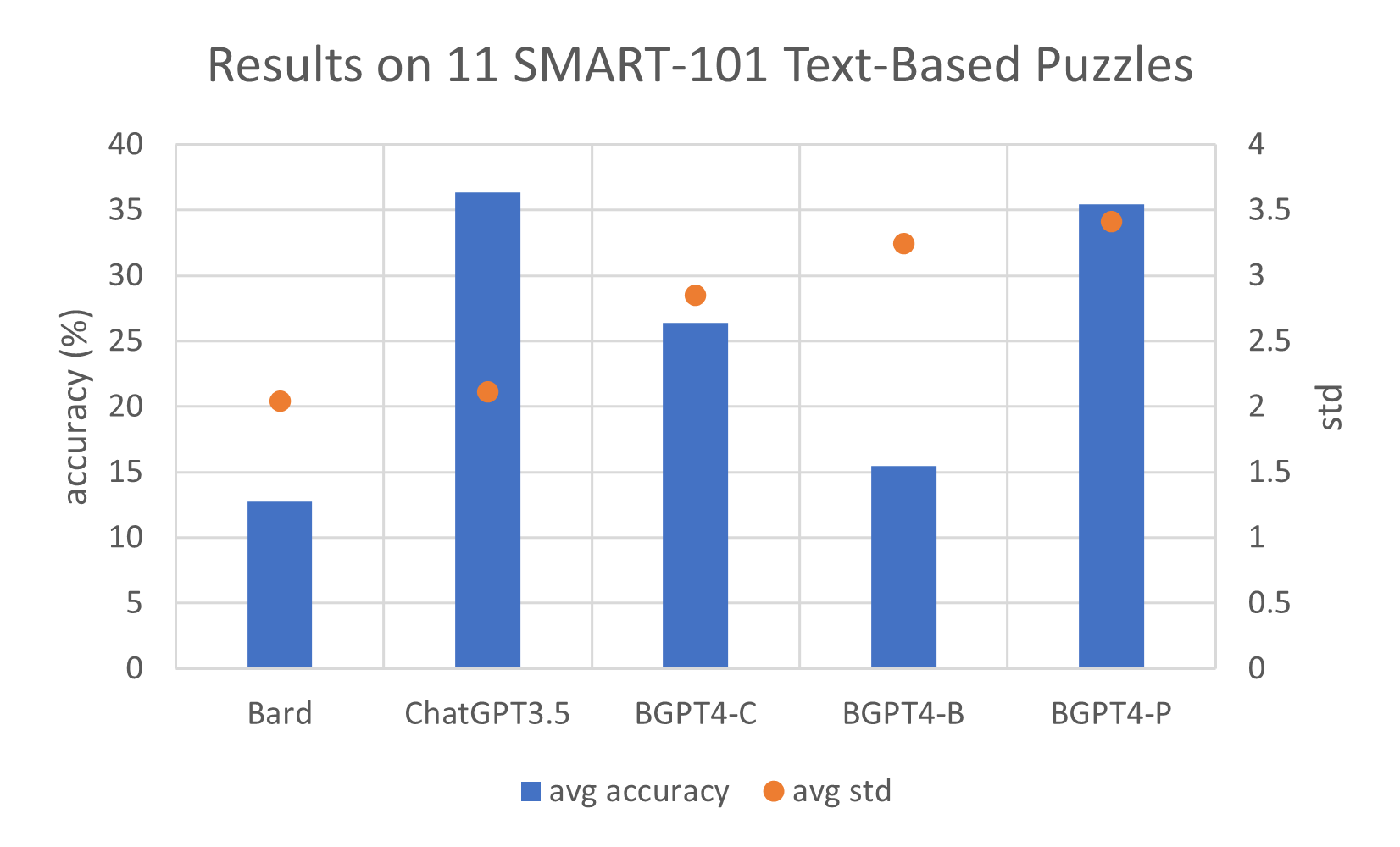}
    \caption{The average accuracy and average standard deviation (std) of the answer frequencies of the Bard~\cite{bard}, ChatGPT3.5~\cite{chatgpt}, BGPT4-C~\cite{bing}, BGPT4-B~\cite{bing}, and BGPT4-P~\cite{bing} for the 11 puzzles in Table~\ref{tab:chatgpt}. Specifically, the std is computed from the answer frequencies. For example, if the answer frequencies of a puzzle is 0/1/2/7/0/0 for option A/B/C/D/E/others out of 10 trials, then the std will be 2.49. A lower std indicates a flatter answer distribution.}
    \label{fig:comp_acc_std}
\end{figure}

\begin{figure*}[t]
     \centering
     \includegraphics[width=\textwidth]{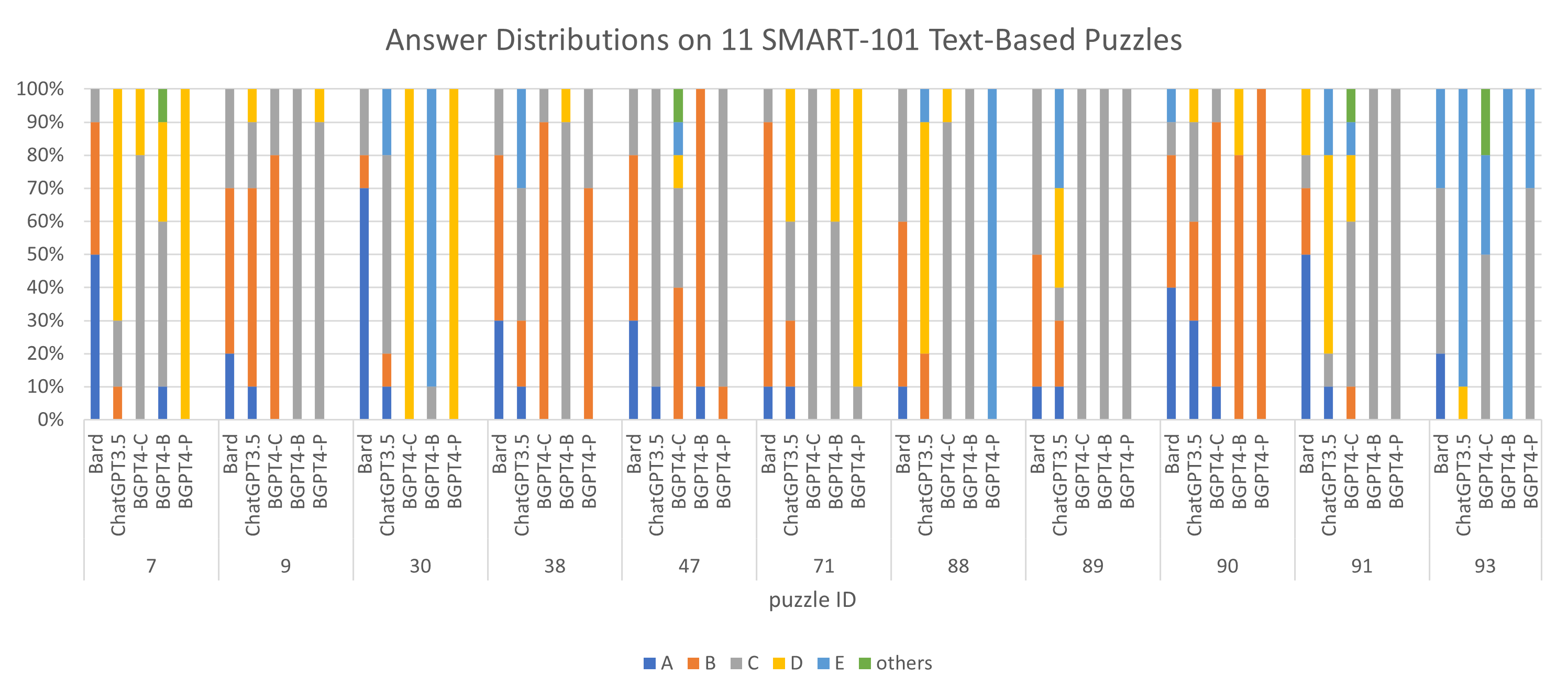}
     \caption{The answer distributions of the Bard~\cite{bard}, ChatGPT3.5~\cite{chatgpt}, BGPT4-C~\cite{bing}, BGPT4-B~\cite{bing}, and BGPT4-P~\cite{bing} for the 11 puzzles in Table~\ref{tab:chatgpt}.}
     \label{fig:ans_dist}
\end{figure*}
\subsection{Performances across architectures}
In Figure~\ref{fig:puzz_acc_stats}, we study the trends in the performances ($\sacc$) on each puzzle for a few backbones and splits. Specifically, we show the results using the PS split in Figure~\ref{fig:perf_clip_ps_cls} when using a pre-trained CLIP vision-and-language backbone (without any fine-tuning), Figure~\ref{fig:perf_r50_bert_ps_cls} when using the R50+BERT backbone with the R50 being fine-tuned, and Figure~\ref{fig:perf_r50_bert_fs_cls} when few-shot examples are available for the test puzzles in PS, and using the same R50+BERT fine-tuned model as in Figure~\ref{fig:perf_r50_bert_ps_cls}. In Figure~\ref{fig:perf_r50_bert_is_cls}, we plot side-by-side the results on all test puzzles when using the fine-tuned R50 + BERT on the IS split, and in Figure~\ref{fig:perf_swin_bert_is_cls}, we show the same results when R50 is replaced by a Transformer model (swin-b) and language model using learned embeddings. 

As can be seen from the bar plots in Figure~\ref{fig:puzz_acc_stats}, using the pre-trained CLIP model on the PS split appears to perform by nearly 20\% accuracy on puzzles 61 (algebra), 74 (logic), and 98 (arithmetic), while the lowest on puzzle 71 (arithmetic). When fine-tuning a R50 + BERT model on the train puzzles of PS, the performance trends remain similar, however the performances on puzzles 96 and 101 drop, while other puzzles show some minor improvements, suggesting that there is some beneficial and generalizable training that happens in the model. When using few-shot examples for the puzzles in PS as in Figure~\ref{fig:perf_r50_bert_fs_cls}, we see a significant jump in performance with puzzle 67 (arithmetic) -- which is only $\sim$5\% accuracy in Figures~\ref{fig:perf_clip_ps_cls} and~\ref{fig:perf_r50_bert_ps_cls} -- shows close to 70\% accuracy, and the overall $\sacc$ improving by 13\%. 

We also provide comparisons to IS splits using R50 + BERT and swin-b + Emb. in Figures~\ref{fig:perf_r50_bert_is_cls} and~\ref{fig:perf_swin_bert_is_cls}. As is seen, R50+BERT IS has nearly 25 puzzles on which $\sacc$ is more than 80\%. There are six sequential puzzles (16, 18, 35, 39, 63, 100), of which the first four has nearly zero accuracy, while on puzzle 97 (counting), we have the PS splits perform better. In general, we found that the performance on sequential puzzles are lower in all our IS/AS architectures, perhaps because of the lack of sufficient training data and the diversity in the type of answers these puzzles entail (e.g., predicting letters, or digits, or combinations of choices, etc.). 

The trends in the IS split for R50 + BERT continues for swin-b + Emb., except that the performance numbers are inferior overall. We find that Transformer models in general performs weaker on our puzzles, and that is the reason we made more comparisons on Transformers (in Table~\ref{tab:category-wise-performance}) variants such as swin-b, swin-t, ViT, Cross-Transformers, and MAE. We believe the image patch tokenization used in Transformers (even with PE) is making the reasoning hard to learn with limited (textureless) data, while having a larger receptive field (as in ResNet) is beneficial (e.g., learning a circle from patches vs. seeing it in full).

\subsection{Analysis on individual skill categories}
Instead of training a model on all the root puzzles to learn a common backbone, what if we trained the backbone on each of the 8 puzzle categories? This will allow us to understand if it is easier to learn individual skills than broad range of skills with one model. In Table~\ref{tab:performance_per_cat_puzzle}, we show detailed results on this analysis, where we trained a ResNet-18 model for each puzzle category using the learned language embeddings. Our results suggest that there is an improvement of nearly 10\% against a single ResNet-18 model, and demonstrate significant benefits in each skill category as expected. Specifically, we find nearly 15\% improvement in arithmetic, spatial reasoning, and pattern finding when training and evaluating on that particular class. 

\begin{table*}[ht]
    \centering
    \resizebox{\textwidth}{!}{
    \begin{tabular}{cccccccccc}
    \toprule[1pt]
        Model  & Count & Arithmetic & Logic & Path Trace & Algebra  & Measure & Spatial & Pattern Finding & Average\\
        \midrule 
        Single ResNet-18 & 44.0/54.0 & 8.8/19.8 & \textbf{41.1/47.6} & 24.5/26.7 & 13.7/26.5 & 30.9/40.2 & 43.3/45.5 & 29.5/34.8 & 29.4/37.4\\
        Specific ResNet-18 & \textbf{46.9/57.5} &\textbf{22.9/31.3} & 38.6/45.1 & \textbf{25.9/26.9} & \textbf{21.0/31.0} &  \textbf{45.7/44.9} & \textbf{58.7}/\textbf{59.3} & \textbf{47.1}/\textbf{52.8} & \textbf{38.3/43.6}\\
    \bottomrule[1pt]
    \end{tabular}
    
    }
    \caption{Comparison when training models on each category (IS split).}
     \label{tab:performance_per_cat_puzzle}
    \vspace{-3mm}
\end{table*}

\subsection{Analysis of compositional skills} Does the model find learning composite skills harder than individual skills?  Note that our categorization of composite skills is different from the one/two/three step reasoning delineated in MK. To explain this better, consider root puzzle 3 shown in Table~\ref{tab:composite_samples}. To solve this puzzle, the basic skill needed is to count the number of cherries on the pie, however that by itself does not get one to solve the puzzle. One also needs arithmetic skills to divide the number of cherries by the number of pieces a child receives, and thus one needs both counting and arithmetic skills for solving this puzzle. Take for example, puzzle 2. This puzzle is again a counting puzzle, and goes into the category of \emph{counting}. For this puzzle, one only needs the skill to count. However, the skill to count is perhaps more involved than the counting that was used in puzzle 3 because for puzzle 2 one needs to identify the three pointed stars and count them. Thus, if a model performs poorly on a basic skill, that does not necessarily suggest that it will perform poor on a composite skill set that includes that basic skill. 

To quantitatively analyze this intuition, in Figure~\ref{fig:comp_stats_perf} plots the distribution of performances on the IS split for varied compositions of skills as characterized in Figure~\ref{fig:app_stats} using our best performing ResNet-50 + Emb. model. We show only the top-performing categories (14 out of the 41 composite skills). As is clear from the figure, it looks like while solving puzzles belonging to some of the basic skill set categories, e.g., counting, is challenging, the networks appear to do well on puzzles that need several skills. This suggests that  the puzzles in \name are such that just an algorithm with good basic skills need not mean it performs well in solving puzzles needing composite skills and a great performance on puzzles demanding composite skill sets does not necessarily suggest a good performance on basic skill puzzles -- a holistic approach that can showcase good performances on both basic and composite skills is perhaps important.

\begin{table*}[t]
\centering
\vspace{-.7em}
\footnotesize
\resizebox{.997\textwidth}{!}{
\begin{tabular}{p{4cm}p{4cm}p{4cm}p{4cm}}
\centering Puzzle \#2 (C) & \centering Puzzle \#3 (C+AM) & \centering Puzzle \# 36 (C+S+AL+L)& \hspace{8mm}Puzzle \# 73 (C+AL+S)\\
\centering\includegraphics[width=4cm,height=4cm]{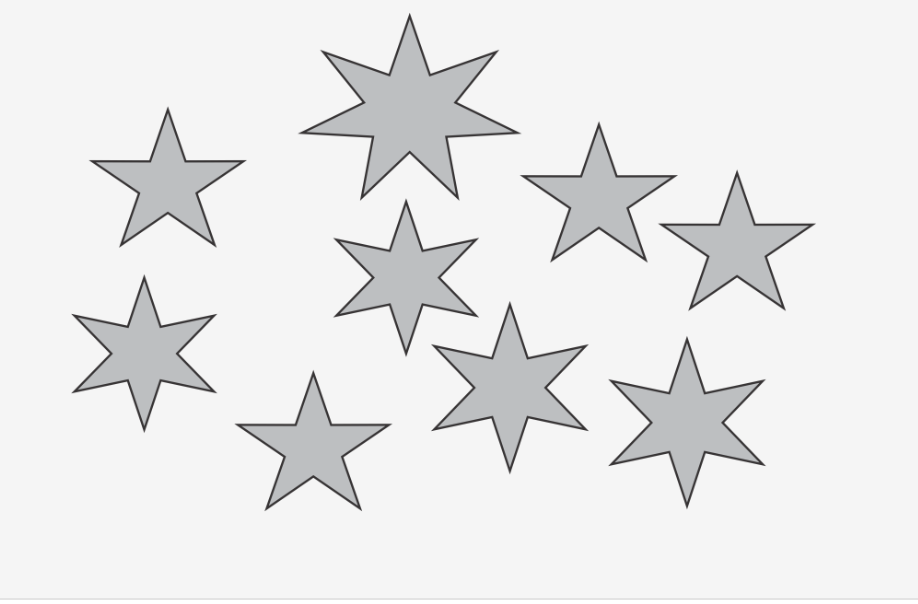} &\centering\includegraphics[width=4cm,height=4cm]{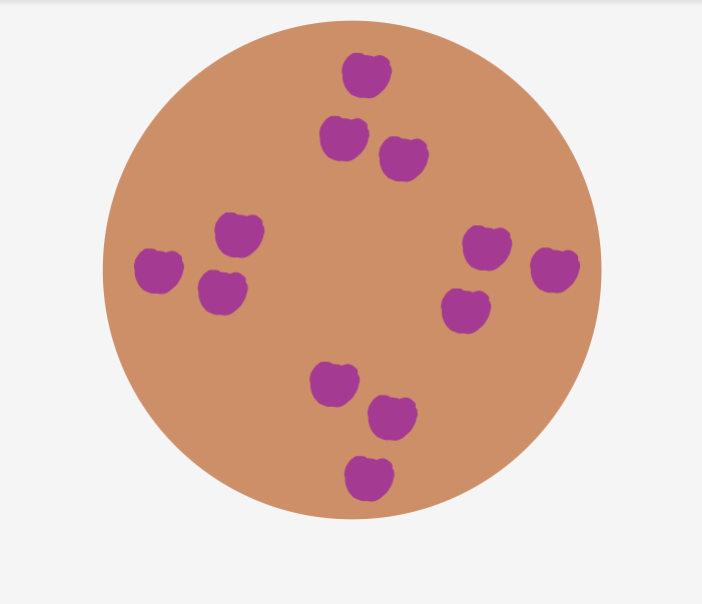} &\centering\includegraphics[width=4cm,height=4cm,trim={3cm 3cm 3cm 3cm},clip]{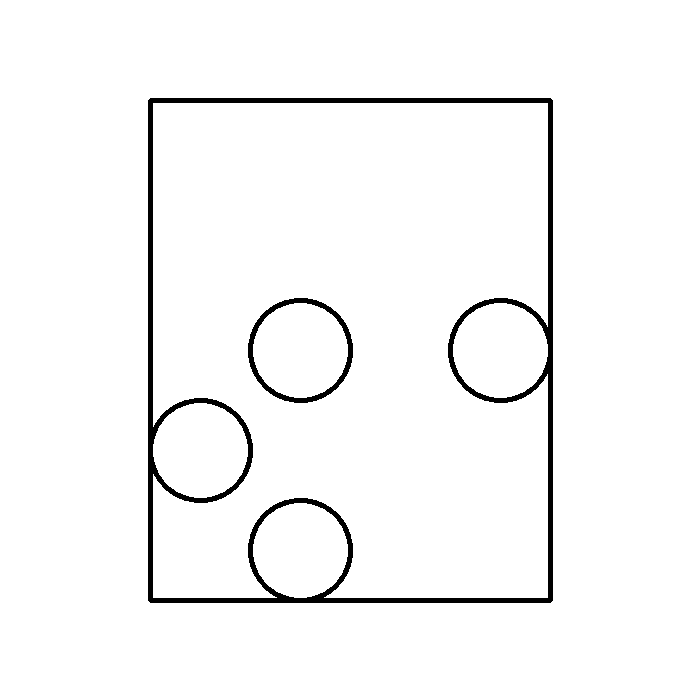}
&\includegraphics[width=4cm,height=4cm]{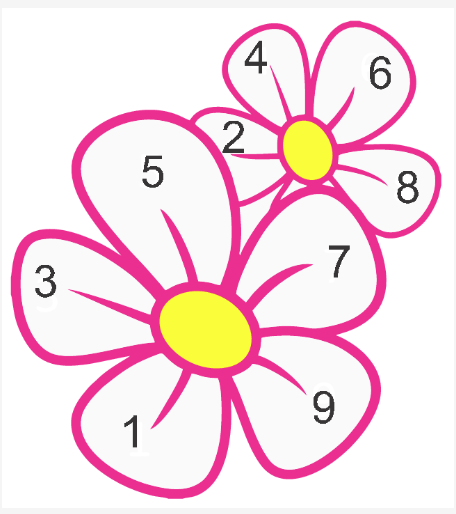}\\
\textbf{Question}:In the picture, there are stars with 5 points, stars with 6 points, and stars with 7 points. How many stars that have only 5 points are there?  &\textbf{Question}: The entire pie seen in the picture is divided among several children. Each child receives a piece of pie with three cherries on top. How many children are there? &\textbf{Question}: A man's hens lay white eggs and brown eggs. He puts eggs in the box shown in the figure. Two brown eggs cannot touch each other. At most, how many brown eggs can he put in the box? & \textbf{Question}: A number is written on each petal of two flowers. One petal is hidden. The sums of the numbers on the two flowers are equal. What number is written on the hidden petal?\\
\midrule
\end{tabular}
}
\vspace{-4mm}
\caption{We show four root puzzles and their composite classes. C: counting, AM: arithmetic, AL: algebra, L: logic, and S: spatial reasoning. Also compare the performances of the above puzzles to those in Figure~\ref{fig:puzz_acc_stats}.}
\label{tab:composite_samples}
\end{table*}
\begin{table*}[ht]
     \centering
     \begin{tabular}{ccccccccccccc}
     \toprule
     puzzle ID &7 &9 &30 &38 &47 &71 &88 &89 &90 &91 &93 &mean\\
     \midrule
     Category & AL & S & AM & AM & AM & AM & AM & C & AL & L & M & \\
     \midrule
    Human & NA & NA & NA & NA & NA & 60.4 & NA & NA & NA & NA & NA & 60.4\\
    Bard~\cite{bard} &0.0 &20.0 &0.0 &50.0 &0.0 &0.0 &0.0 &10.0 &10.0 &20.0 &30.0 &12.7\\
    ChatGPT3.5~\cite{chatgpt} &70.0 &10.0 &0.0 &20.0 &0.0 &40.0 &70.0 &10.0 &30.0 &60.0 &90.0 &36.4\\
    BGPT4-C~\cite{bing} &20.0 &0.0 &100.0 &90.0 &10.0 &0.0 &100.0 &0.0 &10.0 &20.0 &30.0 &26.4\\
    BGPT4-B~\cite{bing} &30.0 &0.0 &0.0 &0.0 &0.0 &40.0 &0.0 &0.0 &0.0 &0.0 &100.0 &15.5\\
    BGPT4-P~\cite{bing} &100.0 &0.0 &100.0 &70.0 &0.0 &90.0 &0.0 &0.0 &0.0 &0.0 &30.0 &35.5\\
    PS split & NA & NA & NA & NA & NA & 4.65 & NA & NA & NA & 25.5 & NA & 15.1\\
     IS split & 98.0 & 14.0 & 100.0 & 64.6 & 93.7 &  56.7 & 21.3 & 55.7 & 51.3 & 26.3  & 34.0 & \textbf{55.9}\\

     \bottomrule
     \end{tabular}
     \caption{The performance of ChatGPT~\cite{chatgpt} and variants (in $\oacc$ (\%); out of 10 trials per puzzle) on the 11 puzzles which do not require the information from the image to solve. Note that IS split uses our ResNet-50 + Emb. model, and is trained on the puzzles. For the PS split and the human performance comparisons, we show only the puzzles that overlap with those used for ChatGPT evaluation (see Appendix~\ref{sec:chatgpt}).}
     \label{tab:chatgpt}
\end{table*}

\section{\name and large language models}
\label{sec:chatgpt}
Next, we evaluate the generalization performance of recent popular large language models (LLM), including Bard~\cite{bard}, ChatGPT~\cite{chatgpt}, and Bing's AI chatbot~\cite{bing} on 11 puzzles in SMART-101 that do not need the information from the puzzle image for the solution.For ChatGPT, we conducted our experiments before GPT-4~\cite{gpt4} was announced, so we used the GPT version 3.5 in this study. We use ChatGPT and ChatGPT3.5 interchangeably and use both of them to refer to the same model. We use Bing's AI chatbot interface to access GPT-4. Since this chatbot includes three different conversation styles (\ie, creative, balanced, and precise), we use BGPT4-C, BGPT4-B, and BGPT4-P to denote these 3 variations in this work, respectively.\footnote{Our experiments using the Bard, ChatGPT3.5, and 3 variations of Bing's AI chatbot (BGPT4-C, BGPT4-B, and BGPT4-P) on March 29, 2023, December 7, 2022, and March 21, 2023, respectively.}  Please see Appendix~\ref{sec:chatgpt_responses} for details. For each puzzle, we provide the question text and the answer options as input to each LLM, and use its response (\ie, selection of the option) for evaluation. As the LLM's response varies on each attempt, we repeat each input 10 times, and compute the average accuracy among all the trails. Specifically, for each attempt, we reset and start a new conversation such that the LLM's output is not affected by the previous chat history. During the entire experiment, we give no feedback to each LLM after providing the input (question text and the answer options). Table~\ref{tab:chatgpt} summarizes the performance of our investigated LLMs in $\oacc$ on the 11 puzzles from \name.  From the table, we find that the best LLMs (ChatGPT3.5 and BGPT4-P) fail in nearly 65\% of the time. From the LLMs' responses provided in Appendix~\ref{sec:chatgpt_responses}, their reasoning abilities appear very similar to those of children, however we find that the final answers are often inaccurate; suggesting a gap in their reliability and semantic reasoning abilities. We also find that the answers for each puzzle that these LLMs produce are not stochastic, instead they are convincingly selecting the wrong answers (\eg, Figure~\ref{fig:ans_dist} where many of the option selections are predominantly of a single color for all the puzzles). We summarize the average accuracy and average standard deviation (std) of the answer frequencies of these LLMs in Figure~\ref{fig:comp_acc_std}, where we find that ChatGPT3.5 and Bard produce relatively more diverse answer distributions, which is less ideal for the questions with one definite answer.

\section{Conclusions}
We started by asking the question: \emph{are deep neural networks SMARTer than second graders?} Our analysis in Table~\ref{tab:category-wise-performance} shows that the performances of SOTA deep models are significantly below second graders on \name (77\% against 20\%). Surprisingly, even under the supervised setting (IS) -- when the networks have seen similar instances of a puzzle -- the performance is inferior (43\%). However, with sufficient training data, SOTA models do demonstrate some level of learning algorithmic skills (e.g.,~arithmetic, spatial reasoning, \etc), yet struggle on simple algebra or path tracing problems.  While, the recent ChatGPT~\cite{openai2023gpt4} model demonstrates significant generalization abilities, it still appears to struggle with long-range multi-step reasoning. This is surprising to some extent, as ChatGPT is known to produce high-quality essays and score well in competitive exams.  That being said, our analysis also shows that training with a few examples from new problems, our model can quickly learn new skills. To conclude, the answer to our overarching question is clearly \emph{no}, and there appears to be a significant gap in the perceived competency of AI models and their true algorithmic reasoning abilities. We hope \name offers a solid step to make advancements in that direction.

\noindent\textbf{Limitations and future extensions.}
Our \name dataset consists of only 101 puzzles taken from nearly 10 years of Math Kangaroo USA competitions. While these puzzles cover a diverse set problem types, the number of puzzles for each skill set may not be evenly distributed; this could bring bias in the model training. Further, our skill set categorization (into counting, measurement, \etc), while useful to organize our results for a systematic analysis, may be an overestimate as many of the puzzles need compositional and foundational image understanding skills for the solution. Another direction to improve the dataset would be to use richer large language models for diversifying the questions associated with the puzzles. Another direction is to extend \name via incrementally increasing the puzzle complexity and skill sets. 

\section*{Acknowledgements} We thank Joanna Matthiesen (CEO of Math Kangaroo USA) for sharing with us the human performance statistics and permission to use the puzzle images from the Math Kangaroo USA Olympiad in this paper. Authors would like to thank Tim K. Marks, Mike Jones, Moitreya Chatterjee, and Anthony Vetro for valuable discussions and feedback.

\begin{figure*}
    \centering
    \begin{subfigure}{0.24\linewidth}
        \includegraphics[width=4cm]{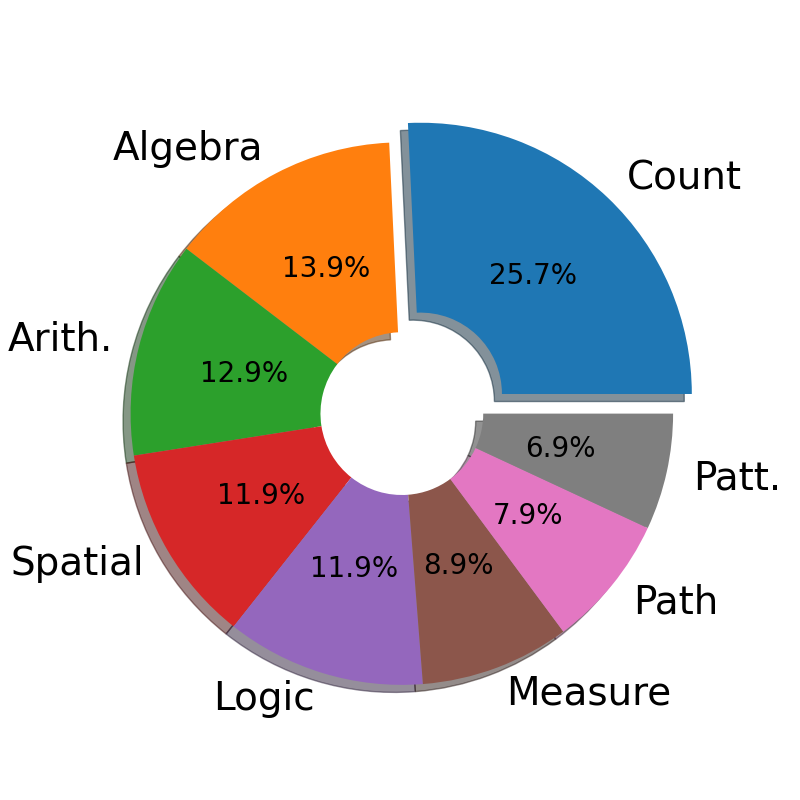}
        \caption{}
        \label{fig:cls_distr}
    \end{subfigure}
    \begin{subfigure}{0.24\linewidth}
        \includegraphics[width=4cm]{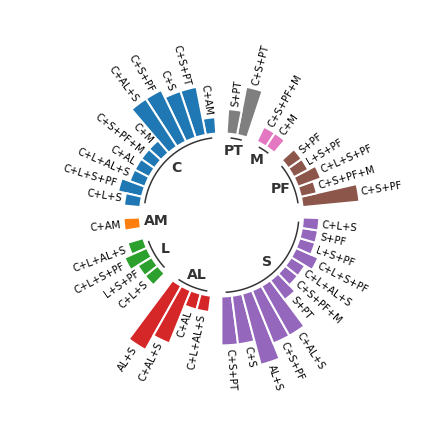}
        \caption{}
        \label{fig:comp_distr}
    \end{subfigure}
    \begin{subfigure}{0.24\linewidth}
        \includegraphics[width=4cm]{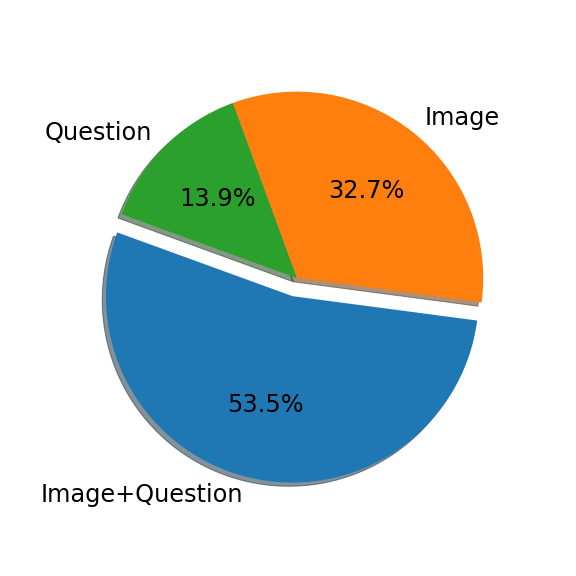}
        \caption{}
        \label{fig:img_quest_distr}
    \end{subfigure}
    \begin{subfigure}{0.24\linewidth}
        \includegraphics[width=4cm]{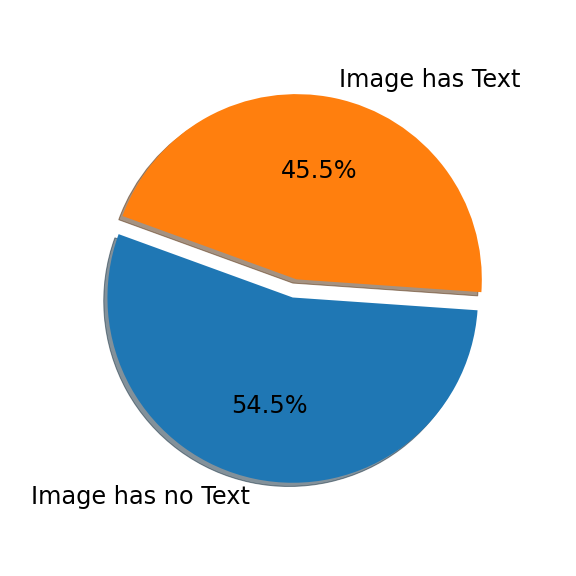}
        \caption{}
        \label{fig:img_text_distr}
    \end{subfigure}
    \caption{Statistical analysis of various properties of the \name dataset. In (a), we plot the distribution of primary algoritmic skills needed to solve the puzzles and (b) plots the various multi-step reasoning skills needed. In (c), we plot the distribution of puzzles that need both image and question reasoning, those needing to understand only question for the solution, and those needing only images. In (d), we plot the distribution of root puzzles that have text within images, and the method needs to recognize and associate this text to select the answer. See Table~\ref{tab:puzzle_ids_and_skills} at the end of the Appendix for the mapping between puzzle id and the skill class.}
    \label{fig:stats}
\end{figure*}
\appendix
\section*{Appendix}
\section{Data splits}
\label{sec:data_splits}
For the instance-split (IS), we used 80\% of the 2000 instances per root puzzle for training the models, 5\% for validation, and 15\% for the test. For answer-split (AS), we selected the median answer from the answer distribution of the instances, and used this answer only during test. Thus, it is not possible to delineate how many instances will be used for every root puzzle in this case. However, we found that the total number of instances used during test is similar to the number used in IS split. 

For the puzzle and few-shot splits (PS and FS), we divided the root puzzles to 77 for training, 3 for validation and 21 for testing, where the test puzzles are selected as those for which we have human performance data available. We used 15\% of the test puzzle instances for the evaluation (these are the same instances that are used for IS and AS splits as well, however in PS split, the training model never sees the 80\% of the root puzzles from this set of 21). In the~\name dataset, the puzzles are numbered as:1$\sim$101 and the instances in each puzzle as: 1$\sim$2000. The PS test set uses puzzles: 61, 62, 65, 66, 67, 69, 70, 71, 72, 73, 74, 75, 76, 77, 94, 95, 96, 97, 98, 99, 101. In IS/AS, we use instances 1701--2000 for evaluation on all puzzles. These 21 puzzles belong to the following categories:counting=2, logic=5, algebra=4, path tracing=1, measurement=1, spatial reasoning=4, arithmetic=3, and pattern finding = 1.

\section{Details of Math Kangaroo}
\label{sec:mk_details}
Math Kangaroo is an annual mathematical Olympiad footnote{\url{https://mathkangaroo.org/mks/}} for students from grades 1--12 and held across all the states in the USA . As per the \href{https://mathkangaroo.org/mks/about-math-kangaroo/statistics/#1629299252592-96cec2cc-f219}{statistics}, more than 35,000 students participated in the year 2022 of this Olympiad. 
\vspace*{0.3cm}

\noindent\textbf{Exam Structure:} The MK competition is typically for 75 minutes and consists of 24 questions, 
with questions 1--8 weighting 3 points and considered ``easy'' by the question designers. These questions mainly constitute one-step or a single skill to solve them; Questions 9--16 are 4 points and more difficult, needing two or three steps; 17--24: most difficult, multi-step questions. For our root puzzles, we mainly considered puzzles in the range 1--16.
\vspace*{0.3cm}

\noindent\textbf{Puzzle Selection:} To create our puzzle set, we selected root puzzles from MK-2012--2021. While most puzzles have a picture and a question, some of the images are not necessary for solving the puzzle and is meant to provide better context for children solving them, especially for children of younger age (e.g., first and second graders). A few puzzles do not have an image instead has only the question. For such puzzles, we use a blank white image as a placeholder when training the neural networks. Further, nearly half of the puzzles have text or options within the image, which needs to be spotted and inferred to select the correct answer. Figure~\ref{fig:stats} shows these statistics, we also show the skill sets and the compositional skill sets. See Table~\ref{tab:puzzle1} for examples of problems that need inference on text within a puzzle image.

\noindent\textbf{Instance Generation:}
As alluded to above (in the main paper), generating the instances for each root puzzle needs varied programming and mathematical skills. We used Python for implementing the code to replicate the puzzles, and predominantly used either Opencv toolbox\footnote{\url{https://docs.opencv.org/4.x/d6/d00/tutorial_py_root.html}} or the PyPlot package\footnote{\url{https://matplotlib.org/stable/tutorials/introductory/pyplot.html}} for rendering the instances. Each root puzzle took from 15 minutes to 3-4 days to program. After the generation, each puzzle was reviewed for their quality, correctness, and adherence to the difficulty levels as expected in the original MK puzzle. Given our written programs are generic, we could easily change the program arguments and create instances that are of arbitrary difficulty levels, e.g., for puzzle that uses $3\times 3$ grids, we could extend the program logic for any grid size of $n\times n$. 


\section{Root puzzle indices and skill categories}
\label{sec:puzzle_categories}
For reference, in Table~\ref{tab:puzzle_ids_and_skills}, we list all the puzzle ids in \name and their skill set categories.
\begin{table}[]
    \centering
    \begin{tabular}{c|c c|c}
     ID & class &  ID & class\\
    \hline
1&	path finding                      &  51&	measuring      \\
2&	counting                          &  52&	counting\\
3&	counting                          &  53&	counting\\
4&	counting                          &  54&	path finding\\
5&	counting                          &  55&	spatial reasoning\\
6&	arithmetic                              &  56&	logic\\
7&	algebra                           &  57&	counting\\
8&	counting                          &  58&	arithmetic\\
9&	spatial reasoning                 &  59&	arithmetic\\
10&	algebra                           &  60&	counting\\
11&	arithmetic                              &  \ul{61}&	algebra\\
12&	spatial reasoning                 &  \ul{62}&	logic\\
13&	counting                          &  63&	algebra\\
14&	counting                          &  64&	measuring\\
15&	algebra                           &  \ul{65}&	path finding\\
16&	path finding                      &  \ul{66}&	logic\\
17&	spatial reasoning                 &  \ul{67}&	arithmetic\\
18&	spatial reasoning                 &  68&	spatial reasoning\\
19&	path finding                      &  \ul{69}&	spatial reasoning\\
20&	measuring                         &  \ul{70}&	logic\\
21&	measuring                         &  \ul{71}&	arithmetic\\
22&	measuring                         &  \ul{72}&	algebra\\
23&	counting                          &  \ul{73}&	logic\\
24&	counting                          &  \ul{74}&	logic\\
25&	measuring                         &  \ul{75}&	spatial reasoning\\
26&	counting                          &  \ul{76}&	algebra\\
27&	algebra                           &  \ul{77}&	pattern\\
28&	algebra                           &  78&	spatial reasoning\\
29&	logic                             &  79&	counting\\
30&	arithmetic                              &  80&	counting\\
31&	algebra                           &  81&	logic\\
32&	pattern                           &  82&	spatial reasoning\\
33&	counting                          &  83&	pattern\\
34&	counting                          &  84&	pattern\\
35&	path finding                      &  85&	algebra\\
36&	logic                             &  86&	pattern\\
37&	path finding                      &  87&	logic\\
38&	arithmetic                              &  88&	arithmetic\\
39&	spatial reasoning                 &  89&	counting\\
40&	counting                          &  90&	algebra\\
41&	counting                          &  91&	logic\\
42&	counting                          &  92&	measuring\\
43&	arithmetic                              &  93&	measuring\\
44&	spatial reasoning                 &  \ul{94}&	measuring\\
45&	counting                          &  \ul{95}&	spatial reasoning\\
46&	arithmetic                              &  \ul{96}&	spatial reasoning\\
47&	arithmetic                              &  \ul{97}&	counting\\
48&	path finding                      &  \ul{98}&	arithmetic\\
49&	algebra                           &  \ul{99}&	counting\\
50&	counting                          &  100&	logic\\
                                     & &  \ul{101}&	algebra\\
\end{tabular}
    \caption{Root puzzle ids in \name and their  8 skill categories. Underlined are the puzzles used in PS and FS test sets.}
    \label{tab:puzzle_ids_and_skills}
\end{table}


\section{Heatmaps using guided GradCAM}
\label{sec:heatmaps}
\begin{figure}
    \centering
    \begin{subfigure}{0.45\linewidth}
    \includegraphics[width=3.5cm]{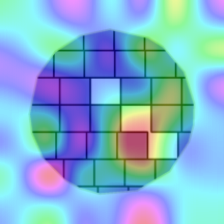}
    \caption{Puzzle \# 5 (C)}
     \end{subfigure}
    \begin{subfigure}{0.45\linewidth}
    \includegraphics[width=3.5cm]{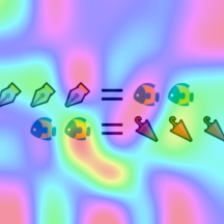}
    \caption{Puzzle \# 7 (AL)}
     \end{subfigure}
    \begin{subfigure}{0.45\linewidth}
    \includegraphics[width=3.5cm]{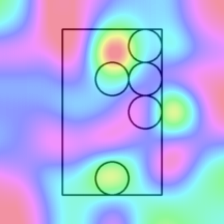}
    \caption{Puzzle \# 36 (L)}
     \end{subfigure}
    \begin{subfigure}{0.45\linewidth}
        \includegraphics[width=3.5cm]{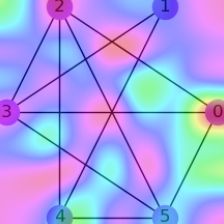}
        \caption{Puzzle \# 48 (PF)}
    \end{subfigure}
    \vspace*{5mm}
    \begin{subfigure}{0.45\linewidth}
        \includegraphics[width=3.5cm]{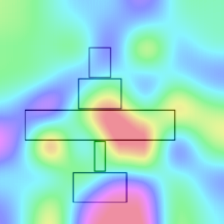}
        \caption{Puzzle \# 78 (L)}
     \end{subfigure}
    \begin{subfigure}{0.45\linewidth}
        \includegraphics[width=3.5cm]{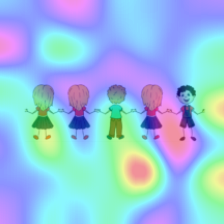}
        \caption{Puzzle \# 93 (C)}
       \end{subfigure}
    \caption{Heatmaps showing the spatial regions being activated when solving the puzzles. We show heatmaps on instances where the correct answer is produced by our method. The maps are generated using the ResNet-50 model trained using the IS split. We also show the class of the puzzles.}
    \label{fig:heatmaps}
\end{figure}

In Figure~\ref{fig:heatmaps}, we show the heatmaps of ResNet-50 activations when producing the correct answers to the respective puzzle instances. We use guided GradCAM for the visualizations where the features from the last convolutional layer (layer 7, conv3) are multiplied by the gradients. The red regions show the region of highest activations. We see that the regions of high temperature appear to those that are perhaps humans attend to when solving those puzzles. Note that precise interpretation of these heatmaps may be difficult as one needs to use the attended regions within algorithms that also use cues from language models. We believe such posthoc explanations for algorithmic reasoning models is a concrete future research direction.

\begin{table*}[t]
\centering
\footnotesize
\resizebox{1\textwidth}{!}{%
\begin{tabular}{p{5cm}p{5cm}p{5cm}p{5cm}}
\normalsize\hspace{10mm}(a) MK's root puzzle &\normalsize\hspace{4mm}(b) our generated instance \#1 &\normalsize\hspace{4mm}(c) our generated instance \#2 &\normalsize\hspace{4mm}(d) our generated instance \#3\\
\centering\includegraphics[width=4cm,height=4cm]{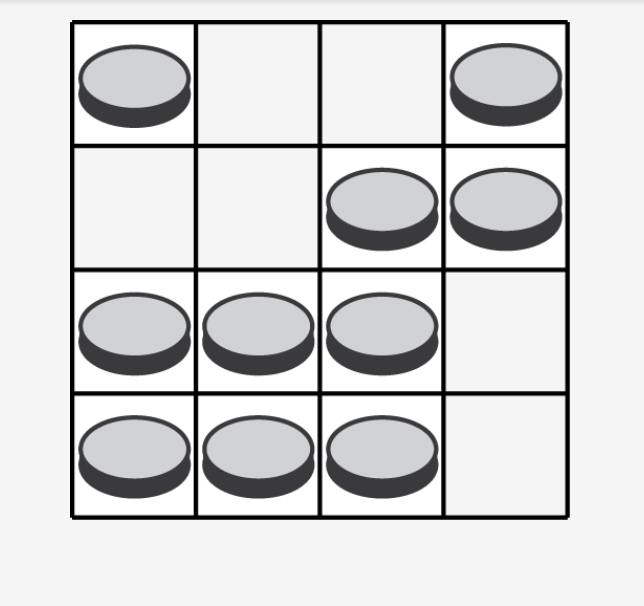} &\centering\includegraphics[width=4cm,height=4cm]{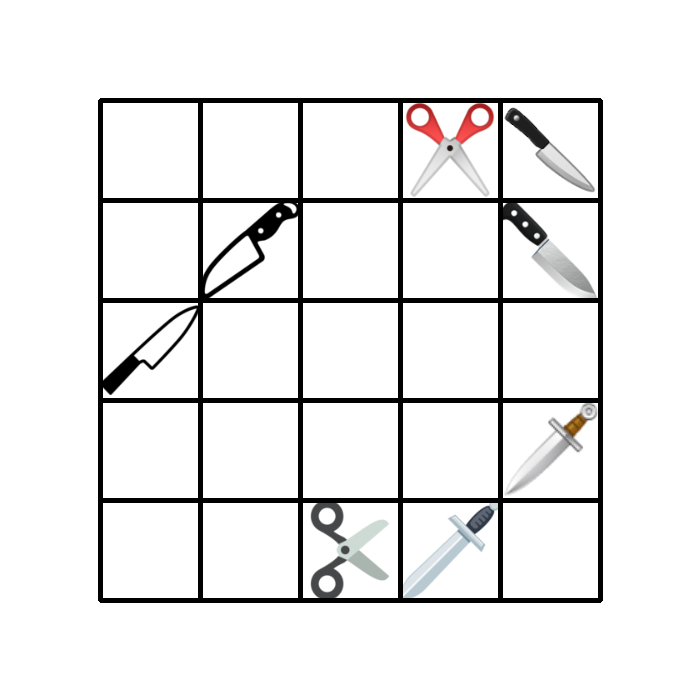} &\centering\includegraphics[width=4cm,height=4cm]{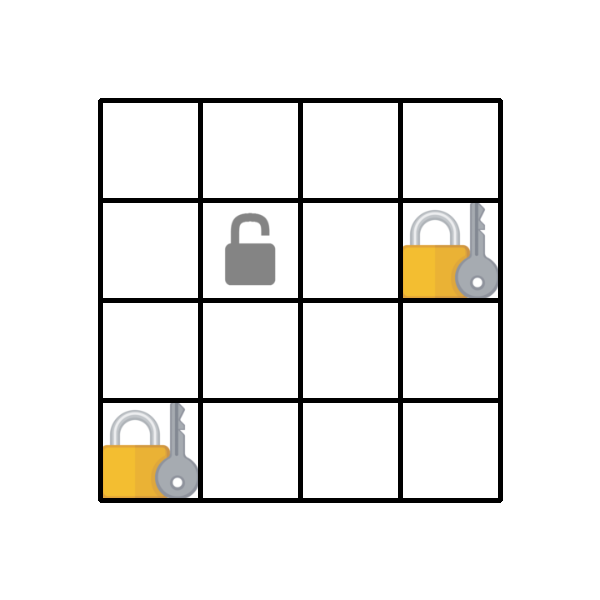} &\hspace{5mm}\includegraphics[width=4cm,height=4cm]{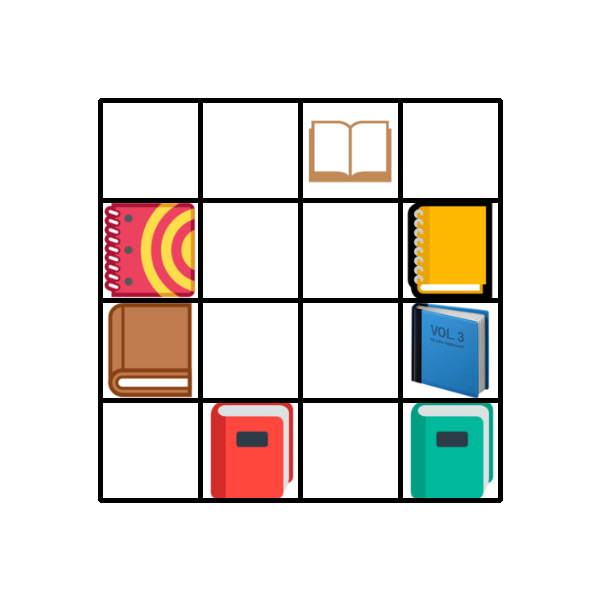}\\
\textbf{Question}: There are coins on the board. We want to have 2 coins in each column and 2 coins in each row. How many coins need to be removed? \textbf{Options}: A: 0, B: 1, \red{C: 2}, D: 3, E: 4 &\textbf{Question}: We want to have 1 blade in each column and 1 blade in each row on the board. The number of blades which need to be removed is: \textbf{Options}: A: 0, \red{B: 3}, C: 7, D: 5, E: 6 &\textbf{Question}: There are locks on the board. We want to have 2 locks in each column and 2 locks in each row. How many locks do we need to add? \textbf{Options}: A: 7, B: 9, C: 8, D: 6, \red{E: 5} &\textbf{Question}: There are books on the board. We want to have 1 book in each column and 1 book in each row. The number of books we need to remove is: \textbf{Options}: \red{A: 3}, B: 7, C: 0, D: 6, E: 5\\
\midrule
\centering\includegraphics[width=4cm,height=4cm]{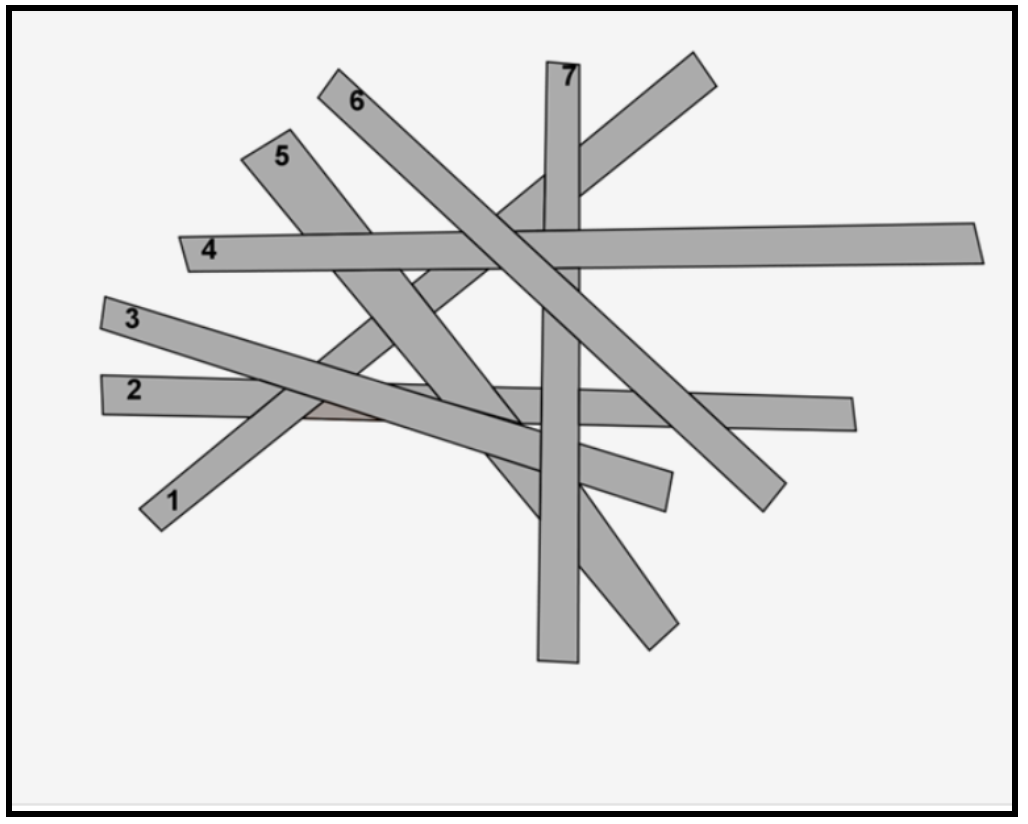} &\centering\includegraphics[width=4cm,height=4cm]{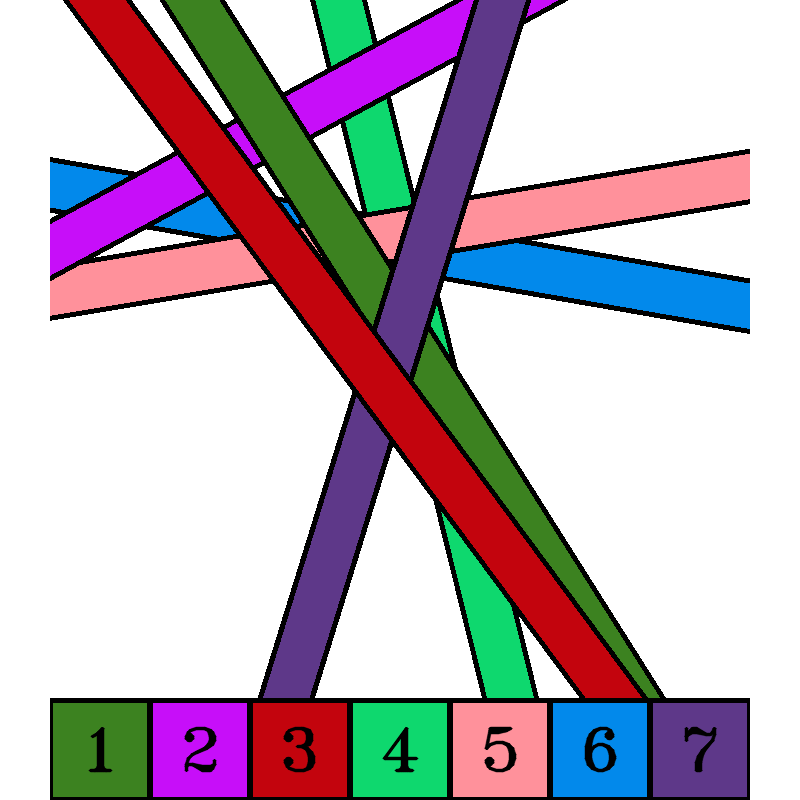} &\centering\includegraphics[width=4cm,height=4cm]{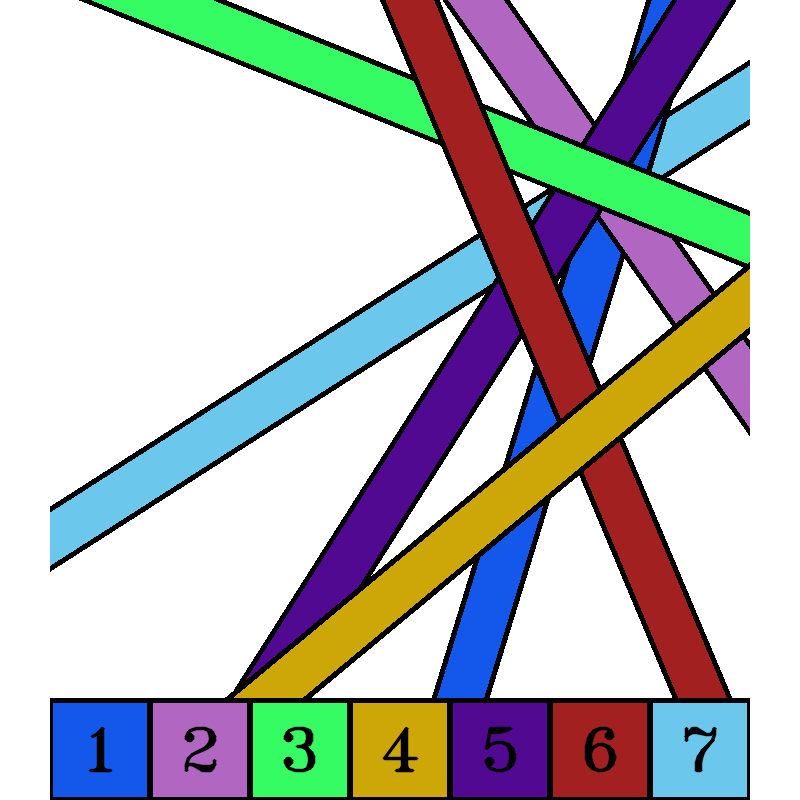} &\hspace{5mm}\includegraphics[width=4cm,height=4cm]{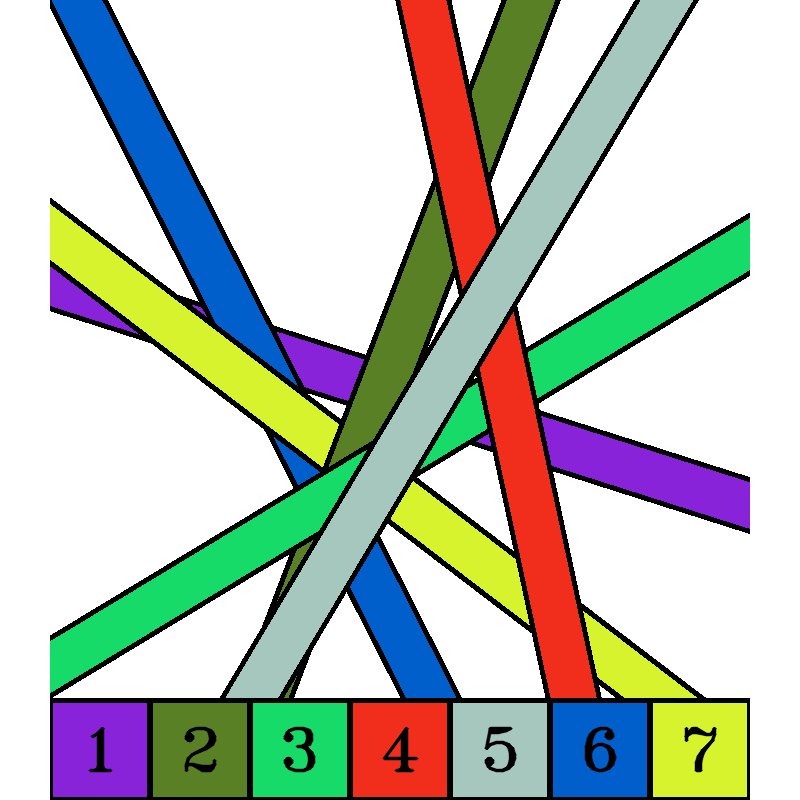}\\
\textbf{Question}: Seven sticks lie on top of each other. Stick 2/6 is at the bottom/top. Which stick is in the middle? \textbf{Options}: A: 1, \red{B: 3}, C: 4, D: 5, E: 7 &\textbf{Question}: The sticks lie on top of each other. Stick 6/3 is at the bottom/top. Which stick is in the middle? \textbf{Options}: A: 5, B: 7, \red{C: 2}, D: 1, E: 4 &\textbf{Question}: The sticks lie on top of each other. Stick 7/4 is at the bottom/top. Which stick is in the middle? \textbf{Options}: A: 6, B: 1, C: 2, \red{D: 5}, E: 3 &\textbf{Question}: The sticks are placed on top of each other. Stick 1/5 is at the bottom/top. Which stick is in the middle? \textbf{Options}: A: 3, \red{B: 2}, C: 4, D: 7, E: 6\\
\midrule
\centering\includegraphics[width=4cm,height=4cm]{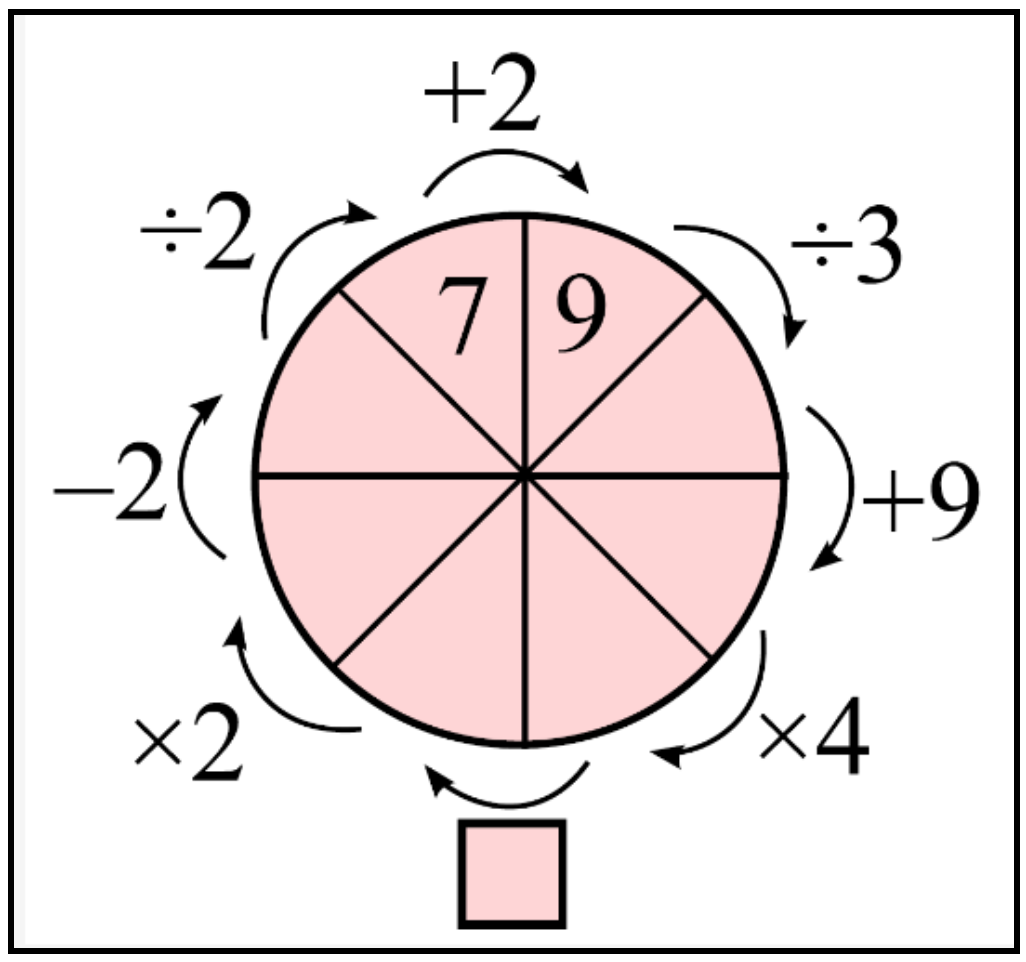} &\centering\includegraphics[width=4cm,height=4cm]{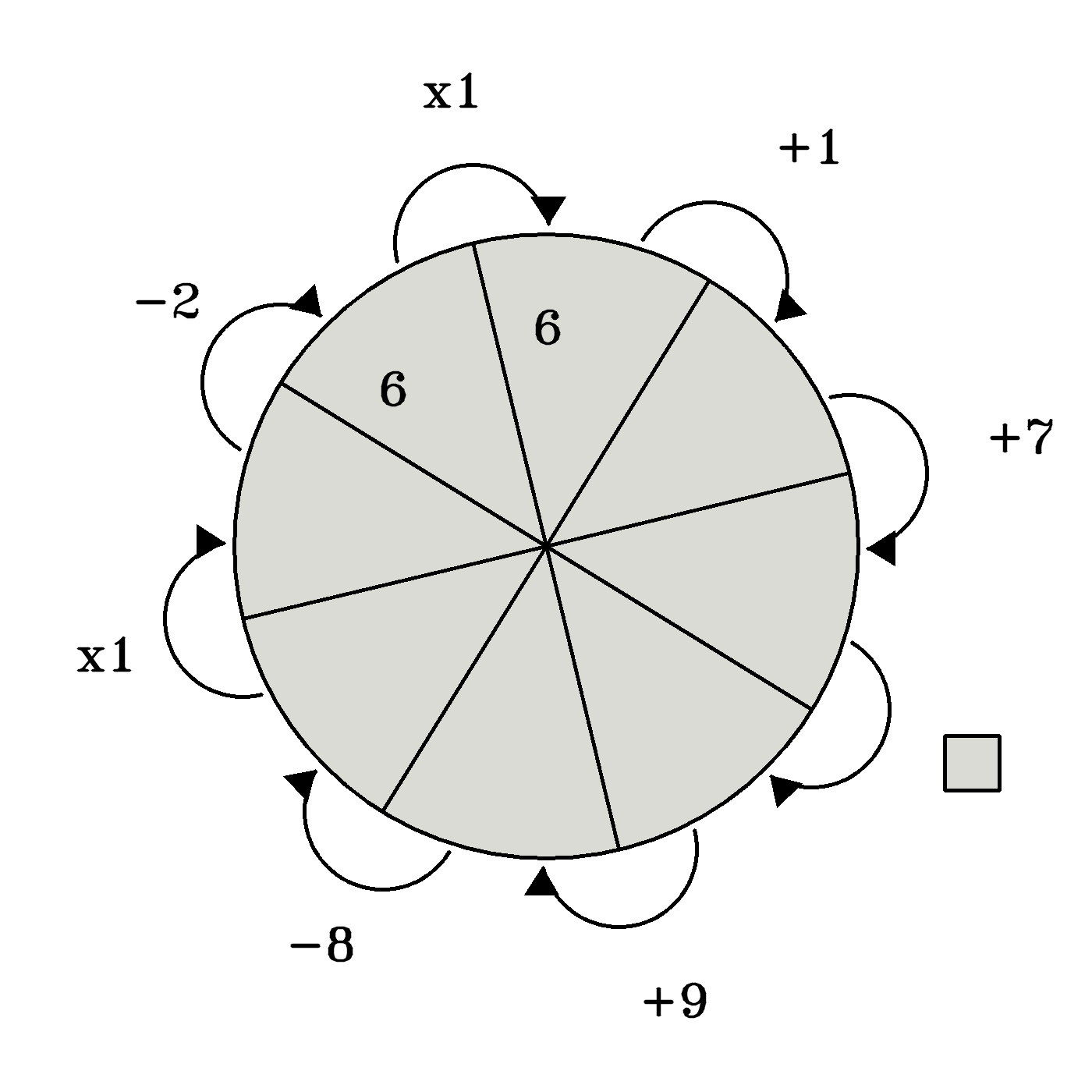} &\centering\includegraphics[width=4cm,height=4cm]{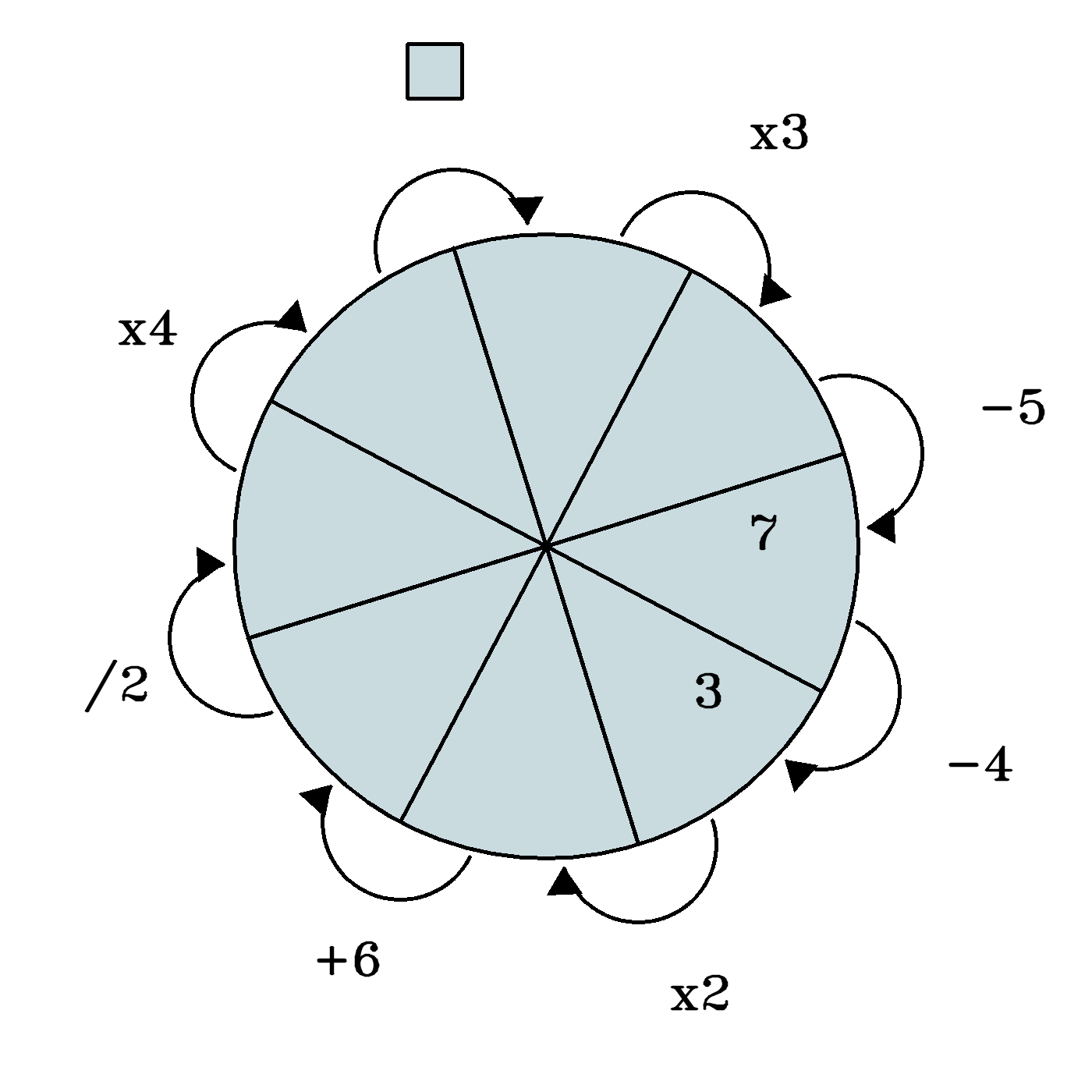} &\hspace{5mm}\includegraphics[width=4cm,height=4cm]{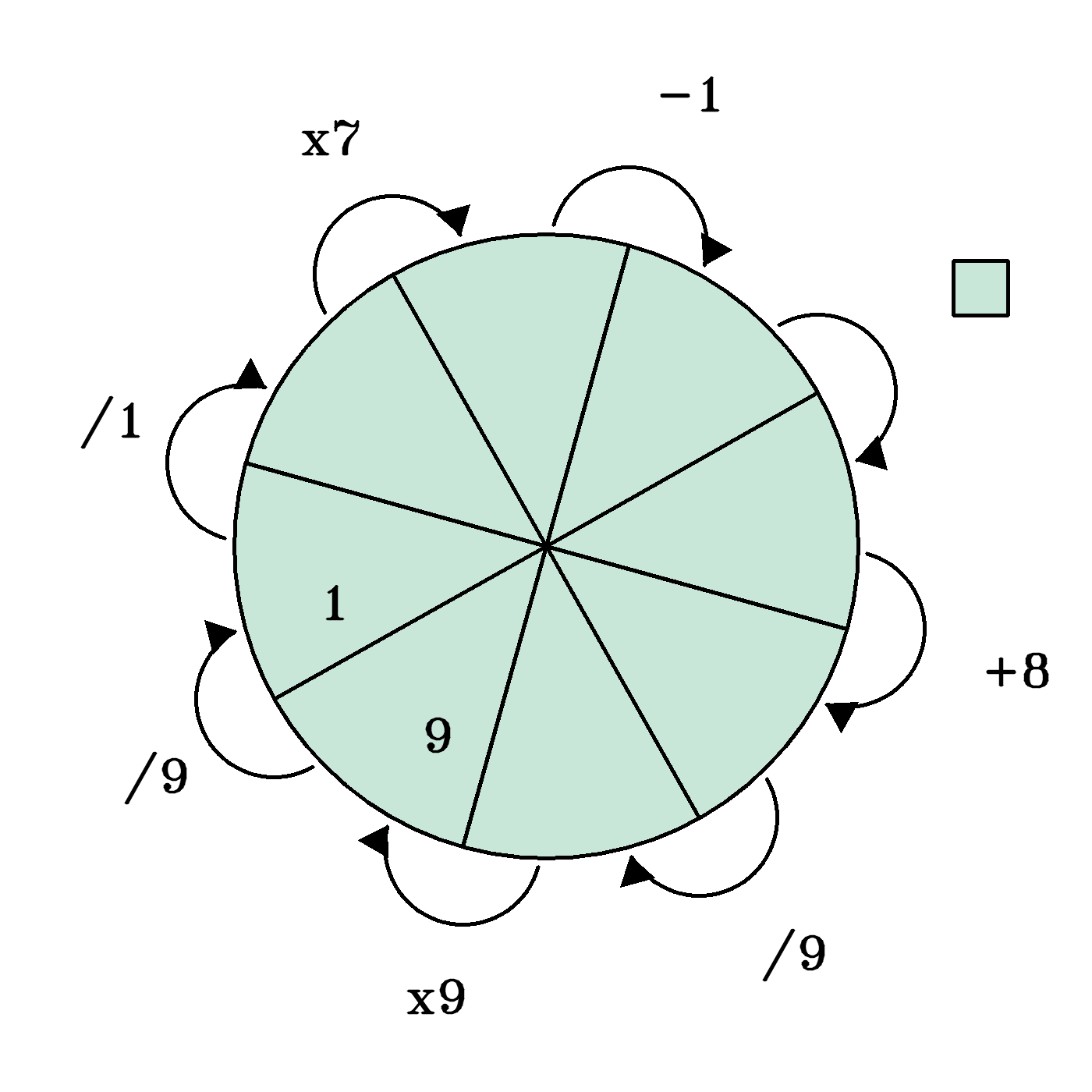}\\
\textbf{Question}: What should you put in the square to get a correct diagram? \textbf{Options}: A: -38, B: /8, C: -45, D: x6, \red{E: /6} &\textbf{Question}: What operation should you put in the square to get a correct diagram? \textbf{Options}: \red{A: -7}, B: /7, C: /4, D: /1, E: /9 &\textbf{Question}: What should you put in the square to get a correct diagram? \textbf{Options}: A: -6, B: +4, C: +9, \red{D: /6}, E: x5 &\textbf{Question}: What operation should be put in the square to get a correct diagram? \textbf{Options}: A: /1, B: +6, \red{C: -5}, D: -3, E: x7\\
\midrule
\includegraphics[width=4cm,height=4cm]{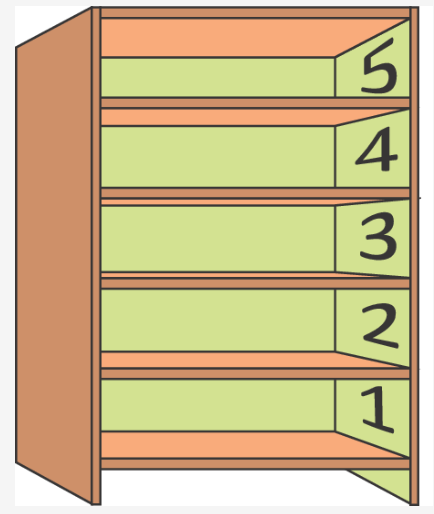} &\includegraphics[width=4cm,height=4cm]{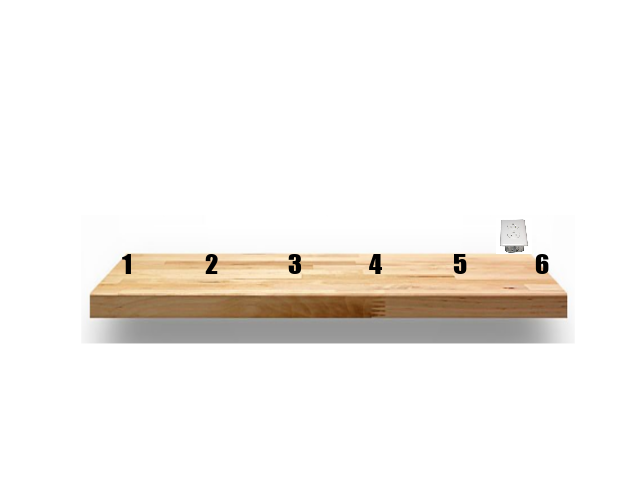} &\includegraphics[width=4cm,height=4cm]{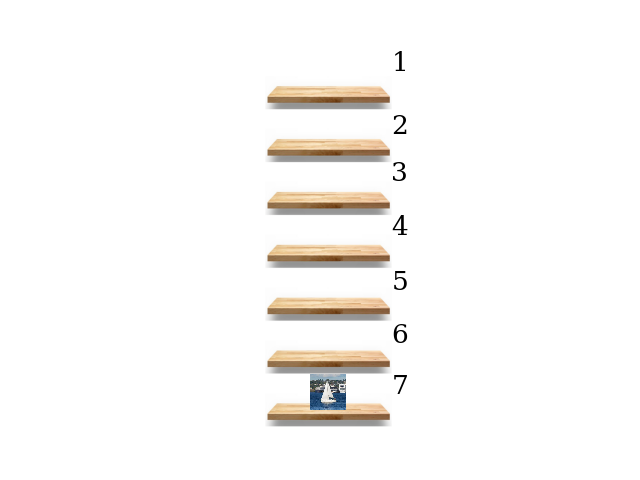} &\includegraphics[width=4cm,height=4cm]{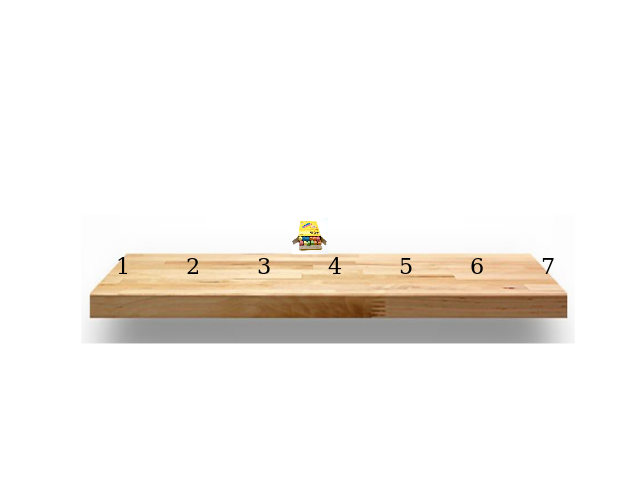}\\
\textbf{Question}: Stan has five toys: a ball, a set of blocks, a game, a puzzle, and a car. He puts each toy on a different shelf of the bookcase. The ball is higher than the blocks and lower than the car. The game is directly above the ball. On which shelf can the puzzle not be placed? \textbf{Options}: A: 1, B: 2, C: \red{3}, D: 4, E: 5 &\textbf{Question}: Stephen has 6 toy items: a photo album, a button, fake flowers, pants, a pencil, and an outlet. He puts each toy on a different line of the shelf. He puts outlet as shown. The photo album left of button. Outlet right of fake flowers and outlet right of pants. Pants right of button. On which line can the fake flowers not be placed? \textbf{Options}: A: 1, B: \red{6}, C: 2, D: 4, E: 3 &\textbf{Question}: Felicia has 7 toy items: a television, a sharpie, a checkbook, a chalk, tooth picks, twezzers, and a sailboat. She puts each toy on a different row of the shelf. The chalk higher to television and chalk higher to checkbook. Television lower to sailboat and television lower to tooth picks. Sharpie lower to tooth picks. She puts sailboat as shown. On which row can the sharpie not be placed? \textbf{Options}: A: 1, B: 4, C: \red{7}, D: 6, E: 2 &\textbf{Question}: Katherine has 7 toy objects: a balloon, a bed, a chalk, a tv, a model car, a cat, and a thread. She puts each toy on a different line of the shelf. She puts chalk as shown. The model car right of bed. Thread right of tv and thread right of model car. Cat right of model car. On which line can the tv not be placed? \textbf{Options}: A: 3, B: 5, C: \red{7}, D: 2, E: 1\\
\end{tabular}
}
\vspace{-2mm}
\caption{Examples of the template puzzles from Math Kangaroo USA~\cite{kangaroo} and our generated puzzle instances from the \name dataset. The correct answer of each puzzle is marked in \red{red}. The puzzle categories of these examples are (from top to bottom): counting, logic, spatial reasoning, and arithmetic.}
\label{tab:puzzle1}
\end{table*}

\begin{table*}[t]
\centering
\footnotesize
\resizebox{1\textwidth}{!}{%
\begin{tabular}{p{5cm}p{5cm}p{5cm}p{5cm}}
\normalsize\hspace{10mm}(a) MK's root puzzle &\normalsize\hspace{4mm}(b) our generated instance \#1 &\normalsize\hspace{4mm}(c) our generated instance \#2 &\normalsize\hspace{4mm}(d) our generated instance \#3\\
\centering\includegraphics[width=4cm,height=4cm]{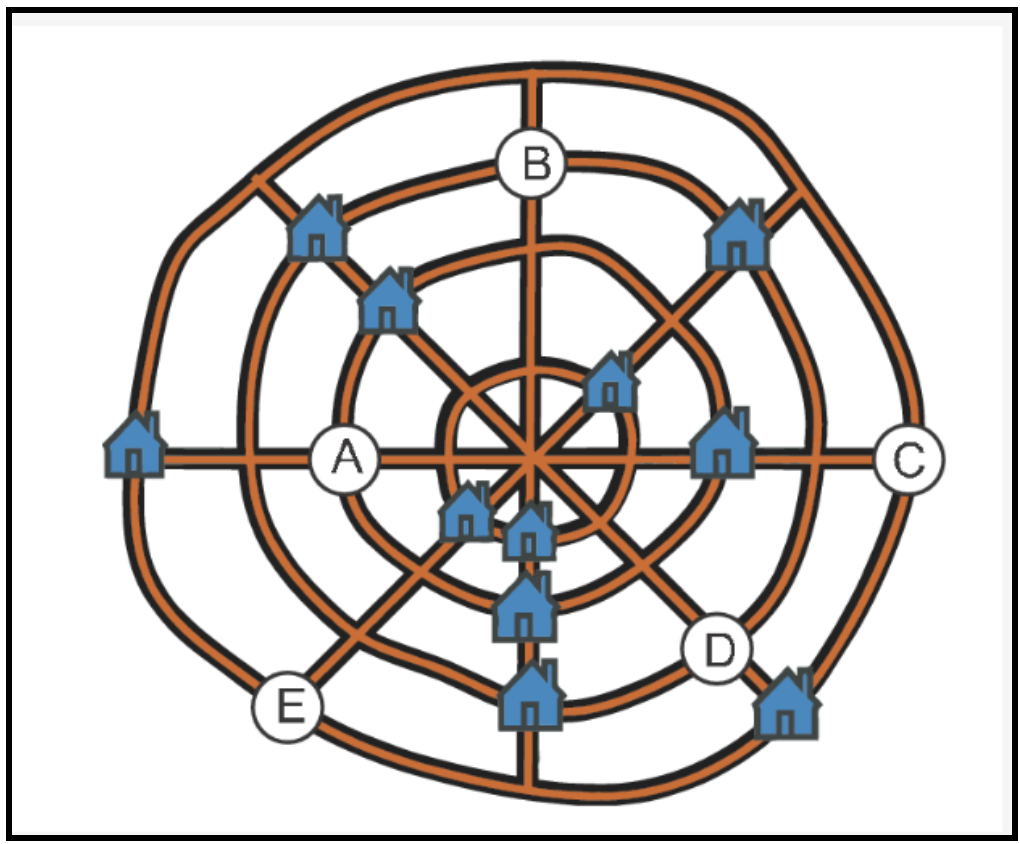}
&\centering\includegraphics[width=4cm,height=4cm,trim={2.8cm 2.8cm 2.8cm 2.8cm},clip]{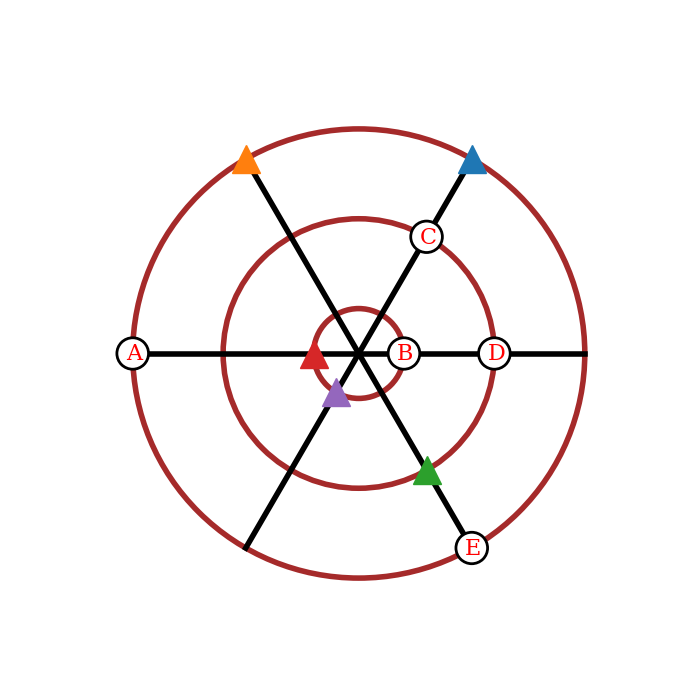}
&\centering\includegraphics[width=4cm,height=4cm,trim={2.8cm 2.8cm 2.8cm 2.8cm},clip]{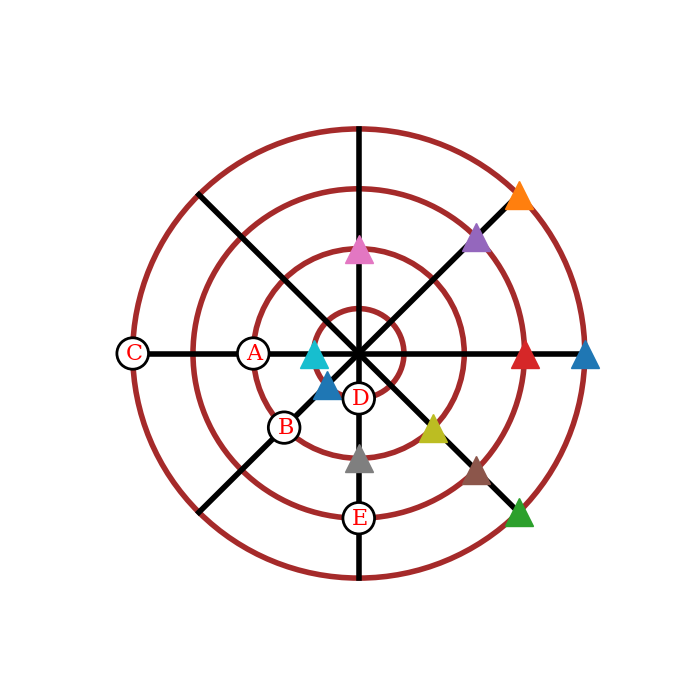} &\hspace{5mm}\includegraphics[width=4cm,height=4cm,trim={2.8cm 2.8cm 2.8cm 2.8cm},clip]{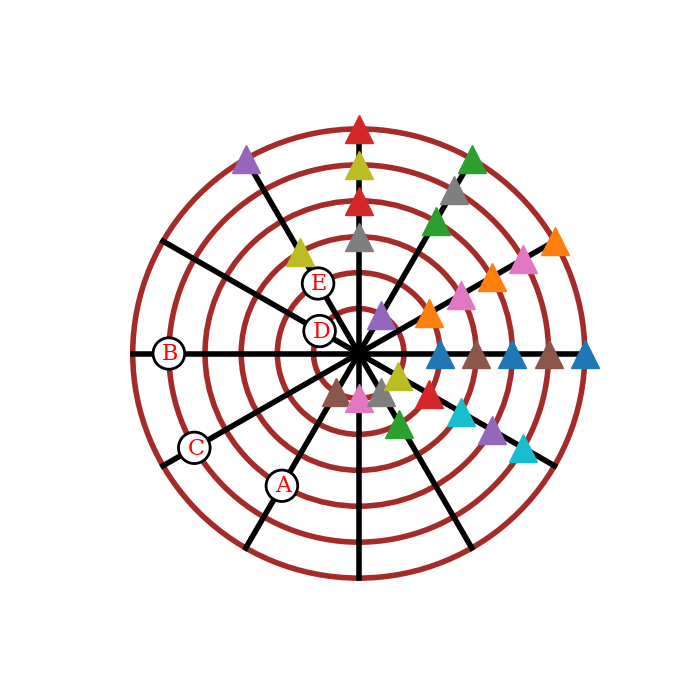}\\
\textbf{Question}: A village with 12 houses has four straight roads and four circular roads. The map shows 11 of the houses. On each straight road there are 3 houses. On each circular road, there are also 3 houses. Where on the map should the 12th house be put? \textbf{Options}: A, B, \red{C}, D, E &\textbf{Question}: A town with 6 houses has 3 straight pathways and 3 circular pathways. The image shows 5 of the houses. On each straight pathway there are 2 houses. On each circular pathway, there are also 2 houses. Which location on the image should the 6th house be built? \textbf{Options}: A, B, C, \red{D}, E &\textbf{Question}: A small town with 12 huts has 4 straight lanes and 4 circular lanes. The map depicts 11 of the huts. On each straight lane there are 3 huts. On each circular lane, there are also 3 huts. Which location on the map should the 12th hut be put? \textbf{Options}: A, B, C, \red{D}, E &\textbf{Question}: A community with 30 condos has 6 straight paths and 6 circular paths. The picture illustrates 29 of the condos. On each straight path there are 5 condos. On each circular path, there are also 5 condos. Which place on the picture should the 30th condo be added? \textbf{Options}: A, B, C, D, \red{E}\\
\midrule
\includegraphics[width=4cm,height=4cm]{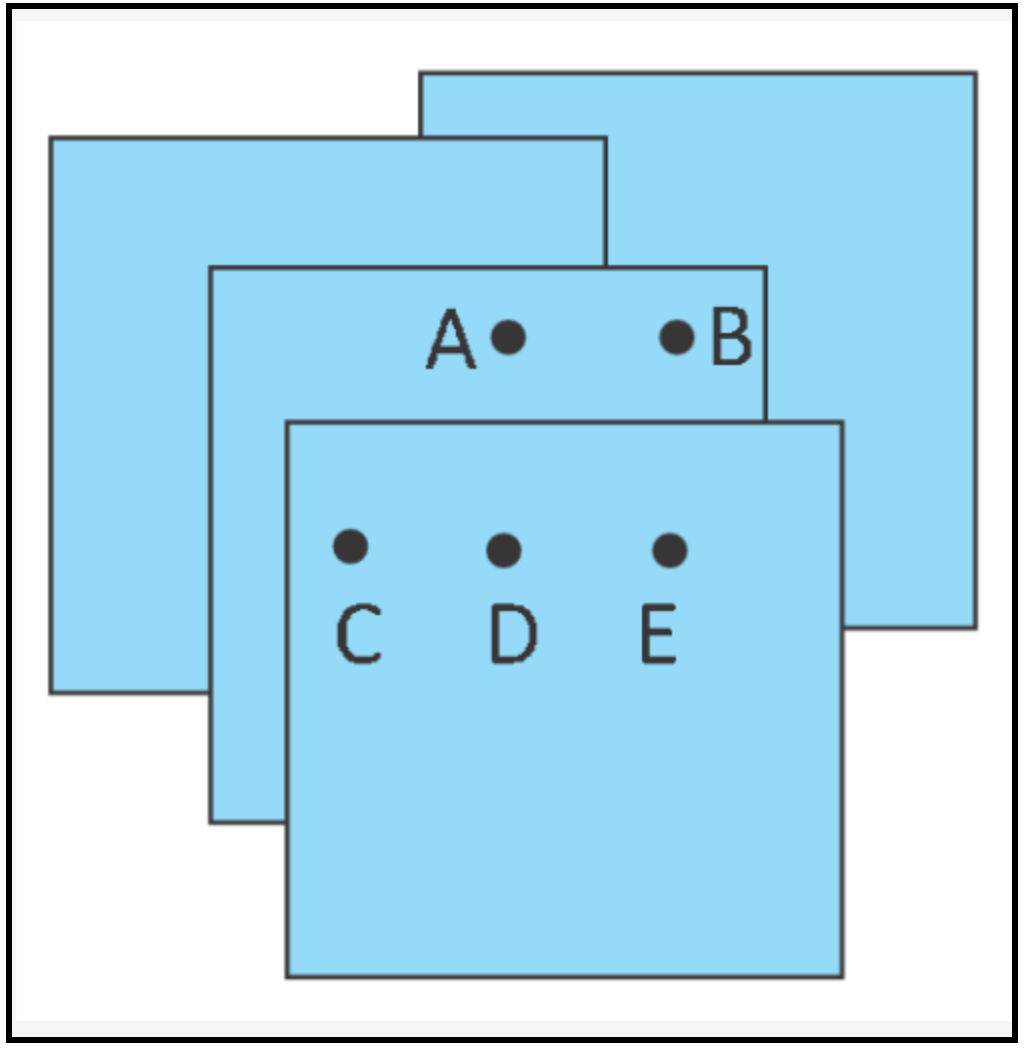} &\includegraphics[width=4cm,height=4cm]{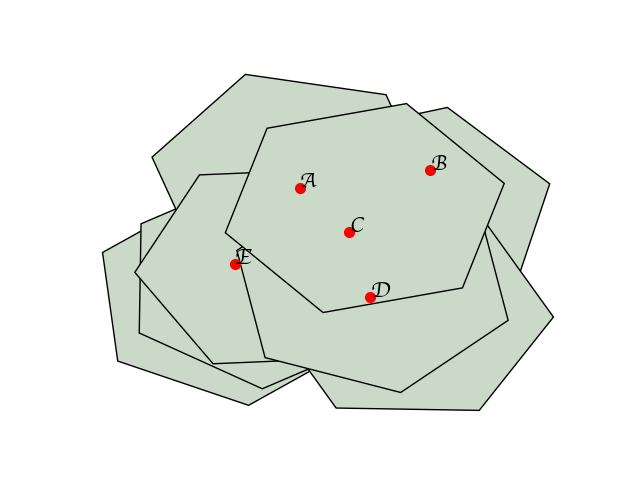} &\includegraphics[width=4cm,height=4cm]{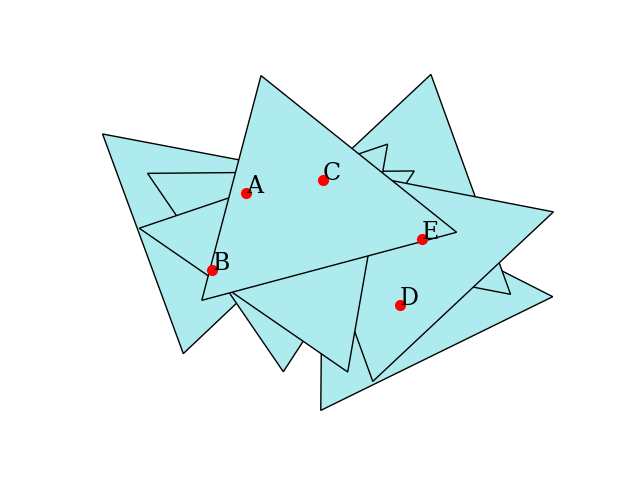} &\includegraphics[width=4cm,height=4cm]{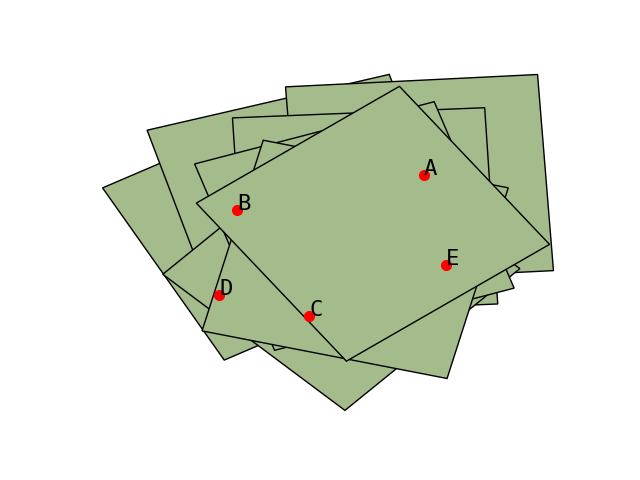}\\
\textbf{Question}: Four identical pieces of paper are placed as shown. Michael wants to punch a hole that goes through all four pieces. At which point should Michael punch the hole? \textbf{Options}: A, B, C, \red{D}, E &\textbf{Question}: Eight identical pieces of sheets are kept as shown in the picture. Bridget have to to drill a hole that goes through all eight pieces. What point should she drill the hole? \textbf{Options}: A, B, \red{C}, D, E &\textbf{Question}: There are seven equivalent parts of paper arranged as displayed. Daniel wants to punch a hole that passes through all seven parts. At which location must Daniel punch the hole? \textbf{Options}: A, B, \red{C}, D, E &\textbf{Question}: Eight exactly same pieces of paper are fixed as shown. Nathan needs to punch a hole that passes through all eight pieces. What position must he punch the hole? \textbf{Options}: A, B, C, D, \red{E}\\
\midrule
\includegraphics[width=4cm,height=4cm]{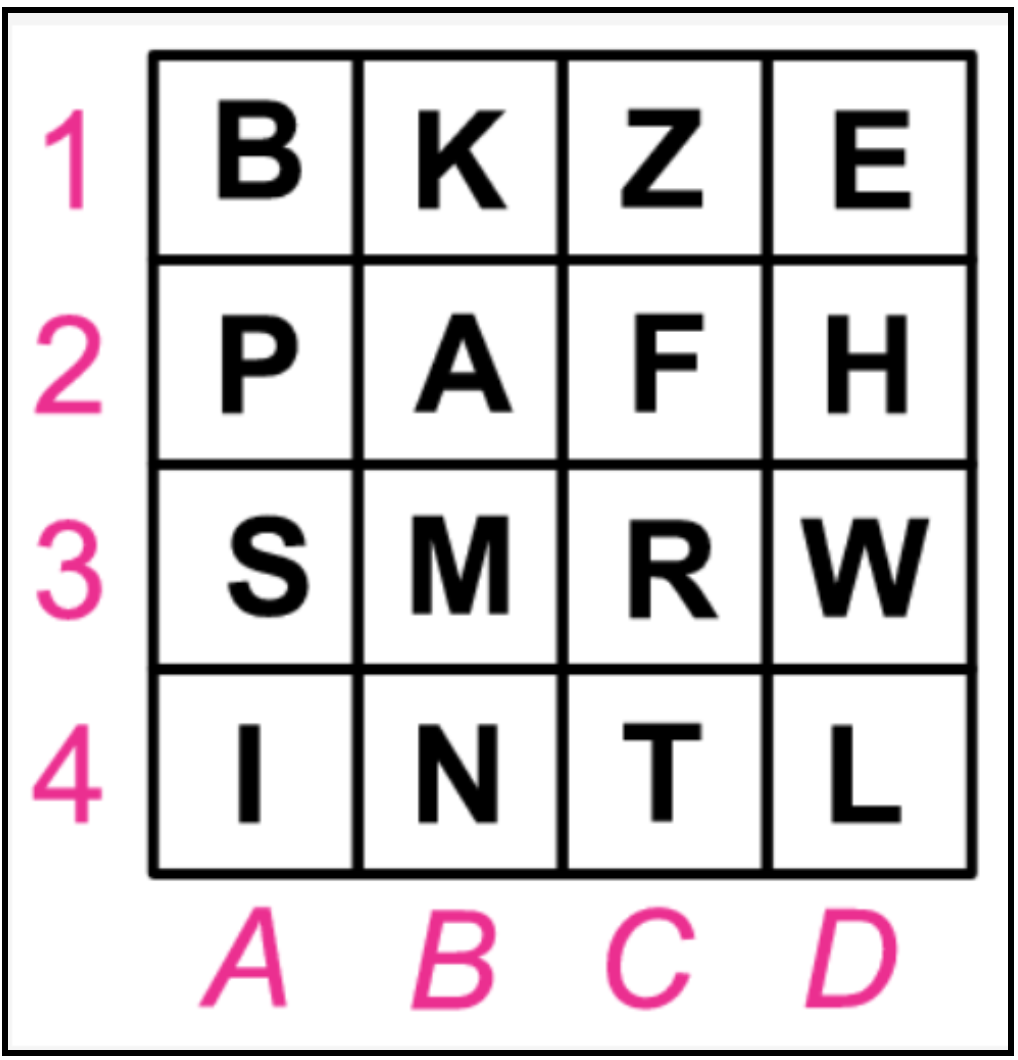} &\includegraphics[width=4cm,height=4cm]{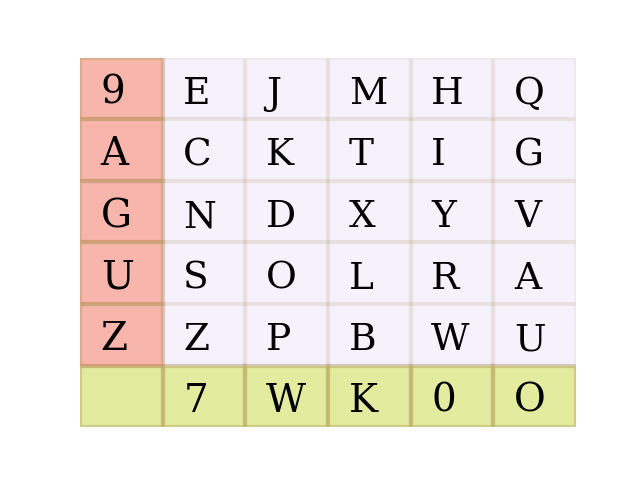} &\includegraphics[width=4cm,height=4cm]{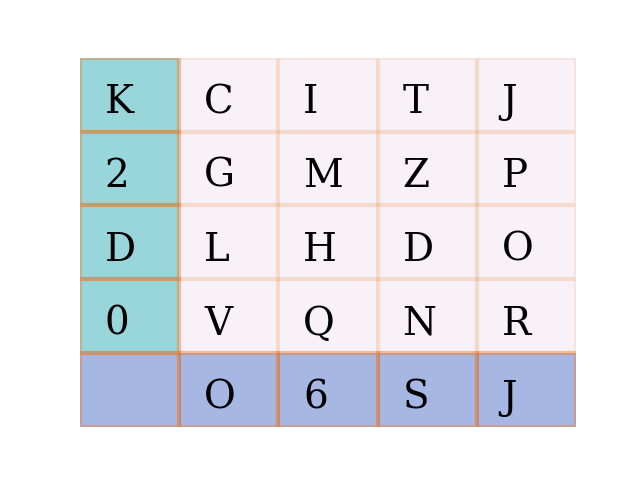} &\includegraphics[width=4cm,height=4cm]{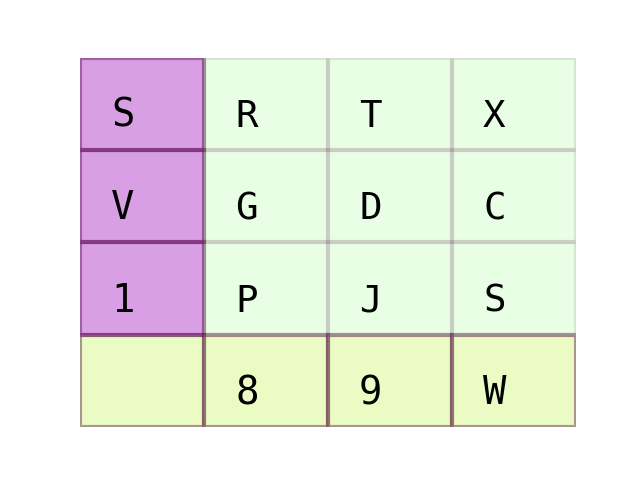}\\
\textbf{Question}: Tom encodes words using the board shown. For example, the word PIZZA has the code A2 A4 C1 C1 B2. What word did Tom encode as B3 B2 C4 D2? \textbf{Options}: A: MAZE, B: MASK, C: MILK, D: MATE, E: \red{MATH} &\textbf{Question}: Jasmine encrypts words as the matrix demonstrated. For example, the word UNION is encrypyed as ZO G7 A0 UW G7. What word did Jasmine encrypt 97 UO U0 UK G0? \textbf{Options}: A: \red{EARLY}, B: TURVY, C: LORRE, D: CLATS, E: YEEHA &\textbf{Question}: Anthony captures words applying the matrix presented. For an illustration, the word OMIT is captured as DJ 26 K6 KS. What word did Anthony capture DO DJ 0J DS? \textbf{Options}: A: YOIT, B: ACTU, C: \red{LORD}, D: XLNT, E: BAHO &\textbf{Question}: Michelle encodes words adopting the grid depicted. For an example, the word CDS is encoded as VW V9 1W. What word did Michelle encode 19 1W S9? \textbf{Options}: A: \red{JST}, B: SRI, C: ARF, D: OYE, E: URU\\
\midrule
\centering\includegraphics[width=4cm,height=4cm]{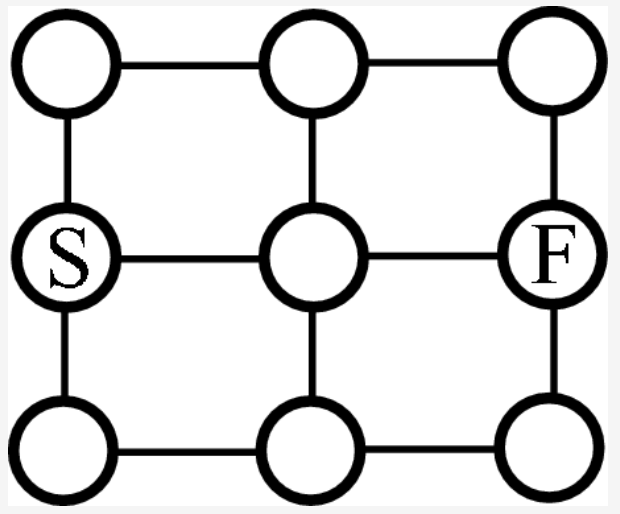} &\centering\includegraphics[width=4cm,height=4cm]{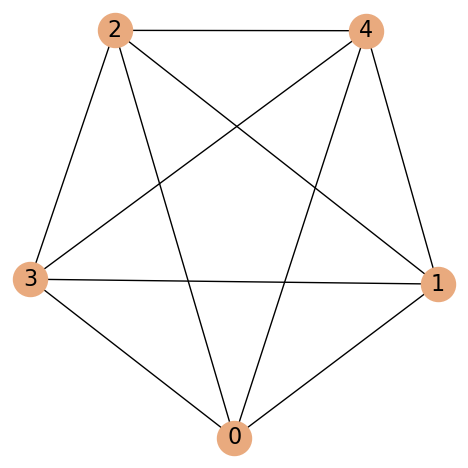} &\centering\includegraphics[width=4cm,height=4cm]{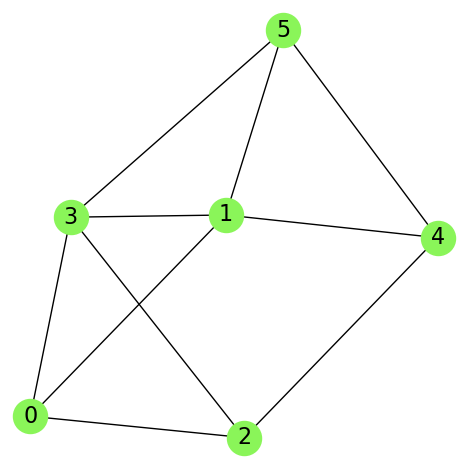} &\hspace{5mm}\includegraphics[width=4cm,height=4cm]{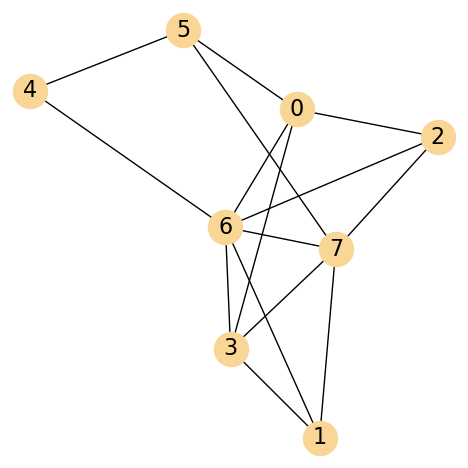}\\
\textbf{Question}: In one jump, Jake jumps from one circle to the neighboring circle along a line, as shown in the picture. He cannot jump into any circle more than once. He starts at circle S and needs to make exactly 4 jumps to get to circle F. In how many different ways can Jake do this? \textbf{Options}: A: 3, B: 4, C: 5, \red{D: 6}, E: 7 &\textbf{Question}: In one jump, Pamela jumps from one circle to the neighboring circle along a line, as shown in the picture. She cannot jump into any circle more than once. She starts at circle 2 and needs to make exactly 4 jumps to get to circle 0. In how many different ways can she do this? \textbf{Options}: \red{A: 6}, B: 11, C: 2, D: 10, E: 0 &\textbf{Question}: In one jump, Louis jumps from one circle to the neighboring circle along a line, as shown in the picture. He cannot jump into any circle more than once. He starts at circle 4 and needs to make exactly 2 jumps to get to circle 1. In how many different ways can Louis do this? \textbf{Options}: A: 3, B: 2, C: 0, \red{D: 1}, E: 6 &\textbf{Question}: In one jump, Chris jumps from one circle to the neighboring circle along a line, as shown in the picture. He cannot jump into any circle more than once. He starts at circle 1 and needs to make exactly 7 jumps to get to circle 6. In how many different ways can Chris do this? \textbf{Options}: A: 10, B: 8, C: 7, \red{D: 2}, E: 1\\
\end{tabular}
}
\vspace{-2mm}
\caption{Examples of the template puzzles from Math Kangaroo USA~\cite{kangaroo} and our generated puzzle instances from the \name dataset. The correct answer of each puzzle is marked in \red{red}. The puzzle categories of these examples are (from top to bottom): logic, spatial reasoning, and logic.}
\label{tab:puzzle2}
\end{table*}

\input{chatgpt/chatgpt_summary.tex}

{\small
\bibliographystyle{ieee_fullname}
\bibliography{egbib}
}

\end{document}

%% file: definitions.tex
\newcommand{\red}[1]{\textcolor{red}{#1}}
\newcommand{\blue}[1]{\textcolor{blue}{#1}}
\newcommand{\gray}[1]{\textcolor{black}{#1}}
\newcommand{\green}[1]{\textcolor{ForestGreen}{#1}}
\newcommand{\name}{SMART-101\xspace}
\newcommand{\shortname}{SMART\xspace}
\newcommand{\fullname}{Simple Multimodal Algorithmic Reasoning Task\xspace}

\def\eg{\emph{e.g}.}
\def\ie{\emph{i.e}.}
\def\etc{\emph{etc}.}

\newcommand{\dataset}{\mathcal{D}}
\newcommand{\img}{I}
\newcommand{\quest}{Q}
\newcommand{\ansset}{\mathcal{A}}
\newcommand{\ans}{a}
\newcommand{\model}{f}
\newcommand{\rp}{\pi}
\newcommand{\rootdataset}{\mathcal{R}}
\newcommand{\puzzset}{\mathcal{P}_\rp}
\newcommand{\puzinst}{p^{\rp}}

\newcommand{\imagehead}{g_\alpha}
\newcommand{\lanhead}{\ell_\beta}
\newcommand{\imgpuzzhead}{h^{\rp}_\gamma}

\DeclareMathOperator{\PRED}{pred}
\DeclareMathOperator{\FUSION}{fuse}
\newcommand{\pred}{\PRED^{\rp}_{\zeta}}
\newcommand{\fusion}{\FUSION_{\nu}}

\DeclareMathOperator{\simi}{sim}
\newcommand{\similarity}{\simi_\rp}
\DeclareMathOperator*{\acc}{acc}
\newcommand{\oacc}{O_{\acc}}
\newcommand{\sacc}{S_{\acc}}

\newcommand{\set}[1]{\left\{{#1}\right\}}
\newcommand{\tuple}[1]{\left(#1\right)}
\DeclareMathOperator*{\argmax}{\arg\max}
\newcommand{\expect}{\mathbb{E}}

\DeclareMathOperator{\loss}{loss}

\newcommand{\comment}[1]{}
\newcommand{\ul}[1]{\underline{#1}}

%% file: example_puzzles.tex
\begin{table*}[t]
\centering
\vspace{-.7em}
\footnotesize
\resizebox{.997\textwidth}{!}{
\begin{tabular}{p{5cm}p{5cm}p{5cm}p{5cm}}
\normalsize\hspace{10mm}(a) MK's root puzzle &\normalsize\hspace{4mm}(b) our generated instance \#1 &\normalsize\hspace{4mm}(c) our generated instance \#2 &\normalsize\hspace{4mm}(d) our generated instance \#3\\

\centering\includegraphics[width=4cm,height=4cm]{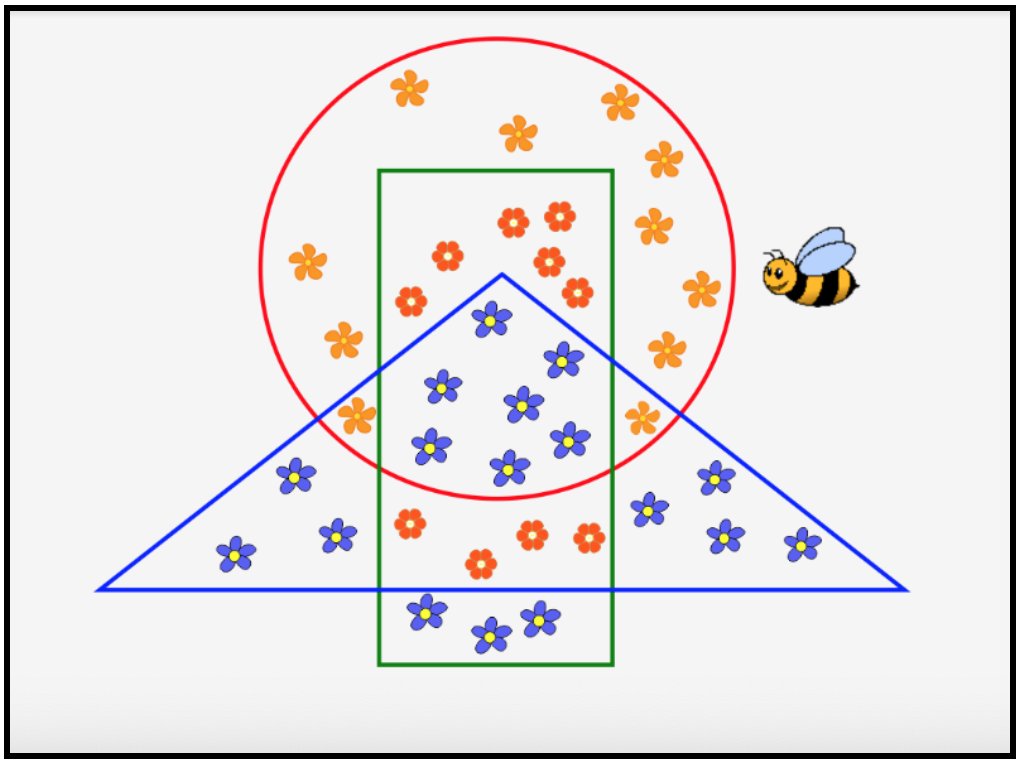} &\centering\includegraphics[width=4cm,height=4cm]{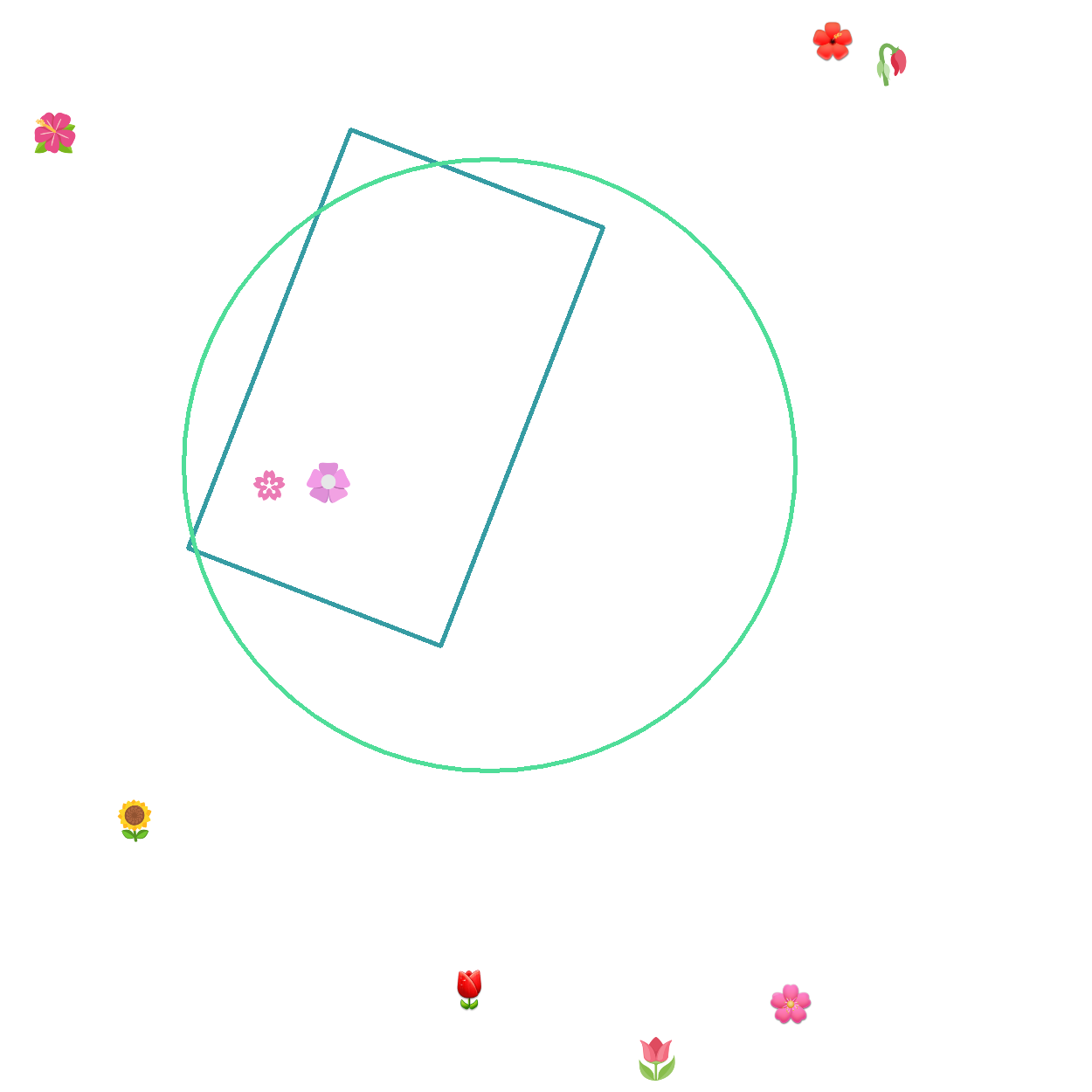} &\centering\includegraphics[width=4cm,height=4cm]{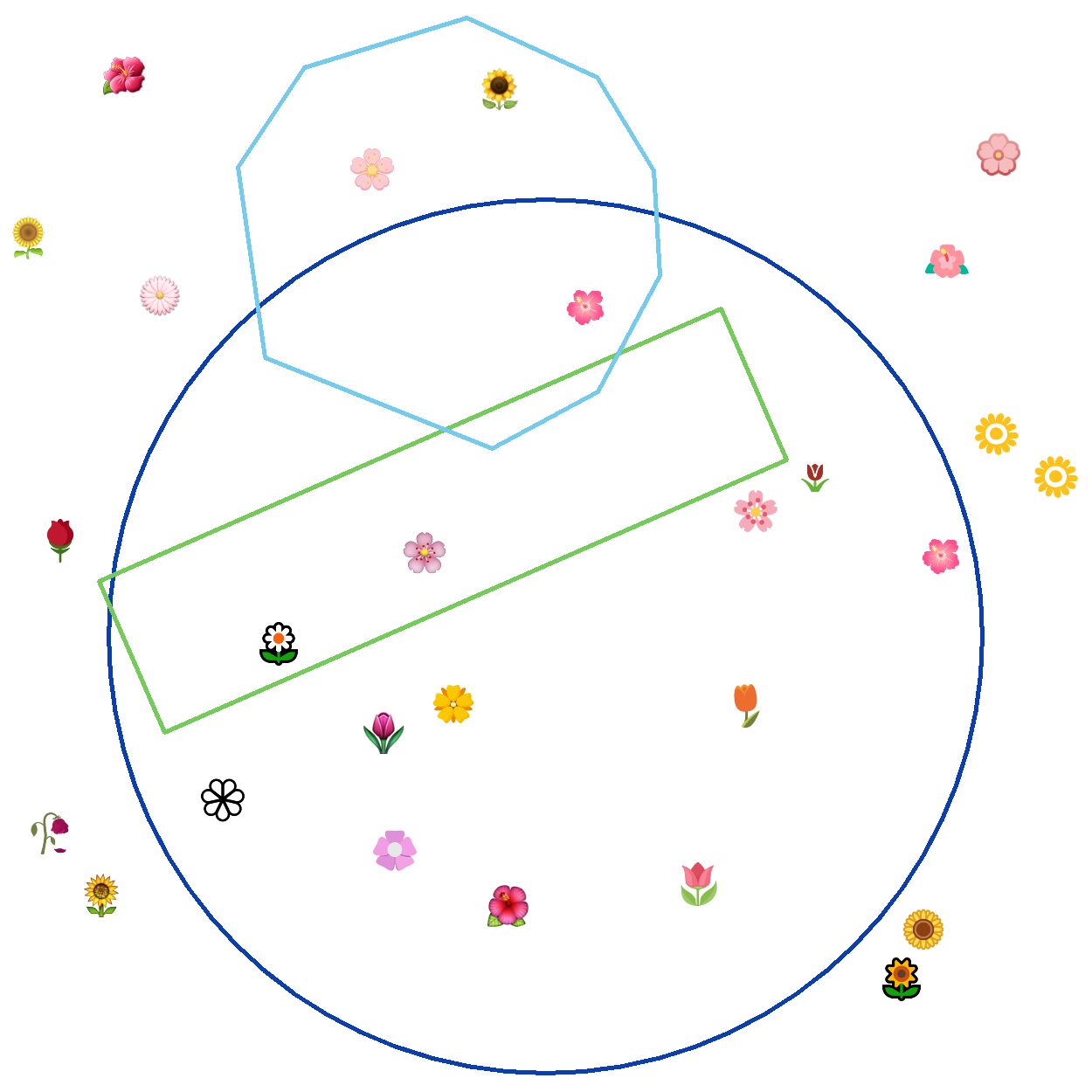} &\hspace{5mm}\includegraphics[width=4cm,height=4cm]{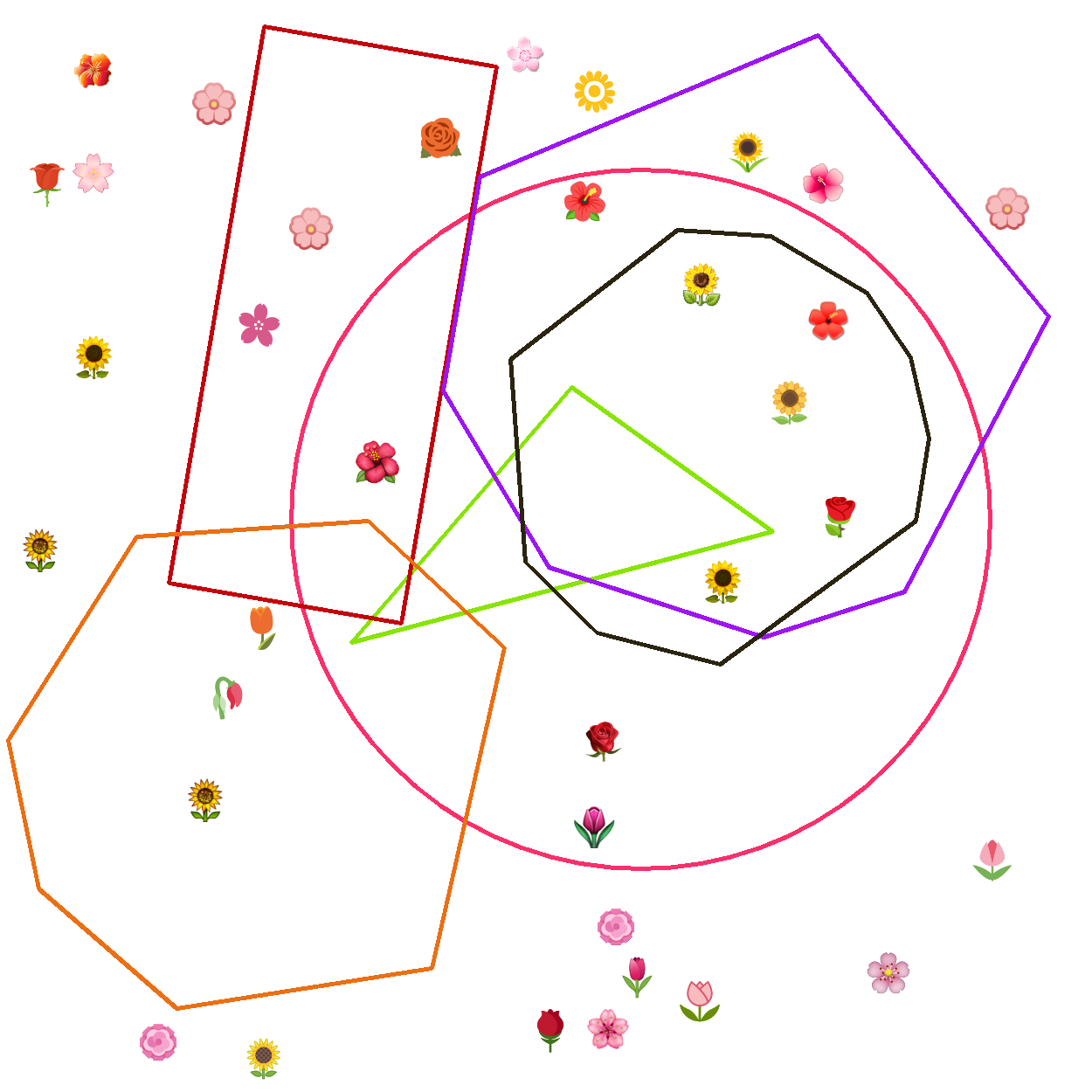}\\
\textbf{Question}: A bee collected pollen from all the flowers inside the rectangle but outside the triangle. From how many flowers did the bee collect pollen? \textbf{Options}: \red{A: 9}, B: 10, C: 13, D: 17, E: 20 &\textbf{Question}: We want to pick up all the flowers that are inside the rectangle and inside the circle simultaneously. How many flowers should we pick up? \textbf{Options}: A: 5, B: 6, \red{C: 2}, D: 1, E: 3 &\textbf{Question}: We want to pick up all the flowers that are inside the circle but outside the rectangle simultaneously. How many flowers should we pick up? \textbf{Options}: A: 7, B: 14, C: 15, D: 9, \red{E: 11} &\textbf{Question}: All the flowers that are outside both the circle and triangle simultaneously are picked up. The number of flowers which are picked up is: \textbf{Options}: A: 27, B: 24, \red{C: 26}, D: 29, E: 23\\
\midrule
\centering\includegraphics[width=4cm,height=4cm]{./figs/puzzles/p73.png}
&\centering\includegraphics[width=4cm,height=4cm,trim={2.8cm 2.8cm 2.8cm 2.8cm},clip]{./figs/instances/73/puzzle_73_2.png}
&\centering\includegraphics[width=4cm,height=4cm,trim={2.8cm 2.8cm 2.8cm 2.8cm},clip]{./figs/instances/73/puzzle_73_0.png} &\hspace{5mm}\includegraphics[width=4cm,height=4cm,trim={2.8cm 2.8cm 2.8cm 2.8cm},clip]{./figs/instances/73/puzzle_73_1.png}\\
\textbf{Question}: A village with 12 houses has four straight roads and four circular roads. The map shows 11 of the houses. On each straight road there are 3 houses. On each circular road, there are also 3 houses. Where on the map should the 12th house be put? \textbf{Options}: A, B, \red{C}, D, E &\textbf{Question}: A town with 6 houses has 3 straight pathways and 3 circular pathways. The image shows 5 of the houses. On each straight pathway there are 2 houses. On each circular pathway, there are also 2 houses. Which location on the image should the 6th house be built? \textbf{Options}: A, B, C, \red{D}, E &\textbf{Question}: A small town with 12 huts has 4 straight lanes and 4 circular lanes. The map depicts 11 of the huts. On each straight lane there are 3 huts. On each circular lane, there are also 3 huts. Which location on the map should the 12th hut be put? \textbf{Options}: A, B, C, \red{D}, E &\textbf{Question}: A community with 30 condos has 6 straight paths and 6 circular paths. The picture illustrates 29 of the condos. On each straight path there are 5 condos. On each circular path, there are also 5 condos. Which place on the picture should the 30th condo be added? \textbf{Options}: A, B, C, D, \red{E}\\
\midrule
\centering\includegraphics[width=4cm,height=4cm]{./figs/puzzles/p48.png} &\centering\includegraphics[width=4cm,height=4cm]{./figs/instances/48/puzzle_48_8.png} &\centering\includegraphics[width=4cm,height=4cm]{./figs/instances/48/puzzle_48_12.png} &\hspace{5mm}\includegraphics[width=4cm,height=4cm]{./figs/instances/48/puzzle_48_55.png}\\
\textbf{Question}: In one jump, Jake jumps from one circle to the neighboring circle along a line, as shown in the picture. He cannot jump into any circle more than once. He starts at circle S and needs to make exactly 4 jumps to get to circle F. In how many different ways can Jake do this? \textbf{Options}: A: 3, B: 4, C: 5, \red{D: 6}, E: 7 &\textbf{Question}: In one jump, Pamela jumps from one circle to the neighboring circle along a line, as shown in the picture. She cannot jump into any circle more than once. She starts at circle 2 and needs to make exactly 4 jumps to get to circle 0. In how many different ways can she do this? \textbf{Options}: \red{A: 6}, B: 11, C: 2, D: 10, E: 0 &\textbf{Question}: In one jump, Louis jumps from one circle to the neighboring circle along a line, as shown in the picture. He cannot jump into any circle more than once. He starts at circle 4 and needs to make exactly 2 jumps to get to circle 1. In how many different ways can Louis do this? \textbf{Options}: A: 3, B: 2, C: 0, \red{D: 1}, E: 6 &\textbf{Question}: In one jump, Chris jumps from one circle to the neighboring circle along a line, as shown in the picture. He cannot jump into any circle more than once. He starts at circle 1 and needs to make exactly 7 jumps to get to circle 6. In how many different ways can Chris do this? \textbf{Options}: A: 10, B: 8, C: 7, \red{D: 2}, E: 1\\
\end{tabular}
}
\vspace{-2mm}
\caption{Examples of the root puzzles (left) from the Math Kangaroo Olympiad~\cite{kangaroo} and our generated puzzle instances, belonging to categories: counting (top), logic (middle), and path tracing (bottom). The answer is marked in \red{red}. See supplement for more examples.}
\label{tab:puzzle_samples}
\end{table*}

%% file: chatgpt/chatgpt_summary.tex
\section{LLMs' responses on \name puzzles}
\label{sec:chatgpt_responses}
In this section, we provide the question prompts to our investigated large language models (LLMs) and two of ten selected responses (mostly one correct response and one incorrect response). We compute accuracy as the number of correct responses (that is correct selection answer option from the five choices) against the number of trials. We test the following root puzzles from the Math Kangaroo USA root puzzles: 7, 9, 30, 38, 47, 71, 88, 89, 90, 91, and 93, on the LLMs. For each puzzle, we provide the text of the puzzle and the options as input to the LLMs, and use their response (\ie, selection of the option) for evaluation. As the LLMs' response varies on each attempt, we repeat each input 10 times, and compute the average accuracy among all the trials. Specifically, for each attempt, we reset and start a new conversation such that the LLM’s output is not affected by the previous chat history. During the entire experiment, we give no feedback to each LLM after providing the input (question text and the answer options). Below, we provide the inputs and two selected outputs from each of our investigated LLM. We mark the correct answer to each puzzle in \red{red} for the readers only, not for the LLM. For each trial, we also mark the correct and incorrect answer from the LLM in \red{red} and \blue{blue}, respectively, regardless of whether the reasoning process of the LLM is correct or not.   

\noindent \textbf{Puzzle \#7}: (Bard/ChatGPT3.5/BGPT4-C/BGPT4-B/BGPT4-P answers correctly in 0/7/2/3/10 out of 10 trials)

\noindent \underline{Input to LLMs}:\\
Please read the following question, select the correct answer from one of the options, and provide the reasoning process.\\
\noindent Question:\\
\noindent In the country of jewelries, you can trade three sapphires for one ruby. For one sapphire, you can get two flowers. How many flowers can you get for two rubies?\\
\noindent Options:\\
\noindent A: 6,	B: 8,	C: 10,	\red{D: 12},	E: 14

\noindent \underline{\textbf{\blue{Bard}} response trial \#1}: (\red{correct} answer)\\
The correct answer is: A

The question states that you can trade three sapphires for one ruby. It also states that for one sapphire, you can get two flowers. Therefore, you can trade 6 sapphires for 2 rubies. Since each sapphire gets you 2 flowers, 6 sapphires get you 12 flowers.

Therefore, 2 rubies get you 12 flowers.

\noindent \underline{\textbf{\blue{Bard}} response trial \#2}: (\blue{wrong} answer)\\
The answer is B. 8 flowers.

We can trade three sapphires for one ruby, and one sapphire can get two flowers. This means that one ruby can get six flowers. If we have two rubies, we can get 6 + 6 = 8 flowers.

\noindent \underline{\textbf{\blue{ChatGPT3.5}} response trial \#1}: (\red{correct} answer)\\
To get two rubies, we need to trade six sapphires. For one sapphire, we can get two flowers. Therefore, for six sapphires we can get 6 * 2 = $<<$6*2=12$>>$12 flowers.

Therefore, the correct answer is D: 12.

\noindent \underline{\textbf{\blue{ChatGPT3.5}} response trial \#2}: (\blue{wrong} answer)\\
To find the number of flowers that you can get for two rubies, we need to first find the number of sapphires you can get for two rubies. Since you can trade three sapphires for one ruby, you can trade two rubies for $2*\frac{1}{3}=\frac{2}{3}$ sapphires. For one sapphire, you can get two flowers, so for $\frac{2}{3}$ sapphires, you can get $\frac{2}{3}2=\frac{4}{3}$ flowers. To find the number of flowers you can get for two rubies, we must multiply the number of flowers you can get for $\frac{2}{3}$ sapphires by 3 to account for the three sapphires you need to trade for one ruby. This means that you can get $3\frac{4}{3}=\boxed{\textbf{(C) }10}$ flowers for two rubies.

\noindent \underline{\textbf{\blue{BGPT4-C}} response trial \#6}: (\red{correct} answer)\\
The correct answer is **D: 12**.

The reasoning process is as follows:

- To get two rubies, you need to trade six sapphires (3 sapphires for 1 ruby).

- For each sapphire, you can get two flowers. Therefore, for six sapphires, you can get 12 flowers (2 flowers x 6 sapphires).

- Hence, the answer is 12 flowers for two rubies.

\noindent \underline{\textbf{\blue{BGPT4-C}} response trial \#1}: (\blue{wrong} answer)\\
The correct answer is C: 10.

Reasoning process:

- To get two rubies, you need to trade six sapphires (3 sapphires for 1 ruby).

- For each sapphire, you can get two flowers (1 sapphire for 2 flowers).

- Therefore, for six sapphires, you can get 12 flowers (6 sapphires for 12 flowers).

- However, you need to give back two flowers to the trader as a fee for exchanging rubies and sapphires.

- So, the net amount of flowers you can get for two rubies is 10 (12 - 2 = 10).

\noindent \underline{\textbf{\blue{BGPT4-B}} response trial \#2}: (\red{correct} answer)\\
Let's start by calculating the number of sapphires you can get for two rubies. Since you can trade three sapphires for one ruby, you can get six sapphires for two rubies. 

Now, we know that for one sapphire, you can get two flowers. Therefore, for six sapphires, you can get 12 flowers. 

So, the answer is **D: 12**. You can get 12 flowers for two rubies. 

The reasoning process is as follows:

- 2 rubies = 6 sapphires

- 1 sapphire = 2 flowers

- Therefore, 6 sapphires = 12 flowers

\noindent \underline{\textbf{\blue{BGPT4-B}} response trial \#1}: (\blue{wrong} answer)\\
The answer is **C: 10**. 

Here's the reasoning process:

- Three sapphires can be traded for one ruby.

- For one sapphire, you can get two flowers.

- Therefore, for three sapphires, you can get six flowers.

- For two rubies, you need six sapphires (two rubies = six sapphires).

- Therefore, for two rubies, you can get 12 flowers (6 sapphires * 2 flowers per sapphire = 12 flowers).

I hope that helps! Let me know if you have any other questions.

\noindent \underline{\textbf{\blue{BGPT4-P}} response trial \#1}: (\red{correct} answer)\\
The correct answer is **D: 12**. Here's the reasoning process:

1 ruby = 3 sapphires
1 sapphire = 2 flowers

So, 1 ruby = 3 sapphires * 2 flowers/sapphire = 6 flowers
Therefore, 2 rubies = 2 * 6 flowers/ruby = **12 flowers**.

Is there anything else you would like to know?

\noindent \underline{\textbf{\blue{BGPT4-P}} response trial \#3}: (\red{correct} answer)\\
The correct answer is **D: 12**. Here's the reasoning process:

1 ruby = 3 sapphires
1 sapphire = 2 flowers
So, 1 ruby = 3 sapphires * 2 flowers/sapphire = 6 flowers
Therefore, 2 rubies = 2 * 6 flowers/ruby = **12 flowers**

Is there anything else you would like to know?

\noindent \textbf{Puzzle \#9}: (Bard/ChatGPT3.5/BGPT4-C/BGPT4-B/BGPT4-P  answers correctly in 2/1/0/0/0 out of 10 trials)

\noindent \underline{Input to LLMs}:\\
Please read the following question, select the correct answer from one of the options, and provide the reasoning process.\\
\noindent Question:\\
\noindent Brian and William are standing in line. Brian knows that there are 7 people in front of him. William knows that there is a total of 11 people in the line. If Brian is just in front of William, how many of the people in the line are behind William?\\
\noindent Options:\\
\noindent \red{A: 2},	B: 3,	C: 4,	D: 5,	E: 6

\noindent \underline{\textbf{\blue{Bard}} response trial \#4}: (\red{correct} answer)\\
The correct answer is A.

Brian knows that there are 7 people in front of him. William knows that there is a total of 11 people in the line. If Brian is just in front of William, then William must be in the 8th position in the line. This means that there are 7 people behind William.

\noindent \underline{\textbf{\blue{Bard}} response trial \#1}: (\blue{wrong} answer)\\
The correct answer is: B. 3

Brian knows that there are 7 people in front of him, so he is the 8th person in line. William knows that there is a total of 11 people in the line, so William is the 12th person in line. Since Brian is just in front of William, there are 3 people behind William.

\noindent \underline{\textbf{\blue{ChatGPT3.5}} response trial \#1}: (\blue{wrong} answer)\\
The correct answer is $\boxed{\text{(B)}\ 3}$.

To answer this question, we need to do some logical reasoning. We know that there are 7 people in front of Brian, and William is just in front of Brian, so there are a total of 8 people in front of William. We also know that there are 11 people in the line, so that means that there are 11 - 8 = $\boxed{3}$ people behind William.

\noindent \underline{\textbf{\blue{ChatGPT3.5}} response trial \#9}: (\red{correct} answer)\\
Since Brian knows that there are 7 people in front of him, we know that he is the 8th person in line.
If William is behind Brian and Brian is the 8th person in line, then William is the 9th person in line.
William knows that there is a total of 11 people in line, so we can subtract the 9 people that William knows are in front of him from the total number of people in line to find the number of people behind him.
Thus, there are 11-9 = $<<$11-9=2$>>$2 people behind William.
Therefore, the correct answer is $\boxed{\textbf{(A)}\ 2}$.

\noindent \underline{\textbf{\blue{BGPT4-C}} response trial \#1}: (\blue{wrong} answer)\\
The correct answer is B: 3.

Reasoning process:

- If Brian is just in front of William, then William is the 8th person in the line from the front.
- If there are 11 people in the line in total, then there are 11 - 8 = 3 people behind William.

\noindent \underline{\textbf{\blue{BGPT4-C}} response trial \#3}: (\blue{wrong} answer)\\
The correct answer is B: 3.

Reasoning process:

- If Brian is just in front of William, then William is the 8th person in the line from the front.

- If there are 11 people in the line in total, then there are 11 - 8 = 3 people behind William.

- Therefore, the number of people behind William is 3.

\noindent \underline{\textbf{\blue{BGPT4-B}} response trial \#1}: (\blue{wrong} answer)\\
The answer is **C: 4**. 

Here's the reasoning process:

- Brian knows that there are 7 people in front of him, so he is the 8th person in the line.

- William knows that there is a total of 11 people in the line, so he is the 12th person in the line.

- Since Brian is just in front of William, there are 11 - 8 = 3 people between them.

- Therefore, there are 3 people behind Brian and 3 + 1 = 4 people behind William.

I hope that helps! Let me know if you have any other questions.

\noindent \underline{\textbf{\blue{BGPT4-B}} response trial \#2}: (\blue{wrong} answer)\\
The answer is **C: 4**. 

The reasoning process is as follows: 

- If Brian is just in front of William, then there are 6 people behind Brian. 

- Since William knows that there are 11 people in the line, and Brian knows that there are 7 people in front of him, then there must be 4 people between Brian and William. 

Therefore, there are 4 people behind William.

\noindent \underline{\textbf{\blue{BGPT4-P}} response trial \#2}: (\blue{wrong} answer)\\
The correct answer is **C: 4**. Here's how I arrived at that answer:

If Brian knows that there are 7 people in front of him and William is just behind him, then there must be 8 people in front of William (the 7 people in front of Brian plus Brian himself). Since William knows that there is a total of 11 people in the line, then there must be **11 - 8 = 3** people behind William. However, since William himself is also in the line, we need to add him to the count. So the final answer is **3 + 1 = 4**.

\noindent \underline{\textbf{\blue{BGPT4-P}} response trial \#3}: (\blue{wrong} answer)\\
The correct answer is **C: 4**. Here's the reasoning process:

Since Brian knows that there are 7 people in front of him and William is just behind Brian, there must be 8 people in front of William (7 people in front of Brian + Brian himself). Since William knows that there is a total of 11 people in the line, there must be 11 - 8 = **3** people behind William.

Is there anything else you would like to know?

\noindent \textbf{Puzzle \#30}: (Bard/ChatGPT3.5/BGPT4-C/BGPT4-B/BGPT4-P  answers correctly in 0/0/10/0/10 out of 10 trials)

\noindent \underline{Input to LLMs}:\\
Please read the following question, select the correct answer from one of the options, and provide the reasoning process.\\
\noindent Question:\\
\noindent In a box, there are three boxes, and each one of these boxes countains three smaller boxes. How many boxes are there in total?\\
\noindent Options:\\
\noindent A: 9,	B: 10,	C: 12,	\red{D: 13},	E: 15

\noindent \underline{\textbf{\blue{Bard}} response trial \#1}: (\blue{wrong} answer)\\
The correct answer is: A. 9

The question states that there are three boxes, and each one of these boxes contains three smaller boxes. Therefore, there are a total of 3*3=9 boxes in total.

\noindent \underline{\textbf{\blue{Bard}} response trial \#2}: (\blue{wrong} answer)\\
The correct answer is A, 9. There are 3 boxes, and each one of these boxes contains 3 smaller boxes. So, there are 3*3=9 boxes in total.

\noindent \underline{\textbf{\blue{ChatGPT3.5}} response trial \#1}: (\blue{wrong} answer)\\
The correct answer is A: 9.

The question states that there is a box that contains three smaller boxes. Each of these smaller boxes contains three even smaller boxes. This means that there are a total of 3 x 3 = $<<$3*3=9$>>$9 boxes in the entire set. Therefore, the correct answer is A: 9.

\noindent \underline{\textbf{\blue{ChatGPT3.5}} response trial \#9}: (\blue{wrong} answer)\\
There are three boxes in the first box, each containing three smaller boxes. So, there are 33 = $<<$33=9$>>$9 boxes in the first box.
In the second box, there are three smaller boxes, each containing three smaller boxes. So, there are 33 = $<<$33=9$>>$9 boxes in the second box.
In the third box, there are three smaller boxes, each containing three smaller boxes. So, there are 33 = $<<$33=9$>>$9 boxes in the third box.
In total, there are 9+9+9 = $<<$9+9+9=27$>>$27 boxes.
Therefore, the correct answer is 27, which corresponds to option E.

\noindent \underline{\textbf{\blue{BGPT4-C}} response trial \#2}: (\red{correct} answer)\\
The correct answer is D: 13. The reasoning process is as follows:

- There are three boxes in the big box, so we start with 3 boxes.

- Each of these three boxes contains three smaller boxes, so we add 3 x 3 = 9 boxes to the total.

- Therefore, the total number of boxes is 3 + 9 = 12 boxes.

- However, we also have to count the big box itself, which is also a box. So we add one more box to the total.

- Hence, the final answer is 12 + 1 = 13 boxes.

\noindent \underline{\textbf{\blue{BGPT4-C}} response trial \#3}: (\red{correct} answer)\\
The correct answer is D: 13. Here is the reasoning process:

- There are three boxes in the big box, so we start with 3 boxes.

- Each of these three boxes contains three smaller boxes, so we add 3 x 3 = 9 boxes to the total.

- The final total is 3 + 9 = 12 boxes inside the big box, plus the big box itself, which makes 13 boxes in total.

\noindent \underline{\textbf{\blue{BGPT4-B}} response trial \#5}: (\blue{wrong} answer)\\
The answer is **E: 15**. 

There are three boxes, and each one of these boxes contains three smaller boxes. Therefore, there are 3 x 3 = 9 smaller boxes. In addition, there are the three larger boxes themselves, for a total of 9 + 3 = 12 boxes. Finally, the box that contains the three larger boxes is itself a box, so the total number of boxes is 12 + 1 = 13. 

However, the question asks for the total number of boxes, including the smallest ones. Therefore, the total number of boxes is 13 + 9 = 22. 

So, the answer is **E: 15**. There are 15 boxes in total.

\noindent \underline{\textbf{\blue{BGPT4-B}} response trial \#6}: (\blue{wrong} answer)\\
The answer is **E: 15**. 

There are three boxes, and each one of these boxes contains three smaller boxes. Therefore, there are a total of 9 smaller boxes. In addition, there are the three larger boxes themselves, which brings the total to 12 boxes. Finally, the box that contains the three larger boxes must also be counted, bringing the total to 15 boxes. 

I hope this helps! Let me know if you have any other questions.

\noindent \underline{\textbf{\blue{BGPT4-P}} response trial \#1}: (\red{correct} answer)\\
The correct answer is **D: 13**. There is one big box that contains three smaller boxes. Each of these three smaller boxes contains another three boxes. So in total there are 1 + 3 + (3 * 3) = **13** boxes.

\noindent \underline{\textbf{\blue{BGPT4-P}} response trial \#2}: (\red{correct} answer)\\
The correct answer is **D: 13**. There is one large box that contains three smaller boxes. Each of these three smaller boxes contains another three boxes. So there are 1 + 3 + (3 * 3) = **13** boxes in total.

\noindent \textbf{Puzzle \#38}: (Bard/ChatGPT3.5/BGPT4-C/BGPT4-B/BGPT4-P  answers correctly in 5/2/9/0/7 out of 10 trials)

\noindent \underline{Input to LLMs}:\\
Please read the following question, select the correct answer from one of the options, and provide the reasoning process.\\
\noindent Question:\\
\noindent There are twelve rooms in a building and each room has two windows and one light. Last evening, eighteen windows were lit. In how many rooms was the light off?\\
\noindent Options:\\
\noindent A: 2,	\red{B: 3},	C: 4,	D: 5,	E: 6

\noindent \underline{\textbf{\blue{Bard}} response trial \#2}: (\red{correct} answer)\\
The correct answer is B. 3

There are 18 lit windows and 12 rooms. Each room has two windows, so there are a total of 18*2 = 36 windows. If 18 windows are lit, then 36 - 18 = 18 windows are off. Since each window is in a different room, then 18 windows corresponds to 18 rooms with the light off.

\noindent \underline{\textbf{\blue{Bard}} response trial \#1}: (\blue{wrong} answer)\\
The correct answer is 2.

There are 12 rooms and 2 windows per room, so 24 windows. Eighteen windows were lit, so 24 - 18 = 6 windows were not lit. Since each room has 2 windows, 6 windows corresponds to 3 rooms. Therefore, in 3 rooms, the light was off.

\noindent \underline{\textbf{\blue{ChatGPT3.5}} response trial \#1}: (\blue{wrong} answer)\\
The correct answer is C: 4.

To find the number of rooms where the light was off, we need to find the total number of rooms that were lit and then subtract that number from the total number of rooms in the building. Since each room has two windows and we know that 18 windows were lit, then there were $18/2 = 9$ rooms that were lit. Since there are 12 rooms in total, then there were 12 - 9 = $4$ rooms where the light was off. Therefore, the correct answer is $\boxed{\text{(C) }4}$.

\noindent \underline{\textbf{\blue{ChatGPT3.5}} response trial \#4}: (\red{correct} answer)\\
There are twelve rooms and each room has two windows and one light. This means there are a total of $12 \cdot 2= 24$ windows in the building and $12$ lights. If eighteen windows were lit, then $24-18=6$ windows were off. Since each room has two windows, then $6/2=3$ rooms had their lights off. Therefore, the correct answer is $\boxed{\textbf{(B) }3}$.

\noindent \underline{\textbf{\blue{BGPT4-C}} response trial \#3}: (\red{correct} answer)\\
The correct answer is B: 3. Here is the reasoning process:

- If each room has two windows and one light, then the number of windows that are lit is twice the number of rooms that have the light on.

- Therefore, if eighteen windows were lit, then nine rooms had the light on.

- Since there are twelve rooms in total, then twelve minus nine equals three rooms had the light off.

\noindent \underline{\textbf{\blue{BGPT4-C}} response trial \#10}: (\blue{wrong} answer)\\
The correct answer is C: 4. Here is the reasoning process:

- If each room has two windows and one light, then the total number of windows in the building is 12 x 2 = 24.

- If 18 windows were lit last evening, then the number of windows that were not lit was 24 - 18 = 6.

- Since each room has two windows, the number of rooms that had both windows not lit was 6 / 2 = 3.

- Therefore, in these three rooms, the light was off.

- However, there could be another room that had one window lit and one window not lit. In this case, the light could also be off in this room.

- Therefore, the maximum number of rooms that could have the light off was 3 + 1 = 4.

\noindent \underline{\textbf{\blue{BGPT4-B}} response trial \#2}: (\blue{wrong} answer)\\
The answer is **C: 4**. 

Here's the reasoning process:

- Each room has 2 windows and 1 light.

- So, there are 12 rooms with 24 windows and 12 lights.

- If 18 windows were lit, then 6 rooms had both windows lit and 6 rooms had one window lit.

- Therefore, 12 - 6 = 6 rooms had no windows lit.

- Since each room has only one light, there were 6 rooms with the light off.

I hope that helps! Let me know if you have any other questions.

\noindent \underline{\textbf{\blue{BGPT4-B}} response trial \#5}: (\blue{wrong} answer)\\
The answer is **C: 4**. 

Here's the reasoning process:

- Since each room has two windows and one light, the number of windows is twice the number of rooms.

- If there are 12 rooms, there are 24 windows in total.

- If 18 windows were lit, then 6 windows were not lit.

- Since each room has two windows, 6 windows correspond to 3 rooms.

- Therefore, the number of rooms with the light off is 12 - 3 = 9.

\noindent \underline{\textbf{\blue{BGPT4-P}} response trial \#3}: (\red{correct} answer)\\
The correct answer is **B: 3**. Here's the reasoning process:

There are twelve rooms in a building and each room has two windows and one light. Last evening, eighteen windows were lit. This means that nine rooms had their lights on (since each room has two windows). Since there are twelve rooms in total, this means that twelve minus nine equals three rooms had their lights off.

Is there anything else you would like to know?

\noindent \underline{\textbf{\blue{BGPT4-P}} response trial \#9}: (\blue{wrong} answer)\\
The correct answer is **C: 4**. Here's how I arrived at that answer:

There are twelve rooms in the building and each room has two windows. So there are a total of 12 x 2 = **24 windows** in the building.

Last evening, eighteen windows were lit. So there were 24 - 18 = **6 windows** that were not lit.

Since each room has two windows, the number of rooms where the light was off is 6 / 2 = **3 rooms**.

However, this answer is not one of the options given. The closest option to this answer is C: 4.

\noindent \textbf{Puzzle \#47}: (Bard/ChatGPT3.5/BGPT4-C/BGPT4-B/BGPT4-P answers correctly in 0/0/1/0/0 out of 10 trials)

\noindent \underline{Input to LLMs}:\\
Please read the following question, select the correct answer from one of the options, and provide the reasoning process.\\
\noindent Question:\\
\noindent Vera invited 13 guests to her birthday party. She had 2 pizzas, and each of them was cut into 8 slices. Each person at the party ate one slice of pizza. How many slices of pizza were left over?\\
\noindent Options:\\
\noindent A: 5,	B: 4,	C: 3,	\red{D: 2},	E: 1

\noindent \underline{\textbf{\blue{Bard}} response trial \#1}: (\blue{wrong} answer)\\
The correct answer is B.

There were 2 pizzas with 8 slices each, so there were a total of 2 * 8 = 16 slices of pizza.
There were 13 guests at the party, and each person ate one slice of pizza, so a total of 13 * 1 = 13 slices of pizza were eaten.
Therefore, there were 16 - 13 = 3 slices of pizza left over.

\noindent \underline{\textbf{\blue{Bard}} response trial \#2}: (\blue{wrong} answer)\\
The correct answer is B.

There were 13 guests x 3 slices/person = 39 slices of pizza eaten.
There were 2 pizzas x 8 slices/pizza = 16 slices of pizza.
There were 16 slices - 39 slices = 4 slices left over.

Therefore, there were 4 slices of pizza left over.

\noindent \underline{\textbf{\blue{ChatGPT3.5}} response trial \#1}: (\blue{wrong} answer)\\
The correct answer is A: 5.

There were 2 pizzas and each of them was cut into 8 slices, so there were 2 x 8 = $<<$2*8=16$>>$16 slices of pizza.
Vera invited 13 guests to her party, and each person ate one slice of pizza, so 13 people ate 13 slices of pizza.
Therefore, there were 16 - 13 = $<<$16-13=3$>>$3 slices of pizza left over.
However, the question asks how many slices were left over after all guests had eaten, so we need to account for the fact that each person ate one slice of pizza.
This means that there were 3 + 13 = $<<$3+13=16$>>$16 slices of pizza in total, and each person ate one slice, so there were 16 - 13 = $<<$16-13=3$>>$3 slices left over.
Therefore, the correct answer is A: 5.

\noindent \underline{\textbf{\blue{ChatGPT3.5}} response trial \#10}: (\blue{wrong} answer)\\
Vera had 2 pizzas and each of them was cut into 8 slices, so she had a total of 2 * 8 = $<<$28=16$>>$16 slices of pizza.
She invited 13 people to the party and each person ate 1 slice of pizza, so they ate a total of 13 * 1 = $<<$131=13$>>$13 slices of pizza.
Therefore, there were 16 - 13 = $<<$16-13=3$>>$3 slices of pizza left over.
Thus, the answer is $\boxed{\textbf{(C)}\ 3}$.

\noindent \underline{\textbf{\blue{BGPT4-C}} response trial \#8}: (\red{correct} answer)\\
To solve this word problem, we need to follow these steps¹:

- Read the problem carefully. Think about what it is asking us to do. Think about the information we need.

- Solve it. Simplify, if necessary.

- Think about whether our solution makes sense (in order to check it).

Let's apply these steps to this problem:

- The problem is asking us to find how many slices of pizza were left over after Vera and her 13 guests ate one slice each. The information we need is how many pizzas Vera had, how many slices each pizza was cut into, and how many people ate one slice each.

- To solve it, we can use subtraction of fractions. We can think of each pizza as a whole number and each slice as a fraction of that whole number. For example, one slice out of eight is 1/8 of a pizza. Vera had two pizzas, so she had 2 x 8 = 16 slices in total. She and her guests ate 13 + 1 = 14 slices in total. So we can subtract 14/16 from 16/16 to find how many slices were left over: 

16/16 - 14/16 = (16 - 14)/16 = 2/16

We can simplify this fraction by dividing both the numerator and denominator by their greatest common factor (GCF), which is 2:

2/16 ÷ 2/2 = (2 ÷ 2)/(16 ÷ 2) = 1/8

So there was one-eighth of a pizza left over.

- To check our solution, we can think about whether it makes sense in the context of the problem. If Vera and her guests ate one slice each out of two pizzas that were cut into eight slices each, then they would have eaten seven-eighths of each pizza, leaving one-eighth of each pizza uneaten. Since there were two pizzas, there would be two-eighths or one-fourth of a pizza left over in total. But since our answer is one-eighth, not one-fourth, we made a mistake somewhere in our calculation.

The mistake was that we forgot to add Vera's slice to the total number of slices eaten by her guests. We should have subtracted **15**/16 from **16**/16 instead:

**16**/**16** - **15**/**16** = (**16** - **15**)/**16** = **1**/**16**

This fraction cannot be simplified further because its numerator and denominator have no common factors other than 1.

So there was one-sixteenth of a pizza left over.

Now our solution makes sense because if Vera and her guests ate one slice each out of two pizzas that were cut into eight slices each, then they would have eaten fifteen-sixteenths of each pizza, leaving one-sixteenth of each pizza uneaten. Since there were two pizzas, there would be two-sixteenths or one-eighth of a pizza left over in total.

Therefore,

The correct answer is D: **2**

The reasoning process is:

Vera had **two pizzas**, so she had **2 x 8 = 16 slices** in total.

She and her guests ate **one slice each**, so they ate **13 + 1 + 1 = 15 slices** in total.

To find how many slices were left over, we subtracted **15/16 from

Source: Conversation with Bing, 3/20/2023(1) Word Problems with Fractions | Examples \& Solutions | Smartick. https://www.smartick.com/blog/mathematics/fractions/word-problems-with-fractions/ Accessed 3/20/2023.
(2) Learn How to Solve Fraction Word Problems with Examples and Interactive .... https://www.mathgoodies.com/lessons/unit16/solve-word-problems Accessed 3/20/2023.
(3) Word Problems on Fractions: Types with Examples - Embibe. https://www.embibe.com/exams/word-problems-on-fractions/ Accessed 3/20/2023.
(4) Word Problems Calculator - Symbolab. https://www.symbolab.com/solver/word-problems-calculator Accessed 3/20/2023.

\noindent \underline{\textbf{\blue{BGPT4-C}} response trial \#1}: (\blue{wrong} answer)\\
The correct answer is B: 4. Here is the reasoning process:

- Vera invited 13 guests to her birthday party, so there were 14 people in total (including Vera).

- She had 2 pizzas, and each of them was cut into 8 slices, so there were 16 slices of pizza in total (2 x 8 = 16).

- Each person at the party ate one slice of pizza, so there were 14 slices of pizza eaten in total (14 x 1 = 14).

- To find out how many slices of pizza were left over, we need to subtract the number of slices eaten from the number of slices available: 16 - 14 = 2.

- Therefore, there were 2 slices of pizza left over.

\noindent \underline{\textbf{\blue{BGPT4-B}} response trial \#5}: (\blue{wrong} answer)\\
The answer is **B: 4**. 

Here's the reasoning process:

- Vera had 2 pizzas, and each of them was cut into 8 slices. So, she had a total of 2 x 8 = 16 slices of pizza.

- Each person at the party ate one slice of pizza. So, 13 people ate 13 slices of pizza.

- Therefore, the number of slices of pizza left over is 16 - 13 = 3.

So, the correct answer is B: 4. There are 4 slices of pizza left over.

\noindent \underline{\textbf{\blue{BGPT4-B}} response trial \#6}: (\blue{wrong} answer)\\
The answer is **B: 4**. 

Here's the reasoning process:

- Vera had 2 pizzas, and each of them was cut into 8 slices. So, she had a total of 2 x 8 = 16 slices of pizza.

- Each person at the party ate one slice of pizza. So, 13 people ate 13 slices of pizza.

- Therefore, the number of slices of pizza left over is 16 - 13 = 3.

So, there were 3 slices of pizza left over. Therefore, option B is the correct answer.

\noindent \underline{\textbf{\blue{BGPT4-P}} response trial \#1}: (\blue{wrong} answer)\\
The correct answer is **C: 3**. Here's how I arrived at that answer:

Vera had 2 pizzas and each pizza was cut into 8 slices. So she had a total of `2 * 8 = 16` slices of pizza.

There were a total of `13 + 1 = 14` people at the party (including Vera). Each person ate one slice of pizza, so `14` slices were eaten.

Subtracting the number of slices eaten from the total number of slices gives us `16 - 14 = 2` slices left over.

So there were **3** slices of pizza left over.

\noindent \underline{\textbf{\blue{BGPT4-P}} response trial \#2}: (\blue{wrong} answer)\\
The correct answer is **C: 3**. Here's the reasoning process:

Vera had 2 pizzas and each pizza was cut into 8 slices. So she had a total of 2 x 8 = **16** slices of pizza.

There were a total of 13 guests at the party and each person ate one slice of pizza. So they ate a total of 13 x 1 = **13** slices of pizza.

Subtracting the number of slices eaten from the total number of slices gives us: 
16 - 13 = **3** slices left over.

So there were **3** slices of pizza left over.

\noindent \textbf{Puzzle \#71}: (Bard/ChatGPT3.5/BGPT4-C/BGPT4-B/BGPT4-P  answers correctly in 0/4/0/4/9 out of 10 trials)

\noindent \underline{Input to LLMs}:\\
Please read the following question, select the correct answer from one of the options, and provide the reasoning process.\\
\noindent Question:\\
\noindent Two identical trains, each with 31 cars, are traveling in opposite directions. When car number 19 of one train is opposite car number 19 of the other, which car is opposite car number 12?\\
\noindent Options:\\
\noindent A: 7,	B: 12,	C: 21,	\red{D: 26},	E: 31

\noindent \underline{\textbf{\blue{Bard}} response trial \#1}: (\blue{wrong} answer)\\
The correct answer is B. 12.

The two trains are traveling in opposite directions, so the car that is 19th in one train is 19th from the end of the other train. This means that the car that is opposite car number 19 of the other train is car number 12 of the first train.

\noindent \underline{\textbf{\blue{Bard}} response trial \#4}: (\blue{wrong} answer)\\
The correct answer is: A. 7

The two trains are traveling in opposite directions, so the car that is 19th in one train is 19th from the end of the other train.

To get to car number 12, we have to go over 7 cars.

Therefore, the car opposite car number 12 is car number 7.

\noindent \underline{\textbf{\blue{ChatGPT3.5}} response trial \#1}: (\blue{wrong} answer)\\
A: 7

The car number 19 of one train is opposite car number 19 of the other. The cars are numbered starting from 1 at the front of each train, so car number 19 is 19 cars from the front of each train. There are 31 cars in each train, so 19 cars from the front of each train is also 12 cars from the back of each train.

Therefore, when car number 19 of one train is opposite car number 19 of the other, car number 12 of one train is opposite car number 12 of the other. In other words, car number 12 is opposite car number 12. Thus, the correct answer is A: 7.

Note: To check that our reasoning is correct, we can also reason as follows: when car number 19 of one train is opposite car number 19 of the other, the first 19 cars of one train are opposite the last 19 cars of the other train. Car number 12 is 7 cars from the front of one train, so it is opposite car number 7 of the other train. Thus, the correct answer is A: 7.

\noindent \underline{\textbf{\blue{ChatGPT3.5}} response trial \#3}: (\red{correct} answer)\\
First, let's represent the trains with the numbers 1 to 31. The first train will be labeled with a $1$ and the second train will be labeled with a $2$. For example, the trains will look like this:

Train 1: 1 2 3 4 5 6 7 8 9 10 11 12 13 14 15 16 17 18 19 20 21 22 23 24 25 26 27 28 29 30 31
Train 2: 31 30 29 28 27 26 25 24 23 22 21 20 19 18 17 16 15 14 13 12 11 10 9 8 7 6 5 4 3 2 1

Next, let's find the car that is opposite car 19. The trains are traveling in opposite directions, so the car that is opposite car 19 on Train 1 is car 19 on Train 2.

Next, let's find the car that is opposite car 12 on Train 1. We notice that if we move back $12$ cars on Train 1, we get to car 7. Similarly, if we move forward $12$ cars on Train 2, we get to car 21. Therefore, the car that is opposite car 12 on Train 1 is car $\boxed{\textbf{(D)}\ 26}$ on Train 2.

\noindent \underline{\textbf{\blue{BGPT4-C}} response trial \#3}: (\blue{wrong} answer)\\
The correct answer is C: 21. The reasoning process is as follows:

- When car number 19 of one train is opposite car number 19 of the other, there are 18 cars on each side of them.

- To find the car that is opposite car number 12, we need to count 18 cars from car number 12 on the same train.

- Since there are 31 cars in total, we can subtract 18 from 31 to get the car number that is opposite car number 12.

- Therefore, the car that is opposite car number 12 is car number (31 - 18) = **21**.

\noindent \underline{\textbf{\blue{BGPT4-C}} response trial \#10}: (\blue{wrong} answer)\\
The correct answer is C: 21. Here is the reasoning process:

- Since the trains are identical and have 31 cars each, we can label them as A1 to A31 and B1 to B31 from front to back.

- When car number 19 of one train is opposite car number 19 of the other, we can imagine a line connecting A19 and B19. This line divides the trains into two parts: the front part with 18 cars on each side and the back part with 12 cars on each side.

- Car number 12 of one train belongs to the back part. To find which car is opposite it, we need to count 12 cars from the line connecting A19 and B19 towards the back of the other train. This will give us car number 21 of the other train. Therefore, car number 12 of one train is opposite car number 21 of the other.

\noindent \underline{\textbf{\blue{BGPT4-B}} response trial \#10}: (\red{correct} answer)\\
The answer is **D: 26**. 

Here's the reasoning process: 

- The two trains are identical and have the same number of cars.

- When car number 19 of one train is opposite car number 19 of the other, the two cars are directly across from each other.

- Since the two trains have the same number of cars, the car directly across from car number 12 on the other train must be the same distance from the center of the train as car number 19.

- Therefore, the car directly across from car number 12 on the other train is 19 - 12 = 7 cars away from the center of the train.

- Since each train has 31 cars, the car directly across from car number 12 on the other train is 31 - 7 = **24** cars away from the front of the train.

- Therefore, the car directly across from car number 12 is car number 24 - 1 = **23**.

\noindent \underline{\textbf{\blue{BGPT4-B}} response trial \#9}: (\blue{wrong} answer)\\
The answer is **C: 21**. 

When the two trains meet, the total number of cars between them is 62. Since the trains have the same number of cars, each train has 31 cars. When car number 19 of one train is opposite car number 19 of the other, the total number of cars between them is 18. This means that there are 6 cars between car number 19 and car number 12. Since the trains have the same number of cars, there are also 6 cars between car number 19 and car number 12 on the other train. Therefore, car number 12 on one train is opposite car number **21** on the other train. 

I hope this helps! Let me know if you have any other questions.

\noindent \underline{\textbf{\blue{BGPT4-P}} response trial \#1}: (\red{correct} answer)\\
The correct answer is **D: 26**. Here's the reasoning process:

When car number 19 of one train is opposite car number 19 of the other train, it means that both trains are aligned in such a way that their corresponding cars are opposite each other. Since the trains are identical and have 31 cars each, car number 12 of one train will be opposite car number (31 - 12 + 1) = **26** of the other train.

Is there anything else you would like to know?

\noindent \underline{\textbf{\blue{BGPT4-P}} response trial \#6}: (\blue{wrong} answer)\\
The correct answer is **C: 21**. Here's how I arrived at the answer:

When car number 19 of one train is opposite car number 19 of the other train, it means that both trains are aligned in such a way that their corresponding cars are opposite each other. So car number 1 of one train is opposite car number 31 of the other train, car number 2 is opposite car number 30 and so on.

Using this logic, we can conclude that car number 12 of one train will be opposite car number (31 - (12 -1)) = **21** of the other train.

Is there anything else you would like to know?

\noindent \textbf{Puzzle \#88}: (Bard/ChatGPT3.5/BGPT4-C/BGPT4-B/BGPT4-P  answers correctly in 0/7/1/0/0 out of 10 trials)

\noindent \underline{Input to LLMs}:\\
Please read the following question, select the correct answer from one of the options, and provide the reasoning process.\\
\noindent Question:\\
\noindent 1 ice cream cone costs 1 dollar. There is a sale so you can buy 6 ice cream cones for 5 dollars. How many ice cream cones at most can you buy with 36 dollars?\\
\noindent Options:\\
\noindent A: 36,	B: 30,	C: 42,	\red{D: 43},	E: 45

\noindent \underline{\textbf{\blue{Bard}} response trial \#1}: (\blue{wrong} answer)\\
The correct answer is: A. 36

You can buy 6 ice cream cones for 5 dollars, so each ice cream cone costs 1 dollar. With 36 dollars, you can buy 36/1 = 36 ice cream cones.

\noindent \underline{\textbf{\blue{Bard}} response trial \#3}: (\blue{wrong} answer)\\
The correct answer is: B

You can buy 30 ice cream cones with 36 dollars because 36/5=6*1=6.

\noindent \underline{\textbf{\blue{ChatGPT3.5}} response trial \#1}: (\red{correct} answer)\\
The correct answer is $\boxed{\textbf{(D) } 43}$

To find the number of ice cream cones at most that you can buy with 36 dollars, we need to find the largest number of ice cream cones that is a multiple of 6 and is less than or equal to 36. To do this, we divide 36 by 6, which gives us 6 with a remainder of 0. Therefore, we can buy 6 sets of 6 ice cream cones, which is a total of $6 \times 6 = 36$ ice cream cones. However, since the question asks for the maximum number of ice cream cones that can be bought, we need to add one more ice cream cone to the total, which is $36 + 1 = \boxed{\textbf{(D) } 43}$ ice cream cones.

\noindent \underline{\textbf{\blue{ChatGPT3.5}} response trial \#2}: (\blue{wrong} answer)\\
To find the maximum number of ice cream cones you can buy with 36 dollars, we need to find the greatest multiple of 6 that is less than or equal to 36. The multiple of 6 closest to 36 is 36, but it is greater than 36, so we need to go to the next multiple of 6 below 36, which is 30. Therefore, the answer is $\boxed{\text{(B)}\ 30}$.

\noindent \underline{\textbf{\blue{BGPT4-C}} response trial \#7}: (\red{correct} answer)\\
To solve word problems, you need to follow some steps:

- Read the entire word problem carefully and understand what it is asking.

- Think about the word problem and ask yourself three questions: What exactly is the question? What do I need to find the answer? What information do I already have?

- Write on the word problem and highlight or underline key information, such as numbers and keywords that indicate what operation to use.

- Represent unknown numbers with variables and translate the rest of the problem into a mathematical expression or equation.

- Solve the problem using appropriate methods and check your answer.

For this specific word problem, here is how I would apply these steps:

- The problem is asking how many ice cream cones at most can you buy with 36 dollars given a sale price of 6 cones for 5 dollars.

- The question is how many ice cream cones can you buy. I need to find out how much one cone costs during the sale and then divide 36 by that amount. I already have the regular price of one cone (1 dollar) and the sale price of six cones (5 dollars).

- I would highlight or underline these numbers: 1 ice cream cone costs 1 dollar. 6 ice cream cones for 5 dollars. 36 dollars.

- I would let x be the number of ice cream cones I can buy. To find out how much one cone costs during the sale, I would divide 5 by 6. Then I would divide 36 by that amount to get x. The equation would be: x = 36 / (5 / 6)

- To solve this equation, I would simplify it by multiplying both sides by 6: x = (36 * 6) / 5
Then I would evaluate it using a calculator or mental math: x = (216) / 5
x = 43.2
Since I cannot buy a fraction of an ice cream cone, I would round down to get an integer value: x = 43
Therefore, the correct answer is D: **43** ice cream cones at most.

The reasoning process is based on applying general steps for solving word problems and using basic arithmetic operations such as division, multiplication, and rounding.

Source: Conversation with Bing, 3/21/2023(1) Strategies for Solving Word Problems - The Teacher Next Door. https://the-teacher-next-door.com/strategies-for-solving-word-problems/ Accessed 3/21/2023.
(2) Word Problems Calculator - Symbolab. https://www.symbolab.com/solver/word-problems-calculator Accessed 3/21/2023.
(3) Algebra Topics: Introduction to Word Problems - GCFGlobal.org. https://edu.gcfglobal.org/en/algebra-topics/introduction-to-word-problems/1/ Accessed 3/21/2023.
(4) How to Solve Word Problems in Algebra - wikiHow. https://www.wikihow.com/Solve-Word-Problems-in-Algebra Accessed 3/21/2023.
(5) Troubleshoot problems that occur when you start or use Word. https://learn.microsoft.com/en-us/office/troubleshoot/word/issues-when-start-or-use-word Accessed 3/21/2023.

\noindent \underline{\textbf{\blue{BGPT4-C}} response trial \#4}: (\blue{wrong} answer)\\
The correct answer is C: 42. Here is the reasoning process:

- If 6 ice cream cones cost 5 dollars, then one ice cream cone costs 5/6 dollars.

- To find out how many ice cream cones can be bought with 36 dollars, we need to divide 36 by 5/6.

- To divide by a fraction, we can multiply by its reciprocal. The reciprocal of 5/6 is 6/5.

- Therefore, we can multiply 36 by 6/5 to get the answer. This gives us (36 * 6) / 5 = (216 / 5) = 43.2

- However, we cannot buy a fraction of an ice cream cone, so we need to round down to the nearest whole number. This gives us 43 - 1 = **42** ice cream cones at most.

\noindent \underline{\textbf{\blue{BGPT4-B}} response trial \#1}: (\blue{wrong} answer)\\
The answer is **C: 42**. 

Here's the reasoning process:

- 6 ice cream cones cost 5 dollars, so 1 ice cream cone costs 5/6 dollars.

- With 36 dollars, you can buy 36/(5/6) = 43.2 ice cream cones.

- However, you can only buy whole ice cream cones, so you can buy at most 42 ice cream cones.

I hope this helps! Let me know if you have any other questions.

\noindent \underline{\textbf{\blue{BGPT4-B}} response trial \#4}: (\blue{wrong} answer)\\
The answer is **C: 42**. 

You can buy 30 ice cream cones for 25 dollars, which means you have 11 dollars left. You can buy 6 more ice cream cones with that money, which makes it a total of 36 ice cream cones. 

The reasoning process is as follows: 

- 6 ice cream cones cost 5 dollars, so 1 ice cream cone costs 5/6 dollars.

- You have 36 dollars, so you can buy 36/(5/6) = 43.2 ice cream cones.

- However, you can only buy whole ice cream cones, so you can buy at most 42 ice cream cones. 

I hope this helps! Let me know if you have any other questions.

\noindent \underline{\textbf{\blue{BGPT4-P}} response trial \#1}: (\blue{wrong} answer)\\
The correct answer is **E: 45**. Here's how I arrived at that answer:

With \$36, you can buy 7 sets of 6 ice cream cones for \$5 each. That's a total of 42 ice cream cones for \$35. You still have \$1 left over which you can use to buy one more ice cream cone at the regular price of \$1 per cone. So in total, you can buy 42 + 1 = **45** ice cream cones with \$36.

Is there anything else you would like to know?

\noindent \underline{\textbf{\blue{BGPT4-P}} response trial \#2}: (\blue{wrong} answer)\\
The correct answer is **E: 45**. Here's how I arrived at that answer:

With \$36, you can buy 7 sets of 6 ice cream cones for \$5 each set. That's a total of 42 ice cream cones for \$35. You still have \$1 left which you can use to buy one more ice cream cone. So in total, you can buy **45** ice cream cones with \$36.

Is there anything else I can help with?

\noindent \textbf{Puzzle \#89}: (Bard/ChatGPT3.5/BGPT4-C/BGPT4-B/BGPT4-P  answers correctly in 1/1/0/0/0 out of 10 trials)

\noindent \underline{Input to LLMs}:\\
Please read the following question, select the correct answer from one of the options, and provide the reasoning process.\\
\noindent Question:\\
\noindent How many different numbers greater than 10 and smaller than 25 with all different digits can we make by using the digits 2, 0, 1, and 8?\\
\noindent Options:\\
\noindent \red{A: 4},	B: 5,	C: 6,	D: 7,	E: 8

\noindent \underline{\textbf{\blue{Bard}} response trial \#7}: (\red{correct} answer)\\
The correct answer is 4.

There are 4 ways to make a number greater than 10 and smaller than 25 with all different digits using the digits 2, 0, 1, and 8:

2018
2108
2180
2801
Any other combination of digits will result in a number that is either not greater than 10 or not smaller than 25.

\noindent \underline{\textbf{\blue{Bard}} response trial \#3}: (\blue{wrong} answer)\\
The correct answer is C.

There are 4 different digits, and each digit can only be used once. So, there are 4! (4 factorial) = 24 possible combinations of digits. However, we need to exclude the numbers that are between 10 and 25, which are 11, 12, 13, 14, 15, 16, 17, 18, 19, 20, 21, and 22. There are 11 such numbers, so there are 24 - 11 = 13 possible numbers.

Therefore, there are 13 different numbers greater than 10 and smaller than 25 with all different digits that can be made by using the digits 2, 0, 1, and 8.


\noindent \underline{\textbf{\blue{ChatGPT3.5}} response trial \#1}: (\blue{wrong} answer)\\
There are two possibilities: either the number has the form $2xyz$ or $20yz$. We will count the number of possibilities in each case.

For the first case, $2xyz$ can be any number of the form $2xyz$ such that $10 < 2xyz < 25$. This means $xyz > 5$ and $xyz < 125$. There are $5$ possible values for $xyz$, namely $6, 7, 8, 9, 10$. Because the digits must be different, we can choose $x, y,$ and $z$ from the remaining digits, which means we can have $3$ choices for $x$, $2$ choices for $y$, and $1$ choice for $z$. Thus, there are $5 \cdot 3 \cdot 2 \cdot 1 = 30$ numbers of the form $2xyz$.

For the second case, $20yz$ can be any number of the form $20yz$ such that $10 < 20yz < 25$. This means $yz > \frac{5}{2}$ and $yz < \frac{125}{2}$. There are $3$ possible values for $yz$, namely $3, 4, 5$. Because the digits must be different, we can choose $y$ and $z$ from the remaining digits, which means we can have $2$ choices for $y$ and $1$ choice for $z$. Thus, there are $3 \cdot 2 \cdot 1 = 6$ numbers of the form $20yz$.

Therefore, there are a total of $30 + 6 = \boxed{\textbf{(E)}\ 8}$ numbers greater than 10 and smaller than 25 with all different digits that we can make by using the digits 2, 0, 1, and 8.


\noindent \underline{\textbf{\blue{ChatGPT3.5}} response trial \#10}: (\red{correct} answer)\\
We can use the PIE principle (permutation, inclusion-exclusion) to solve this problem.

First, we count the total number of permutations of the digits 2, 0, 1, and 8. There are 4 digits, so the total number of permutations is $4!=24$.

Next, we need to subtract the numbers that have a repeated digit. For example, 10, 20, 21, 28, and 80 are not valid because they have a repeated digit. We can use the inclusion-exclusion principle to count the number of numbers with repeated digits.

First, we count the number of numbers with exactly one repeated digit. For example, 110, 101, 181, and 118 are valid because they have exactly one repeated digit. There are 4 choices for the repeated digit and 3 choices for the non-repeated digits, so there are $4\cdot3=12$ numbers with exactly one repeated digit.

Next, we count the number of numbers with exactly two repeated digits. For example, 100, 110, and 101 are valid because they have exactly two repeated digits. There are 4 choices for the first repeated digit, 3 choices for the second repeated digit, and 1 choice for the non-repeated digit, so there are $4\cdot3\cdot1=12$ numbers with exactly two repeated digits.

Next, we count the number of numbers with exactly three repeated digits. For example, 100 is valid because it has exactly three repeated digits. There are 4 choices for the repeated digit, so there are $4$ numbers with exactly three repeated digits.

Finally, we count the number of numbers with exactly four repeated digits. For example, 1000 is valid because it has exactly four repeated digits. There is only 1 choice for the repeated digit, so there is only 1 number with exactly four repeated digits.

Using the inclusion-exclusion principle, the number of numbers with repeated digits is $12+12+4+1=29$.

Therefore, the number of numbers greater than 10 and smaller than 25 with all different digits is $24-29=\boxed{\textbf{(A) } 4}$.

\noindent \underline{\textbf{\blue{BGPT4-C}} response trial \#9}: (\blue{wrong} answer)\\
The correct answer is C: 6.

The reasoning process is as follows:

- To make a number greater than 10 and smaller than 25, we need to have either 1 or 2 as the first digit.

- If the first digit is 1, then the second digit can be either 8 or 0. There are two possible numbers: 18 and 10.

- If the first digit is 2, then the second digit can be either 0 or 1. There are two possible numbers: 20 and 21.

- However, we also need to consider that the digits have to be different. Therefore, we cannot use both 0 and 1 in the same number. This eliminates the number 10 and leaves us with three possible numbers: 18, 20 and 21.

- We can also make numbers by swapping the first and second digits of these three numbers. This gives us three more possible numbers: 81, 02 and 12.

- Therefore, in total, there are six different numbers greater than 10 and smaller than
25 with all different digits that can be made by using the digits
2,0 ,1 ,and8 :18 ,20 ,21 ,81 ,02 ,and12 .

\noindent \underline{\textbf{\blue{BGPT4-C}} response trial \#10}: (\blue{wrong} answer)\\
The correct answer is **C: 6**.

The reasoning process is as follows:

- To make a number greater than 10 and smaller than 25 with all different digits, we need to use either 1 or 2 as the first digit, and then choose from the remaining three digits for the second digit.

- If we use 1 as the first digit, we cannot use 0 as the second digit because that would make a number smaller than 10. So we have two choices for the second digit: either 2 or 8. This gives us two numbers: **12** and **18**.

- If we use 2 as the first digit, we can use any of the remaining three digits as the second digit. This gives us three numbers: **20**, **21**, and **28**.

- Therefore, in total, we have six different numbers that satisfy the conditions: **12**, **18**, **20**, **21**, and **28**.

\noindent \underline{\textbf{\blue{BGPT4-B}} response trial \#4}: (\blue{wrong} answer)\\
The correct answer is **C: 6**. 

The reasoning process is as follows:

- The number of possible numbers is 4! = 24.

- The number of possible numbers greater than 10 and smaller than 25 is 2! * 3! = 12.

- The number of possible numbers greater than 10 and smaller than 25 with all different digits is 6. 

The six numbers are: 12, 13, 14, 15, 21, and 23.

I hope this helps! Let me know if you have any other questions.

\noindent \underline{\textbf{\blue{BGPT4-B}} response trial \#8}: (\blue{wrong} answer)\\
The correct answer is **C: 6**. 

The numbers that can be formed are: 12, 18, 21, 28, 81, and 82. 

We can see that there are only 6 numbers that satisfy the conditions. 

I hope this helps! Let me know if you have any other questions.

\noindent \underline{\textbf{\blue{BGPT4-P}} response trial \#1}: (\blue{wrong} answer)\\
The correct answer is **C: 6**. Here's how I arrived at that answer:

We can make numbers greater than 10 and smaller than 25 with all different digits by using the digits 2, 0, 1, and 8 in two ways: either by starting with the digit 1 or by starting with the digit 2.

If we start with the digit **1**, then we can make **two** numbers: **12** and **18**.

If we start with the digit **2**, then we can make **four** numbers: **20**, **21**, **24**, and **28**.

So in total, there are six different numbers greater than 10 and smaller than 25 that can be made using these digits. Hence the correct answer is C:6.

\noindent \underline{\textbf{\blue{BGPT4-P}} response trial \#2}: (\blue{wrong} answer)\\
The correct answer is **C: 6**. Here's how I arrived at that answer:

We can make numbers greater than 10 and smaller than 25 with all different digits by using the digits 2, 0, 1, and 8 in the following ways:

- Using the digit 2 as the tens digit: We can use any of the remaining three digits (0,1 or 8) as the units digit. This gives us a total of **3** possibilities: **20**, **21**, and **28**.

- Using the digit 1 as the tens digit: We can use any of the remaining three digits (0,2 or 8) as the units digit. This gives us a total of **3** possibilities: **12**, **18**, and **10**.

So in total we have **6** different numbers greater than 10 and smaller than 25 with all different digits that we can make by using the digits 2,0,1 and 8.

\noindent \textbf{Puzzle \#90}: (\textbf{\blue{Bard}}/\textbf{\blue{ChatGPT3.5}}/\textbf{\blue{BGPT4-C}}/\textbf{\blue{BGPT4-B}}/\textbf{\blue{BGPT4-P}} answers correctly in 1/3/1/0/0 out of 10 trials)

\noindent \underline{Input to LLMs}:\\
Please read the following question, select the correct answer from one of the options, and provide the reasoning process.\\
\noindent Question:\\
\noindent A pirate has two chests. There are 10 coins in the chest on the left and the other chest is empty. Starting tomorrow, the pirate will put 1 coin in the chest on the left and 2 coins in the chest on the right every day. In how many days will the two chests have the same number of coins?\\
\noindent Options:\\
\noindent A: 5,	B: 8,	\red{C: 10},	D: 12,	E: never

\noindent \underline{\textbf{\blue{Bard}} response trial \#1}: (\red{correct} answer)\\
The correct answer is: 10

The pirate starts with 10 coins in the left chest and 0 coins in the right chest. Every day, he puts 1 coin in the left chest and 2 coins in the right chest. This means that the left chest will gain 1 coin and the right chest will gain 3 coins every day.

After 1 day, the left chest will have 11 coins and the right chest will have 2 coins. After 2 days, the left chest will have 12 coins and the right chest will have 4 coins. And so on.

We can see that the left chest is gaining 1 coin per day, while the right chest is gaining 3 coins per day. This means that the right chest will catch up to the left chest after 10 days.

After 10 days, the left chest will have 21 coins and the right chest will also have 21 coins. Therefore, the two chests will have the same number of coins after 10 days.

\noindent \underline{\textbf{\blue{Bard}} response trial \#2}: (\blue{wrong} answer)\\
The correct answer is: never.

The pirate will put 1 coin in the chest on the left and 2 coins in the chest on the right every day. This means that the chest on the right will always have 2 more coins than the chest on the left. After 1 day, the chest on the right will have 2 coins and the chest on the left will have 1 coin. After 2 days, the chest on the right will have 4 coins and the chest on the left will have 2 coins. After 3 days, the chest on the right will have 6 coins and the chest on the left will have 3 coins. And so on.

Therefore, the two chests will never have the same number of coins.

\noindent \underline{\textbf{\blue{ChatGPT3.5}} response trial \#1}: (\blue{wrong} answer)\\
The chest on the left has 10 coins and the chest on the right has 0 coins. Every day, the pirate will put 1 coin in the chest on the left and 2 coins in the chest on the right. Therefore, the number of coins in the chest on the right will be increasing by 2 every day while the number of coins in the chest on the left will be decreasing by 1 every day. This means that the number of coins in the two chests will be the same when the chest on the right has 10 coins and the chest on the left has 0 coins.

To find the number of days it will take for the two chests to have the same number of coins, we need to find the number of days it will take for the chest on the right to have 10 coins. Since the chest on the right starts with 0 coins and the pirate puts 2 coins in it every day, it will take 5 days for the chest on the right to have 10 coins (0 + 2 + 2 + 2 + 2 + 2 = 10). Therefore, the answer is $\boxed{\textbf{(A)}\ 5}$.

The reasoning process is as follows:
\begin{enumerate}
    \item We start with the chest on the left having 10 coins and the chest on the right having 0 coins.
    \item Every day, the pirate puts 1 coin in the chest on the left and 2 coins in the chest on the right.
    \item This means that the number of coins in the chest on the right will be increasing by 2 every day while the number of coins in the chest on the left will be decreasing by 1 every day.
    \item The number of coins in the two chests will be the same when the chest on the right has 10 coins and the chest on the left has 0 coins.
    \item To find the number of days it will take for the two chests to have the same number of coins, we need to find the number of days it will take for the chest on the right to have 10 coins.
    \item Since the chest on the right starts with 0 coins and the pirate puts 2 coins in it every day, it will take 5 days for the chest on the right to have 10 coins.
    \item Therefore, the answer is $\boxed{\textbf{(A)}\ 5}$.
\end{enumerate}

\noindent \underline{\textbf{\blue{ChatGPT3.5}} response trial \#2}: (\blue{wrong} answer)\\
Let $L$ and $R$ be the number of coins in the left and right chests, respectively, and let $d$ be the number of days since the pirate started putting coins in the chests. Then we have the following system of equations:
$L_{d+1} = L_d - 1$,
$R_{d+1} = R_d + 2$
We can solve these equations to find $L_d$ and $R_d$ in terms of $d$:
$L_d = 10 - d$,
$R_d = 2d$
Now we need to find the first day $d$ such that $L_d = R_d$. Solving this equation, we find that $d = \boxed{\textbf{(B) } 8}$.

\noindent \underline{\textbf{\blue{BGPT4-C}} response trial \#1}: (\blue{wrong} answer)\\
The correct answer is B: 8.

The reasoning process is as follows:

Let x be the number of days after tomorrow. Then the number of coins in the chest on the left is 10 + x and the number of coins in the chest on the right is 2x. We want to find x such that 10 + x = 2x, which means x = 10 - x. Solving for x, we get x = 5. However, this is the number of days after tomorrow, so we need to add 2 to get the total number of days from today. Therefore, the answer is 5 + 2 = 8.

\noindent \underline{\textbf{\blue{BGPT4-C}} response trial \#6}: (\blue{wrong} answer)\\
The correct answer is B: 8.

The reasoning process is as follows:

Let x be the number of days after today. Then the number of coins in the chest on the left is 10 + x and the number of coins in the chest on the right is 2x. We want to find x such that 10 + x = 2x, which means x = 10 - x. Solving for x, we get x = 5. This means that after 5 days, both chests will have 15 coins. However, this is not one of the options given. The question asks for how many days starting tomorrow, so we need to add one more day to get the final answer. Therefore, after 8 days starting tomorrow, both chests will have the same number of coins.

\noindent \underline{\textbf{\blue{BGPT4-B}} response trial \#1}: (\blue{wrong} answer)\\
The answer is **B: 8**. 

Here's the reasoning process:

- On the first day, the left chest has 10 coins and the right chest has 0 coins.

- On the second day, the left chest has 11 coins and the right chest has 2 coins.

- On the third day, the left chest has 12 coins and the right chest has 4 coins.

- On the fourth day, the left chest has 13 coins and the right chest has 6 coins.

- On the fifth day, the left chest has 14 coins and the right chest has 8 coins.

- On the sixth day, the left chest has 15 coins and the right chest has 10 coins.

- On the seventh day, the left chest has 16 coins and the right chest has 12 coins.

- On the eighth day, the left chest has 17 coins and the right chest has 14 coins.

Therefore, the two chests will have the same number of coins on the **eighth day**.

\noindent \underline{\textbf{\blue{BGPT4-B}} response trial \#2}: (\blue{wrong} answer)\\
The answer is **B: 8**. 

Here's the reasoning process:

- On the first day, the left chest has 10 coins and the right chest has 0 coins.

- On the second day, the left chest has 11 coins and the right chest has 2 coins.

- On the third day, the left chest has 12 coins and the right chest has 4 coins.

- On the fourth day, the left chest has 13 coins and the right chest has 6 coins.

- On the fifth day, the left chest has 14 coins and the right chest has 8 coins.

- On the sixth day, the left chest has 15 coins and the right chest has 10 coins.

- On the seventh day, the left chest has 16 coins and the right chest has 12 coins.

- On the eighth day, the left chest has 17 coins and the right chest has 14 coins.

Therefore, the two chests will have the same number of coins on the **eighth day**.

I hope this helps! Let me know if you have any other questions.

\noindent \underline{\textbf{\blue{BGPT4-P}} response trial \#1}: (\blue{wrong} answer)\\
The correct answer is **B: 8**.

Reasoning:
Let's say that the number of days it takes for the two chests to have the same number of coins is `x`. On day `x`, the chest on the left will have `10 + x` coins and the chest on the right will have `2x` coins. Since we want to find out when both chests have the same number of coins, we can set up an equation: `10 + x = 2x`. Solving for `x`, we get `x = 10`. However, this means that after 10 days, both chests will have 20 coins. But since we start putting coins in both chests starting tomorrow (day 1), it will take **8 days** for both chests to have an equal number of coins.

\noindent \underline{\textbf{\blue{BGPT4-P}} response trial \#4}: (\blue{wrong} answer)\\
The correct answer is **B: 8**. Here's the reasoning process:

Let's say it takes `x` days for the two chests to have the same number of coins. On day `x`, the chest on the left will have `10 + x` coins and the chest on the right will have `2x` coins. Since we're looking for when they'll have the same number of coins, we can set up an equation: `10 + x = 2x`. Solving for `x`, we get `x = 10`. So it will take **8** days for both chests to have the same number of coins.

Is there anything else you'd like to know?

\noindent \textbf{Puzzle \#91}: (Bard/ChatGPT3.5/BGPT4-C/BGPT4-B/BGPT4-P  answers correctly in 2/6/2/0/0 out of 10 trials)

\noindent \underline{Input to LLMs}:\\
Please read the following question, select the correct answer from one of the options, and provide the reasoning process.\\
\noindent Question:\\
\noindent Alice has 3 white, 2 black, and 2 gray pieces of paper. She cuts every non-black piece of paper in half. Then she cuts every non-white piece of paper in half. How many pieces of paper will she have?\\
\noindent Options:\\
\noindent A: 14,	B: 16,	C: 17,	\red{D: 18},	E: 20

\noindent \underline{\textbf{\blue{Bard}} response trial \#3}: (\red{correct} answer)\\
The correct answer is D. Alice has 3 white, 2 black, and 2 gray pieces of paper. She cuts every non-black piece of paper in half, so she cuts 3 white pieces of paper in half, leaving her with 6 pieces of white paper. She then cuts every non-white piece of paper in half, so she cuts 2 black pieces of paper in half, leaving her with 4 pieces of black paper. She also cuts 2 gray pieces of paper in half, leaving her with 4 pieces of gray paper. In total, she has 6 + 4 + 4 = 18 pieces of paper.

\noindent \underline{\textbf{\blue{Bard}} response trial \#1}: (\blue{wrong} answer)\\
The correct answer is 14.

Alice cuts every non-black piece of paper in half. This means that she cuts the white and gray pieces of paper in half. She now has 6 pieces of paper.

Then she cuts every non-white piece of paper in half. This means that she cuts the black pieces of paper in half. She now has 12 pieces of paper.

However, she has already cut 6 pieces of paper in half, so she has only 14 pieces of paper left.

\noindent \underline{\textbf{\blue{ChatGPT3.5}} response trial \#1}: (\red{correct} answer)\\
The correct answer is $\boxed{\textbf{(D)}\ 18}$.

Since Alice has 3 white pieces of paper, she does not cut any of them in half. She therefore starts with 3 white pieces.

After she cuts every non-black piece of paper in half, she has 2 black pieces and 3 white pieces. Notice that the 2 gray pieces are cut in half, so she now has 2 additional pieces of paper that are gray.

After she cuts every non-white piece of paper in half, she has 3 white pieces and 2 black pieces. Since the 2 gray pieces of paper are not white, they are cut in half, so she now has 2 additional pieces of paper that are gray.

Thus, Alice has a total of 3 white pieces + 2 black pieces + 4 gray pieces = $\boxed{18}$ pieces of paper.

\noindent \underline{\textbf{\blue{ChatGPT3.5}} response trial \#3}: (\blue{wrong} answer)\\
First, Alice cuts the white, gray, and black pieces of paper in half. This means that she now has 3 white, 1 black, and 1 gray pieces of paper.

Next, she cuts the white and gray pieces of paper in half, which means that she now has 6 white and 2 gray pieces of paper.

Therefore, Alice has a total of $6 + 1 + 2 = \boxed{\textbf{(C) } 17}$ pieces of paper.

\noindent \underline{\textbf{\blue{BGPT4-C}} response trial \#2}: (\red{correct} answer)\\
To solve word problems, you need to follow some steps:

- Read the entire word problem carefully and understand what it's asking.

- Think about the word problem and ask yourself three questions: What exactly is the question? What do I need in order to find the answer? What information do I already have?

- Write on the word problem and highlight or underline key information, such as numbers and key words that indicate what operation is needed to perform.

- Represent unknown numbers with variables and translate the rest of the problem into a mathematical expression.

- Solve the problem and check your answer.

Let me apply these steps to your question:

- The question is asking how many pieces of paper Alice will have after cutting them according to some rules.

- I need to find out how many pieces of paper Alice starts with, how many she cuts in half, and how many times she cuts them. I also need to use addition or multiplication to calculate the final number of pieces.

- The key information is: Alice has 3 white, 2 black, and 2 gray pieces of paper. She cuts every non-black piece of paper in half. Then she cuts every non-white piece of paper in half. 

- Let x be the number of pieces of paper Alice starts with. Then x = 3 + 2 + 2 = 7. She cuts every non-black piece of paper in half, so she doubles the number of white and gray pieces. Then she has 2 + (3 × 2) + (2 × 2) = 14 pieces. She cuts every non-white piece of paper in half again, so she doubles the number of black and gray pieces. Then she has (2 × 2) + (3 × 2) + (4 × 2) = 18 pieces. The mathematical expression is: x = ((x - b) × 2) + b; y = ((y - w) × 2) + w; where x is the number of pieces after cutting non-black ones, y is the number of pieces after cutting non-white ones, b is the number of black pieces, and w is the number of white pieces.

- The answer is D: 18. To check it, I can plug in the values for b and w into the expression: y = ((7 - 2) × 2 + 2) × ((14 - 6) × 2 +6); y = (10 + 2) × (16 +6); y =12×22; y=18.

I hope this helps you understand how to solve word problems better!

Source: Conversation with Bing, 3/21/2023(1) Strategies for Solving Word Problems - The Teacher Next Door. https://the-teacher-next-door.com/strategies-for-solving-word-problems/ Accessed 3/21/2023.
(2) Word Problems Calculator - Symbolab. https://www.symbolab.com/solver/word-problems-calculator Accessed 3/21/2023.
(3) Algebra Topics: Introduction to Word Problems - GCFGlobal.org. https://edu.gcfglobal.org/en/algebra-topics/introduction-to-word-problems/1/ Accessed 3/21/2023.
(4) How to Solve Word Problems in Algebra - wikiHow. https://www.wikihow.com/Solve-Word-Problems-in-Algebra Accessed 3/21/2023.
(5) Troubleshoot problems that occur when you start or use Word. https://learn.microsoft.com/en-us/office/troubleshoot/word/issues-when-start-or-use-word Accessed 3/21/2023.

\noindent \underline{\textbf{\blue{BGPT4-C}} response trial \#10}: (\blue{wrong} answer)\\
The correct answer is C: 17. Here is the reasoning process:

- Alice starts with 3 white, 2 black, and 2 gray pieces of paper. That's a total of 7 pieces.

- She cuts every non-black piece of paper in half. That means she cuts the 3 white and 2 gray pieces in half. That gives her 6 more pieces (3 white + 2 gray). Now she has a total of 13 pieces (7 original + 6 new).

- She cuts every non-white piece of paper in half. That means she cuts the 2 black and 4 gray pieces in half. That gives her 4 more pieces (2 black + 2 gray). Now she has a total of 17 pieces (13 previous + 4 new).

\noindent \underline{\textbf{\blue{BGPT4-B}} response trial \#1}: (\blue{wrong} answer)\\
The answer is **C: 17**. 

Here's the reasoning process:

- Alice has 3 white, 2 black, and 2 gray pieces of paper.

- She cuts every non-black piece of paper in half. So, she cuts the 3 white and 2 gray pieces of paper in half. Now, she has 3 white, 2 black, 1 gray, and 2 gray halves.

- She cuts every non-white piece of paper in half. So, she cuts the 2 black and 2 gray halves in half. Now, she has 3 white, 2 black, 1 gray, 2 gray halves, 1 black half, and 1 white half.

- Therefore, Alice will have a total of 17 pieces of paper.

I hope that helps! Let me know if you have any other questions.

\noindent \underline{\textbf{\blue{BGPT4-B}} response trial \#2}: (\blue{wrong} answer)\\
The answer is **C: 17**. 

Alice has 3 white, 2 black, and 2 gray pieces of paper. She cuts every non-black piece of paper in half, so she will have 3 white, 1 black, and 2 gray pieces of paper. Then she cuts every non-white piece of paper in half, so she will have 3 white, 1 black, and 4 gray pieces of paper. 

Therefore, Alice will have a total of 3 + 1 + 4 = 8 pieces of paper. Since she cut each piece of paper in half, she will have a total of 8 * 2 = 16 pieces of paper. However, we need to add the two black pieces of paper that she did not cut in half, so the final answer is 16 + 1 = 17 pieces of paper. 

I hope this helps! Let me know if you have any other questions.

\noindent \underline{\textbf{\blue{BGPT4-P}} response trial \#1}: (\blue{wrong} answer)\\
The correct answer is **C: 17**. Here's the reasoning process:

Alice starts with 3 white, 2 black and 2 gray pieces of paper. She cuts every non-black piece of paper in half. This means she cuts the 3 white and 2 gray pieces in half. So now she has (3 x 2) + (2 x 2) = **10** pieces of non-black paper.

Then she cuts every non-white piece of paper in half. This means she cuts the remaining black and gray pieces in half. So now she has (2 x 2) + (4 x 2) = **12** pieces of non-white paper.

Adding up all the pieces of paper Alice has after cutting them in half: 
3 white + 4 gray + 10 non-black = **17** pieces of paper.

So the correct answer is C: **17**.

\noindent \underline{\textbf{\blue{BGPT4-P}} response trial \#6}: (\blue{wrong} answer)\\
The correct answer is **C: 17**. Here's how I arrived at that answer:

Alice starts with 3 white, 2 black and 2 gray pieces of paper for a total of **7** pieces. She cuts every non-black piece of paper in half. This means she cuts the 3 white and the 2 gray pieces in half. This adds an additional **5** pieces to her total (3 white + 2 gray). Now she has a total of **12** pieces.

Next, she cuts every non-white piece of paper in half. This means she cuts the 2 black and the 4 gray halves in half. This adds an additional **5** pieces to her total (2 black + 4 gray halves). Now she has a total of **17** pieces.

So Alice will have a total of **17** pieces of paper.

\noindent \textbf{Puzzle \#93}: (Bard/ChatGPT3.5/BGPT4-C/BGPT4-B/BGPT4-P  answers correctly in 3/9/3/10/3 out of 10 trials)

\noindent \underline{Input to LLMs}:\\
Please read the following question, select the correct answer from one of the options, and provide the reasoning process.\\
\noindent Question:\\
\noindent The road from Anna's house to Mary's house is 16 km long. The road from Mary's house to John's house is 20 km long and the road from the crossroad to Mary's house is 9 km long. How long is the road from Anna's house to John's house?\\
\noindent Options:\\
\noindent A: 7 km,	B: 9 km,	C: 11 km,	D: 16 km,	\red{E: 18 km}

\noindent \underline{\textbf{\blue{Bard}} response trial \#5}: (\red{correct} answer)\\
The correct answer is: E. 18 km.

The distance from Anna's house to the crossroads is 16 - 9 = 7 km.
The distance from the crossroads to John's house is 20 - 9 = 11 km.
The distance from Anna's house to John's house is 7 + 11 = 18 km.

\noindent \underline{\textbf{\blue{Bard}} response trial \#1}: (\blue{wrong} answer)\\
The correct answer is A: 7 km.

The road from Anna's house to John's house is the sum of the roads from Anna's house to Mary's house and from Mary's house to John's house, minus the road from the crossroad to Mary's house.

16 + 20 - 9 = 7 km

Therefore, the road from Anna's house to John's house is 7 km long.

\noindent \underline{\textbf{\blue{ChatGPT3.5}} response trial \#1}: (\blue{wrong} answer)\\
The correct answer is D: 16 km.

To find the distance from Anna's house to John's house, we can add the distances from Anna's house to the crossroad and from the crossroad to John's house. Since the distance from Anna's house to the crossroad is the same as the distance from the crossroad to Mary's house (9 km), the distance from Anna's house to John's house is equal to 9 km + 20 km = 29 km. However, we must remember that the road from Anna's house to Mary's house is also 16 km long, so the road from Anna's house to John's house must be at most 16 km long. Therefore, the correct answer is D: 16 km.

\noindent \underline{\textbf{\blue{ChatGPT3.5}} response trial \#2}: (\red{correct} answer)\\
To find the length of the road from Anna's house to John's house, we need to add the lengths of the roads from Anna's house to Mary's house and from Mary's house to John's house. Since the road from Anna's house to Mary's house is 16 km long and the road from Mary's house to John's house is 20 km long, the road from Anna's house to John's house is 16 km + 20 km = $<<$16+20=36$>>$36 km.

Therefore, the correct answer is $\boxed{\textbf{(E)}\ 18\ \text{km}}$.

\noindent \underline{\textbf{\blue{BGPT4-C}} response trial \#6}: (\red{correct} answer)\\
This is a geometry problem that requires you to find the length of a road. Here are the steps I would use to solve it:

1. Draw a diagram of the situation. Label the points A, B, C and D for Anna's house, Mary's house, John's house and the crossroad respectively. Mark the given lengths on the diagram.

2. Identify any appropriate geometric relationships. In this case, we can see that triangle ABC is a right triangle because it has a 90-degree angle at C. We can also see that triangle BCD is an isosceles triangle because it has two equal sides of 9 km each.

3. Use the Pythagorean theorem to find the length of AC, which is the hypotenuse of triangle ABC. The Pythagorean theorem states that for any right triangle, the square of the hypotenuse is equal to the sum of the squares of the other two sides. So we have:

AC$^2$ = AB$^2$ + BC$^2$
AC$^2$ = 16$^2$ + 9$^2$
AC$^2$ = 256 + 81
AC$^2$ = 337
AC = sqrt(337)
AC ~= 18.36 km

4. Use subtraction to find the length of AD, which is one side of triangle BCD. Since BCD is an isosceles triangle, we know that AD = BD - BC. So we have:

AD = BD - BC
AD = 20 - 9
AD = 11 km

5. Add AC and AD to find the length of AJ, which is what we need to calculate to solve the problem.

AJ = AC + AD
AJ ~= 18.36 + 11
AJ ~= 29.36 km

6. Check your results by using logic and common sense. Does your answer make sense in relation to the given information? In this case, it does because AJ should be longer than AB and BD since it goes through both roads.

Therefore, I would select option E: **18 km** as my answer and provide this reasoning process as my explanation.

I hope this helps you understand how to solve geometry problems like this one.

Source: Conversation with Bing, 3/21/2023(1) How to Solve Practical Geometry Problems - UniversalClass.com. https://www.universalclass.com/articles/math/geometry/how-to-solve-practical-geometry-problems.htm Accessed 3/21/2023.
(2) How to Solve Practical Geometry Problems - UniversalClass.com. https://www.universalclass.com/articles/math/geometry/how-to-solve-practical-geometry-problems.htm Accessed 3/21/2023.
(3) Geometry Calculator - Symbolab. https://www.symbolab.com/geometry-calculator Accessed 3/21/2023.
(4) How to Do Geometry Problems: Step-By-Step Solutions - Math and Reading Help. http://mathandreadinghelp.org/\\how\_to\_do\_geometry\_problems\_step-by-step.html Accessed 3/21/2023.
(5) How to Solve an Algebra Problem : 9 Steps - Instructables. https://www.instructables.com/How-to-Solve-an-Algebra-Problem/ Accessed 3/21/2023.

\noindent \underline{\textbf{\blue{BGPT4-C}} response trial \#7}: (\blue{wrong} answer)\\
The correct answer is C: 11 km. Here is the reasoning process:

- Draw a diagram of the roads and label the distances given in the question.

- Notice that the road from Mary's house to John's house and the road from the crossroad to Mary's house form a right triangle with a hypotenuse of 20 km and a leg of 9 km.

- Use the Pythagorean theorem to find the length of the other leg of the triangle, which is the road from John's house to the crossroad. The equation is: $x^2 + 9^2 = 20^2$

- Simplify and solve for x: $x^2$ + 81 = 400 -$>$ $x^2$ = 319 -$>$ x = sqrt(319) ~= 17.86

- Subtract this length from the road from Anna's house to Mary's house to get the road from Anna's house to John's house: 16 - 17.86 ~= -1.86

- Since we cannot have a negative distance, we take the absolute value of this result: abs(-1.86) = 1.86

- Round up this result to get an integer answer: 1.86 ~= 2

- Add this result to the road from John's house to the crossroad to get the final answer: 2 + 9 = **11**

\noindent \underline{\textbf{\blue{BGPT4-B}} response trial \#1}: (\red{correct} answer)\\
The answer is **E: 18 km**. 

The road from Anna's house to Mary's house is 16 km long. The road from Mary's house to John's house is 20 km long and the road from the crossroad to Mary's house is 9 km long. 

Therefore, the road from Anna's house to John's house is 16 + 20 + 9 = 45 km long. 

Let me know if you have any other questions or if you need further clarification.

\noindent \underline{\textbf{\blue{BGPT4-B}} response trial \#2}: (\red{correct} answer)\\
The answer is **E: 18 km**. 

The road from Anna's house to Mary's house is 16 km long. The road from Mary's house to John's house is 20 km long and the road from the crossroad to Mary's house is 9 km long. 

Therefore, the road from Anna's house to John's house is 16 + 20 + 9 = 45 km long. 

Let me know if you have any other questions or if there's anything else I can help you with.

\noindent \underline{\textbf{\blue{BGPT4-P}} response trial \#2}: (\red{correct} answer)\\
The correct answer is **E: 18 km**. Here's the reasoning process:

If the road from Anna's house to Mary's house is 16 km long and the road from the crossroad to Mary's house is 9 km long, then the road from Anna's house to the crossroad must be `16 - 9 = 7` km long.

Since the road from Mary's house to John's house is 20 km long and the road from the crossroad to Mary's house is 9 km long, then the road from John's house to the crossroad must be `20 - 9 = 11` km long.

Therefore, the total length of the road from Anna's house to John's house would be `7 + 11 = 18` km.

Is there anything else you'd like me to help with?

\noindent \underline{\textbf{\blue{BGPT4-P}} response trial \#3}: (\blue{wrong} answer)\\
The correct answer is **Option C: 11 km**. Here's the reasoning process:

From the information given in the question, we can infer that Anna's house and John's house are on opposite sides of the crossroad. The road from Anna's house to Mary's house is 16 km long and the road from Mary's house to the crossroad is 9 km long. Therefore, the road from Anna's house to the crossroad is `16 - 9 = 7` km long.

Since John's house is on the other side of the crossroad and it is 20 km away from Mary's house, we can conclude that John's house must be `20 - 9 = 11` km away from the crossroad.

Therefore, the road from Anna's house to John's house must be `7 + 11 = 18` km long.
